\documentclass{article}

\usepackage{microtype}
\usepackage{graphicx}

\usepackage{booktabs} % for professional tables
\usepackage{hyperref}
\usepackage{subcaption}
\usepackage{longtable} % add this to the preamble
\usepackage{placeins}

\usepackage[accepted]{icml2025}

\usepackage{tikz}
\usepackage{amsmath}
\usepackage{amssymb}
\usepackage{mathtools}
\usepackage{amsthm}
\usepackage{subcaption}
\usepackage[justification=centering]{caption}
\usepackage[capitalize,noabbrev]{cleveref}
\usepackage{listings}
\usepackage{xcolor}

\definecolor{promptbg}{RGB}{245,245,245}
\definecolor{promptborder}{RGB}{200,200,200}

\lstdefinestyle{prompt}{
  backgroundcolor=\color{promptbg},
  basicstyle=\ttfamily\small,
  frame=single,
  rulecolor=\color{promptborder},
  breaklines=true,
  showstringspaces=false,
  captionpos=b,
  xleftmargin=1em,
  xrightmargin=1em,
  aboveskip=1em,
  belowskip=1em
}

\theoremstyle{plain}

\theoremstyle{definition}

\theoremstyle{remark}

\usepackage[textsize=tiny]{todonotes}

\icmltitlerunning{BIMgent: Towards Autonomous Building Modeling via Computer-use Agents}

\begin{document}

\twocolumn[
\icmltitle{BIMgent: Towards Autonomous Building Modeling via Computer-use Agents}




\begin{icmlauthorlist}
\icmlauthor{Zihan Deng}{aff1,aff2}
\icmlauthor{Changyu Du}{aff1,aff2}
\icmlauthor{Stavros Nousias}{aff1,aff2}
\icmlauthor{André Borrmann}{aff1,aff2}\\
\href{https://tumcms.github.io/BIMgent.github.io/}
{https://tumcms.github.io/BIMgent.github.io/}
%\icmlauthor{Firstname4 Lastname4}{sch}
%\icmlauthor{Firstname5 Lastname5}{yyy}
%\icmlauthor{Firstname6 Lastname6}{sch,yyy,comp}
%\icmlauthor{Firstname7 Lastname7}{comp}
%\icmlauthor{Firstname8 Lastname8}{sch}
%\icmlauthor{Firstname8 Lastname8}{yyy,comp}
\end{icmlauthorlist}

\icmlaffiliation{aff1}{Chair of Computing in Civil and Building Engineering, Technical University of Munich, Germany}
\icmlaffiliation{aff2}{TUM Georg Nemetschek Institute, Munich, Germany}

\icmlcorrespondingauthor{Zihan Deng}{zihan.deng@tum.de}
\icmlcorrespondingauthor{Changyu Du}{changyu.du@tum.de}

\icmlkeywords{Machine Learning, ICML}

\vskip 0.3in
]



\printAffiliationsAndNotice{}  % leave blank if no need to mention equal contribution

\begin{abstract}
Existing computer-use agents primarily focus on general-purpose desktop automation tasks, with limited exploration of their application in highly specialized domains. In particular, the 3D building modeling process in the Architecture, Engineering, and Construction (AEC) sector involves open-ended design tasks and complex interaction patterns within Building Information Modeling (BIM) authoring software, which has yet to be thoroughly addressed by current studies. In this paper, we propose \textbf{BIMgent}, an agentic framework powered by multimodal large language models (LLMs), designed to enable autonomous building model authoring via graphical user interface (GUI) operations. BIMgent automates the architectural building modeling process, including multimodal input for conceptual design, planning of software-specific workflows, and efficient execution of the authoring GUI actions. We evaluate BIMgent on real-world building modeling tasks, including both text-based conceptual design generation and reconstruction from existing building design. The design quality achieved by BIMgent was found to be reasonable. Its operations achieved a 32\% success rate, whereas all baseline models failed to complete the tasks (0\% success rate). Results demonstrate that BIMgent effectively reduces manual workload while preserving design intent, highlighting its potential for practical deployment in real-world architectural modeling scenarios. Code available at: \href{https://github.com/ZihanDDD/BIMgent}{https://github.com/ZihanDDD/BIMgent}

\end{abstract}

\begin{figure}[ht]
    \centering
    \includegraphics[width=\columnwidth]{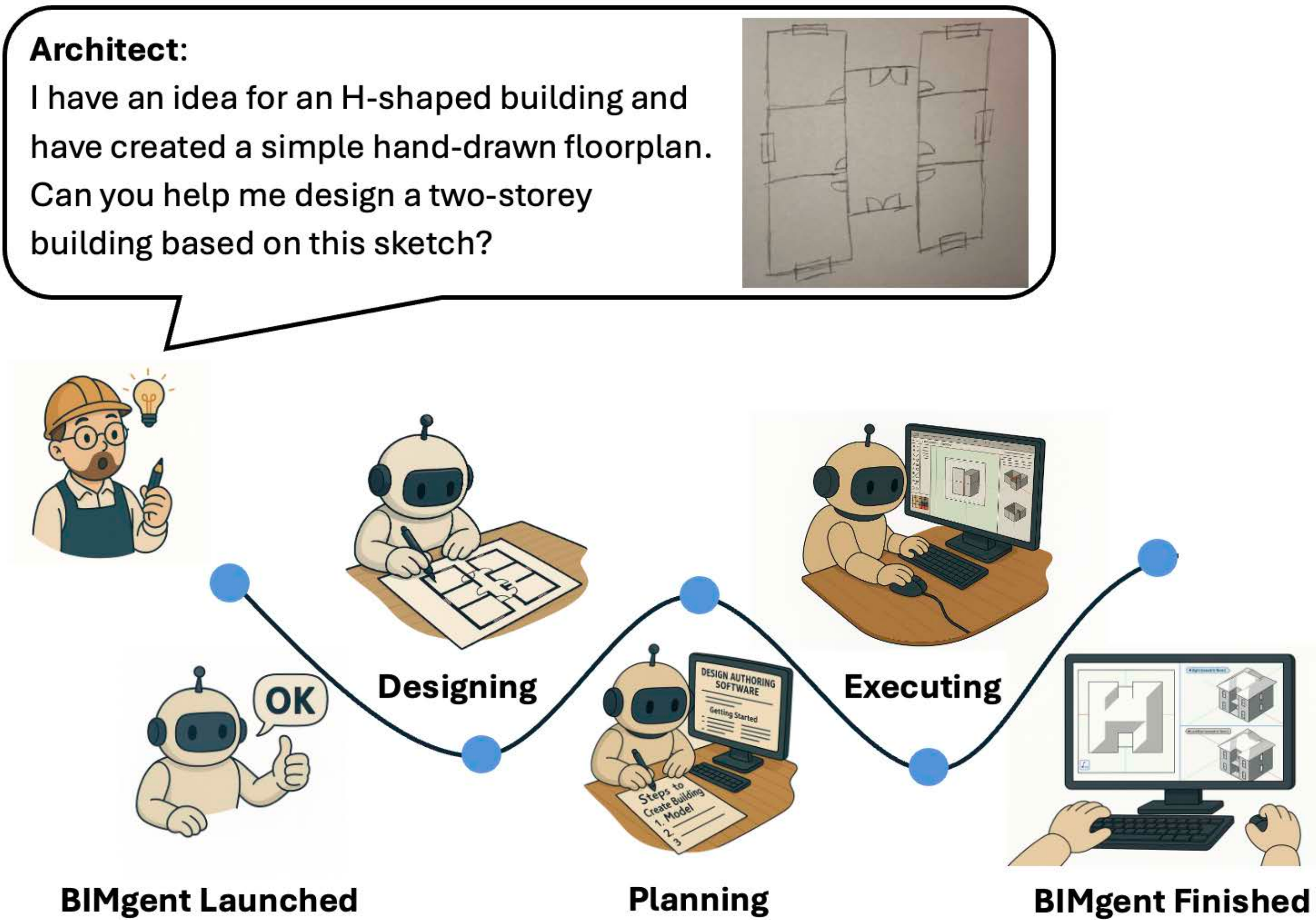}
    \caption{The \textbf{BIMgent} framework, enabling the architectural building modeling process to be performed autonomously through computer control.
 }
    \label{fig:BIMgentgeneral}
\end{figure}
 
\section{Introduction}
\label{Introduction}

To achieve generality in current autonomous agents, researchers are exploring computer-use agents that operate directly through graphical user interfaces (GUIs) \cite{anthropic2025computeruse, openai2025computeragent}. These agents perceive the same screens as humans, generate plans based on the current GUI state, and produce keyboard and mouse actions to perform tasks autonomously \cite{zhang2411large}. Today, many computer-use agents are being developed to automate tasks across a variety of environments, including the web \cite{zhengGPT4VisionGeneralistWeb, zhangUFO2DesktopAgentOS2025, zhengSkillWeaverWebAgents2025}, mobile devices \cite{wuReachAgentEnhancingMobile2025, zhengVEMEnvironmentFreeExploration2025}, and video games \cite{teamScalingInstructableAgents2024, tan2024cradle}.

However, computer-use agents are still underexplored in the Architecture, Engineering, and Construction (AEC) sector. In recent years, Building Information Modeling (BIM) has become indispensable. A BIM model is a digital representation that captures not only the 3D geometry of a building but also includes rich semantic and topological information, enabling support throughout the building’s entire lifecycle \cite{borrmann2018building}.
Before actual construction begins, architects and engineers typically design and create BIM models using professional BIM authoring software ~\cite{baduge2022artificial}.
However, there are two main challenges in the modeling process using such software. First, the commands and GUI of design software are often highly complex, resulting in a steep learning curve and high training costs \cite{hossain2022introducing}. Second, the design and modeling workflow involves numerous repetitive operations, which further increases the manual effort and time consumption \cite{heaton2019design}.

Computer-use agents can automate the building modeling process by interacting directly with the software GUI, replacing cumbersome manual operations \cite{agashe2024agent}. Despite extensive research on computer-use agents in other domains, their application in highly specialized building designs poses unique challenges:
(1) Agents must understand the conceptual intent of human design and accurately translate it into command flows within the BIM authoring software for 3D building modeling. 
(2) BIM authoring software exposes highly parameterized operations and multiple interaction modes. Agents need robust strategies to navigate these options.
(3) The richly detailed, noise-filled GUI of design software can overwhelm vision-based agents. Filtering out irrelevant elements while retaining essential visual cues is crucial for dependable operation.
(4) Modeling a building involves hundreds of interdependent operations. Agents must not only plan these steps efficiently but also manage error recovery and state tracking across a lengthy workflow.
% \begin{itemize}
% \item How can the agent understand the conceptual intent of human design and accurately translate it into complex command flows within the design software to generate the corresponding 3D building model? 
% \item Design authoring software involves highly parameterized operations and multiple interaction modes. How can these complexities be effectively interpreted and managed? 
% \item Design authoring software's GUI is typically complex and information-dense. How can we effectively handle numerous irrelevant visual elements and noise present in screenshots?
% \item How can the hundreds of interdependent operation steps typically involved in building modeling be efficiently planned and managed?

% \end{itemize}

In response, we introduce \textbf{BIMgent}, an agentic framework designed for the autonomous architectural building modeling process, as illustrated in Figure~\ref{fig:BIMgentgeneral}. BIMgent is capable of transforming multimodal design intents, including textual building descriptions or rough 2D floorplan sketches, into a 3D BIM model. It not only goes beyond basic interface-level tasks to handle open-ended design generation but also incorporates domain-specific knowledge to plan and execute modeling tasks within complex software environments.

To enhance the GUI agent's capability to operate the design authoring software, we propose a hierarchical planning structure. A high-level planner is designed for generating general design steps. It decomposes the overall modeling workflow into element-level steps (e.g., design layers, walls, etc.). Each step is then passed to a low-level planner, which retrieves relevant information from official software documentation to learn the usage of tools and commands, and generate appropriate GUI actions. We design two distinct workflows for action execution. The \textit{Pure-Action Workflow} handles tasks such as keyboard shortcuts and element placement actions, which are common usage patterns in design software. GUI pixel-level coordinates required for keyboard and mouse operations are either mapped from the floorplan image or retrieved from the software documentation. Inspired by speculative multi-action execution in UFO-2~\cite{zhangUFO2DesktopAgentOS2025}, we pre-generate the entire action sequence during the planning phase, rather than issuing one action at a time through repeated agent calls during execution to reduce latency. For the \textit{Vision-Driven Workflow} that requires fine-grained information comprehension in GUI, we propose a dynamic GUI grounding method that reduces visual noise by restricting the grounding process to regions associated with relevant interactions for action generation.  
Considering that the modeling tasks often require hundreds of sequential actions, where a single mistake can lead to cascading errors, we use LLMs as judges to evaluate each action step and provide real-time feedback to correct mistakes, enabling BIMgent to self-reflect.

Existing benchmarks for GUI agents primarily focus on web, mobile, or office software \cite{bonatti2024windows, xie2024osworld}. 
To systematically assess the BIMgent, we design a Mini Building Benchmark tailored to assess GUI agent performance in building modeling scenarios. It includes 25 real-world building modeling tasks, which evaluate the agent’s ability to handle open-ended design requirements (design evaluation) and performance throughout the modeling process within the BIM authoring software (operation evaluation).
The experimental results show that BIMgent achieves an average score of 3 out of 5 across six critical design evaluation criteria. In terms of operation, it achieves a 32\% end-to-end success rate across 25 design tasks. Notably, when decomposing the building modeling tasks, it achieves success rates of 86.58\% and 95.12\% in creating repetitive and redundant elements: walls and openings respectively. By contrast, the strongest baseline model finished none of the end-to-end tasks (0\%) and achieves only 31.70\% and 35.36\% success on walls and openings respectively in our Mini Building Benchmark.  These results demonstrate its potential to significantly reduce human effort in the building modeling process.

\section{Related Work}

\textbf{Generative AI in AEC. }
The advent of generative AI has transformed the architecture, engineering, and construction (AEC) domain. Luo and Huang \yrcite{luo2022floorplangan} proposed FloorplanGAN, which integrates vector-based generation with raster-based discrimination for architectural floorplan generation. As the field shifts toward 3D design, Ennemoser and Mayrhofer-Hufnagl \yrcite{ennemoser2023design} introduced a 3DGAN model that reconstructs architectural forms through a voxel-to-image-to-voxel pipeline; however, their approach emphasizes geometry and lacks semantic detail. Addressing this gap, Gao et al. \yrcite{gao2024diffcad} developed DiffCAD, a weakly supervised probabilistic model that retrieves and aligns CAD models from RGB images. They essentially have the ability to handle open-ended tasks and fulfill building design requirements. However, existing methods still lack the capability to accept multimodal inputs for handling design changes.

\textbf{Multimodal LLM Agents.}
Recent multimodal LLM agents have demonstrated strong capabilities in interacting with software environments, even when addressing open-ended design tasks. For example, in video game environments, Voyager leveraged LLMs to autonomously explore and acquire diverse skills in Minecraft, though it relied on internal APIs for action execution \cite{wangVoyagerOpenEndedEmbodied2023}. TWOSOME combined LLMs with reinforcement learning to improve decision-making in complex scenarios \cite{tan2024true}. In the domain of 3D scene generation, Hu et al. \yrcite{hu2024scenecraft} translated natural language into 3D environments by implementing an LLM agent that generates Python code. Similarly, Du et al. \yrcite{du2024text2bim} used LLMs to create multiple early-stage building models through internal APIs. However, these methods face limitations due to their reliance on software-specific APIs, which restrict generalizability across different platforms and tools.

\textbf{GUI Agents.} To address the limitations posed by API restrictions, recent GUI agents have demonstrated strong proficiency in interface manipulation. State-of-the-art systems such as UFO-2 \cite{ren2020ufo}, AgentS2 \cite{agashe2025agent}, SEEACT \cite{zhengGPT4VisionGeneralistWeb}, and FRIDAY \cite{wu2024copilot} have achieved high performance on GUI-based tasks. Their integration of strategies such as knowledge retrieval and hierarchical planning further enhances their efficiency and ability to handle complex, multi-step tasks across web environments. Similarly, Anthropic \cite{anthropic2025computeruse} and OpenAI \cite{openai2025computeragent} have developed screenshot-driven agents that automate operations using visual input rather than relying on APIs.

However, these systems face limitations when applied to more complex, domain-specific environments. Most existing use cases are centered around web-based interfaces, which tend to be relatively static and less complex. In contrast, adapting to environments like BIM authoring software is significantly more challenging due to complex GUIs and intricate multi-step operations. A comparable situation can be observed in video game environments, which also demand creativity and involve complex usage patterns. For example, SIMA is a GUI agent that interacts via keyboard and mouse across diverse 3D video game environments \cite{teamScalingInstructableAgents2024}.  VillagerAgent demonstrated the ability to manage complex task dependencies in large open-ended environments in Minecraft \cite{dong2024villageragent}. Similarly, Tan et al. \yrcite{tan2024cradle} proposed a generalized GUI agent framework Cradle, which is capable of operating across both video games and web applications. Despite these advances, a shared limitation among these agents is the lack of well-defined task completion signals and standardized benchmarks, which hinders consistent evaluation. As a result, many of these works design their own tasks and experimental protocols. Compared to video games, the building modeling process involves less real-time animation but poses its own challenges, such as an open-ended design process, non-intuitive operations, more complex planning and management, and intricate GUIs that are harder to interpret and ground.

\begin{figure*}[ht]
\centering
\includegraphics[width=\textwidth, trim=0 10 0 0, clip]{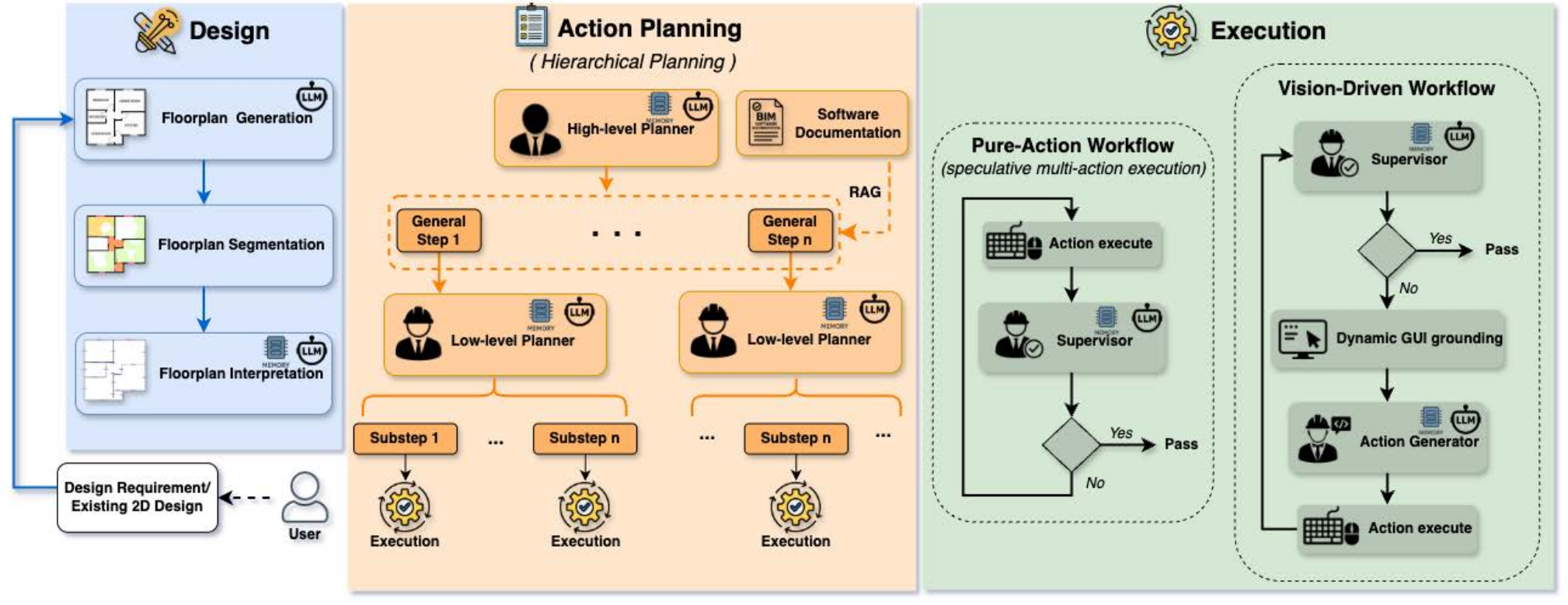}
\caption{Overview of the BIMgent framework. Given the multimodal design requirements provided by the user, the \textit{Design Layer} first transforms them into a refined floorplan and extracts the necessary semantic and geometric information to guide the modeling process.
Based on the interpreted design information and domain knowledge, the \textit{Action Planning Layer} hierarchically organizes the modeling procedure and decomposes it into detailed substeps, guided by the official software documentation. These substeps are then executed through specialized action workflows in the \textit{Execution Layer}, each equipped with verification mechanisms.
Execution trajectories are stored in a memory module, which supports both self-reflection and cooperation among different parts of the framework.}
\label{fig:BIMgent-workflow}
\end{figure*}

\section{BIMgent Framework}

\textbf{BIMgent} enables an autonomous architectural building modeling process through computer control, spanning from design concept interpretation to final 3D building modeling.
As depicted in Figure~\ref{fig:BIMgent-workflow}, the framework consists of three key layers:
(1) \textbf{Design Layer} transforms conceptual or existing designs into 2D floorplans aligned with the GUI coordinate;
(2) \textbf{Action Planning Layer} generates knowledge-based, hierarchical operation steps based on software documentation; and
(3) \textbf{Execution Layer} executes actions under agent-based supervision to complete the modeling workflow. 
The details of each component are explained in the following sections.
% BIMgent is powered by various LLMs, with each model assigned to tasks that match its strengths. A long-term memory is maintained throughout the entire process, enabling cooperation among components and supporting self-improvement.

\subsection{Design Layer}

The Design Layer is responsible for transforming a building description or existing design sketches into a 2D floorplan image. 
In contrast to existing methods that stop at image generation, this layer further extracts necessary design information and maps the image resolution-level coordinates of floorplan elements to the pixel-level coordinates, enabling downstream GUI grounding, planning, and execution.

\textbf{Floorplan Generation.} We explore three approaches to generate 2D floorplan representations from design intent, specifically, (1) we adopt a generative adversarial network (GAN)-based image generation method~\cite{nauata2021house, fu2024anyhome} to generate floorplans from textual prompts; (2) we leverage LLMs to produce SVG floorplans from textual descriptions or image input; and (3) we employ advanced multimodal LLMs to directly generate floorplan images either from text or conditioned on existing floorplan visuals. Based on empirical comparison, we find that multimodal LLMs demonstrate better capability in translating both abstract design intent and existing visual layouts into coherent and functional 2D floorplans. This enables our framework to support both text-to-building generation and floorplan-to-building transformation. Detailed comparisons of the three approaches are presented in the Appendix \ref{compare_floorplan}.

\textbf{Floorplan Segmentation.} Despite the strong image generation and understanding capabilities of current LLMs, they still exhibit limitations in recognizing and localizing accurate architectural components within floorplans. To address this, we integrate a floorplan segmentation model to identify and classify architectural elements such as walls and openings in the generated floorplan.

\textbf{Floorplan Interpretation.} We further employ an additional multimodal LLM to proofread and enhance the results, enabling the extraction of more fine-grained and accurate design details for downstream GUI grounding. For example, it helps distinguish between internal and external walls, or between doors and windows among the openings.
Following classification, we apply a rule-based algorithm to map the image-resolution coordinates of the identified components  \((x_i, y_i)\) to their corresponding pixel-level coordinates \((x_{\text{gui}}, y_{\text{gui}})\) in the GUI. This mapping can be expressed by the following formula:

\[
x_{\text{gui}} = \frac{x_i}{w_{\text{img}}} \cdot w_{\text{gui}}, \quad
y_{\text{gui}} = \frac{y_i}{h_{\text{img}}} \cdot h_{\text{gui}}
\]

Here, \(w_{\text{img}}\) and \(h_{\text{img}}\) denote the width and height of the image resolution, while \(w_{\text{gui}}\) and \(h_{\text{gui}}\) refer to the dimensions of the GUI design panel of the BIM authoring software. This scaling is used to convert coordinates from the image space to their corresponding pixel locations in the GUI.

\subsection{Action Planning Layer}
The entire building modeling process typically involves hundreds of sequential steps. To mitigate errors and improve accuracy, we designed a hierarchical planning process that employs two agents for authoring software action planning: the high-level planner and the low-level planner. 

\textbf{High-Level Planner.}  After analyzing modeling patterns of several architects, we designed a high-level planner that generates a sequence of general steps based on standard building modeling workflows. For example, the typical process begins with setting up corresponding design layers, followed by creating walls and other elements in a high-level plan. This design imitates the architects' modeling thought process. Additionally, the planner identifies which specific elements from the generated floorplan should be created or configured in each step.

\textbf{Low-Level Planner.} The detailed modeling action steps in the authoring software are complex and involve multiple operation modes, which vary across different users. To handle this complexity, we leverage the official software documentation and adopt a retrieval-augmented generation (RAG) approach \cite{Du:2024:Copilot_BIM}. This allows the agent to explore autonomously and learn software usage dynamically like humans, rather than hard-coding static actions within prompts.

\begin{figure}[h!]
  \centering
  \includegraphics[width=\columnwidth]{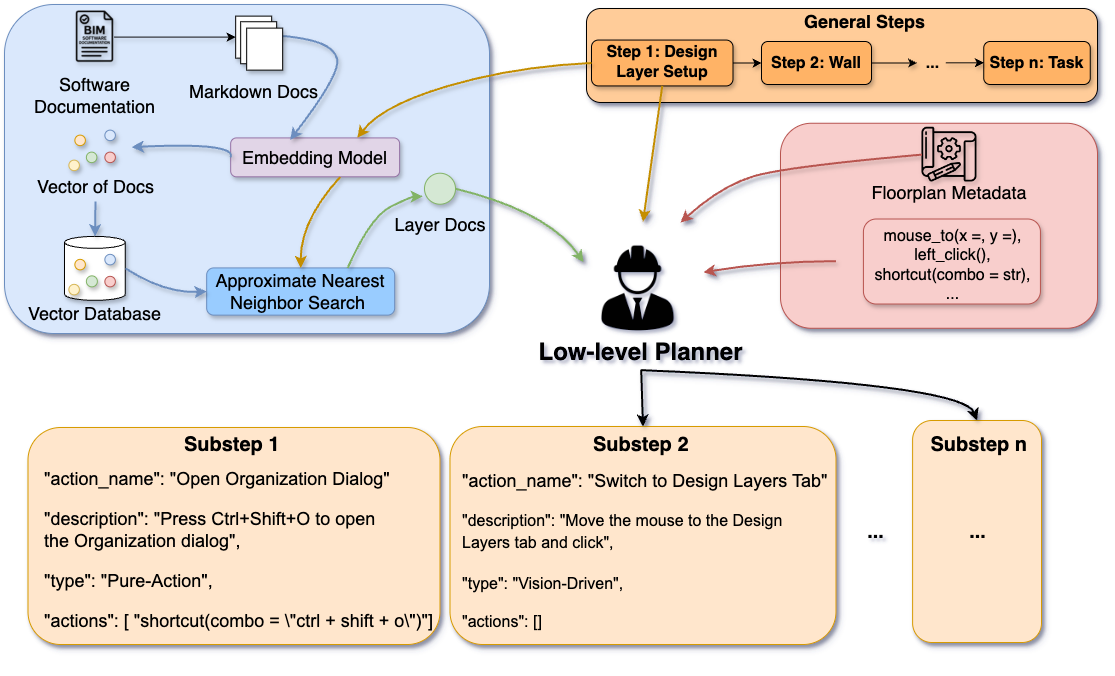}
  \caption{Low-level planner. The general steps generated by the high-level planner are embedded and used to query the official documentation to retrieve the most relevant detailed guidance. Combined with floorplan metadata and the necessary action definitions, these results are forwarded to the agent for detailed subtask generation.}

  \label{low-level-plan}
\end{figure}

As shown in Figure \ref{low-level-plan}, the documentation is embedded into vector representations and stored in a vector database.
Given a general step from the high-level planner, the most relevant sections from the documentation are retrieved. The low-level planner then references both the retrieved documentation and general steps from the high-level planner to generate detailed substeps. It breaks down each general step into multiple actionable substeps, generating them iteratively until the full sequence is completed. We identify two types of substeps generated from the low-level planner, which cover the main usage patterns of human designers interacting with the BIM authoring software: \textit{Vision-Driven} and \textit{Pure-Action}. 

\textit{Vision-Driven} substeps (e.g., Substep 2 in Figure \ref{low-level-plan}) require grounding in the GUI, such as switching tabs or setting up parameters, which necessitate specific pixel-level coordinates for accurate interactions.
These substeps are not directly converted into actions at this stage due to the lack of visual input.
In contrast, \textit{Pure-Action} substeps (e.g., Substep 1), such as issuing keyboard shortcuts and placing elements, are deterministic actions, with information typically available in the floorplan metadata or software documentation.
The low-level planner directly generates executable actions for Pure-Action tasks without additional GUI grounding. The available actions are detailed in Appendix \ref{app:action_def}.

\subsection{Execution Layer}

The final execution layer implements the planned GUI operations to model buildings within the design software.
It sequentially executes the substeps produced by the low-level planner for each general step. Once all substeps for the current general step are completed, it proceeds to the next general step and repeats the process. 
For the two types of substeps planned by the low-level planner, we accordingly designed dedicated workflows for their execution.

\textbf{Pure-Action Workflow.} Pure-action substeps are executed directly through the planned action sequences. We implement speculative multi-action execution~\cite{zhangUFO2DesktopAgentOS2025}, allowing each substep to be executed without requiring separate API calls for individual actions. After each action is completed, the supervisor is invoked. The supervisor receives the current GUI screenshot and analyzes the result. A common feature of BIM authoring software is that the GUI typically displays metadata about the currently created building elements. Leveraging this, the agent inspects the displayed information to verify whether the created element's type and semantic attributes are correct. If the result is valid, the workflow proceeds to the next task; otherwise, the failed substep is undone and regenerated by the supervisor agent. A detailed visualization of the process is provided in Appendix~\ref{trajectories}.

\begin{figure}[h!]
  \centering
  \includegraphics[width=\columnwidth]{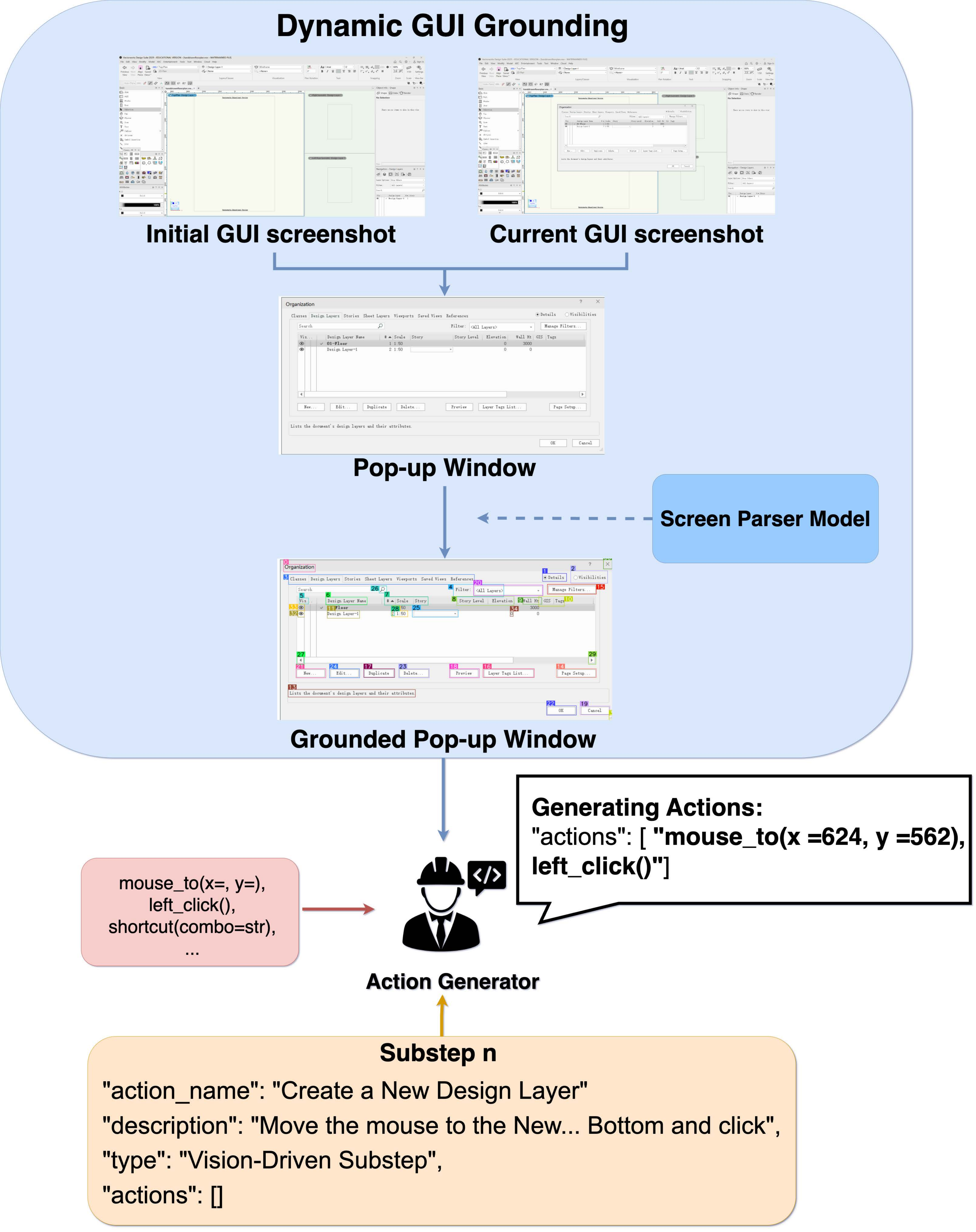}
  \caption{Dynamic GUI Grounding. The initial GUI screenshot is subtracted from the current GUI screenshot to highlight changes, allowing pop-up windows to be specifically visualized and reducing noise during the grounding process. The grounded pop-up window is then passed to the action generator, which combined with the relevant action definitions and substep information, produces the corresponding actions.
}
  \label{dynamic}
\end{figure}

\textbf{Vision-Driven Workflow.} Vision-driven substeps require dynamic action generation based on the GUI screenshot state. In this workflow, the supervisor agent is invoked first. It captures a GUI screenshot and assesses the current GUI state. If the current state satisfies the substep requirements, the system proceeds to the next substep. If not, we apply a dynamic GUI grounding mechanism for accurate interaction. 
As shown in Figure \ref{dynamic}, to detect visual changes, a screenshot is captured at the beginning of each general step. For example, before starting general step 1, an initial screenshot is taken as a reference. The current GUI screenshot is compared with the initial GUI screenshot to detect visual changes within the differing region, typically a pop-up window. The grounding process then focuses exclusively on this region. This design choice is based on the observation that, in design software, Vision-Driven substeps usually involve parameter settings, which are commonly presented through dedicated pop-up windows. As the surrounding interface remains largely static and irrelevant to the current task, excluding it reduces visual noise. This approach mimics human behavior, where attention is naturally directed toward the changing parts of the interface, resulting in more accurate and efficient grounding. 
A screen parser model converts screenshots of pop-up windows into structured representations with UI element bounding boxes and text descriptions (i.e., Set-of-Marks). These representations are then passed to a VLM-based action generator, which generates and executes the appropriate actions based on the grounding result. 
Finally, the supervisor agent re-evaluates the GUI state to verify the success of the operation.

\textbf{Reflection.} 
In all workflows, a supervisor is integrated to monitor the GUI state and assess whether the current substeps have been successfully completed. In the Vision-Driven Workflow, if a substep fails, the supervisor provides a failure reason to guide future adjustments and support the subsequent Action Generator Agent in regenerating actions. In contrast, in the Pure-Action Workflow, where the required actions are relatively simple, the Supervisor Agent regenerates the actions directly without delegating to another agent.

\section{Experiments}

\subsection{Implementation details}

\textbf{Hybrid Multi-Agent Framework.} We employ a hybrid of different LLMs/VLMs as the backbones. We choose OpenAI's gpt-image-1 for floorplan generation due to its advanced image generation and editing capabilities \cite{openai2025imagegeneration}. For the interpretation floorplan component, we employ Gemini 2.5 Pro, which demonstrates state-of-the-art image reasoning capabilities \cite{team2023gemini}. For action planning, including the high-level planner, low-level planner, and action generator, we utilize GPT-4.1 to handle more complex reasoning and instruction-following tasks \cite{openai2025_41}. For the supervisor and action generator, we use o4-mini due to its lower cost and faster response time \cite{openai2025_4o}.

\textbf{Non-Agent Components.}
Our framework incorporates several non-LLM components.
For floorplan segmentation, we adopt the DeepFloorPlan~\cite{zeng2019deep}, a multi-task network trained on about 12k annotated plans that jointly predicts room-boundary primitives (walls and openings) and room-type masks.
For software documentation and task embedding, we use OpenAI’s text-embedding-3-small \cite{openai2024embeding}.
Finally, during the dynamic GUI grounding process, we employ Omni-Parser-v2 \cite{yu2025omniparser} for accurate screen parsing and component detection.

% Due to varying building model requirements, such as the number of floors, it is difficult to standardize the overall time and number of steps. Therefore, we limit the number of redo attempts for each substep during evaluation. 

\subsection{Mini Building Benchmark Introduction}

Compared to existing computer-use benchmarks \cite{zhou2023webarena, xie2024osworld, deng2023mind2web}, where each task typically involves around 10 steps, building modeling tasks are significantly more complex, with each design task usually requiring approximately 100 action steps on average. To evaluate our system, we constructed a custom Mini Building Benchmark consisting of 25 real-world 3D building modeling tasks, all executed within the BIM authoring software Vectorworks. The benchmark includes five tasks for generating 3D buildings from pure textual design requirements, five based on hand-sketched 2D floorplan images, five sourced from the CubiCasa5K floorplan dataset \cite{kalervo2019cubicasa5k}, five from hand-sketched floorplans with additional modification requirements, and five from CubiCasa5K floorplans with modification requirements. In total, the benchmark involves over 2000 action steps. Further details are provided in the Appendix \ref{bench}.

% \subsection{Evaluation Method}
Unlike existing research benchmarks, building modeling tasks lack clear signals for automated evaluation, which makes it difficult to objectively determine whether a task has been successfully completed. To address this, the evaluation process is divided into two phases: \textbf{design evaluation} and \textbf{operation evaluation}. We conduct human evaluation based on predefined criteria, with assessments performed by architects for design and the final 3D building model. The details are shown in the Appendix \ref{Criteria}. Because no existing computer-use agents are specifically designed for autonomous building modeling, we choose GPT-4o and Claude 3.7 as baseline models.

\begin{table*}[t]
\caption{Success rates (\%) on the proposed Mini Building Benchmark test set consisting of 25 building modeling tasks, along with results from the ablation study. N/A indicates 0\% success rate.}
\label{tab:sroftool}
\vskip 0.15in
\begin{center}
\begin{small}
\begin{sc}
\begin{tabular}{lcccccccc}
\toprule
Method & End-to-End (25) & Layer (41) & Wall (82) & Slab (41) & Openings (82) & Roof (25)\\
\midrule

gpt-4o & N/A& N/A & 4.87 & 2.43 & 12.19 & N/A \\
claude 3.7 & N/A & N/A & 31.70 & 21.95 & 35.36 & N/A \\
\midrule
BIMgent             & \textbf{32.00} &  \textbf{46.34} & \textbf{86.58} & \textbf{73.81} & \textbf{95.12}& \textbf{60.00} \\
w/o Dynamic GUI Grounding  & 13.33  & 34.15 & 86.58 & 73.81 & 92.68  & 38.46 \\
w/o Supervision           & 11.53  & 24.19 & 84.14 & 70.73 & 92.68 & 48.00 \\
w/o Hierarchical Planning & N/A & 2.43 & 41.46 & 58.53 & 46.34 & 12.00 \\

\bottomrule
\end{tabular}
\end{sc}
\end{small}
\end{center}
\vskip -0.1in
\end{table*}

\section{Results and Analysis}

\subsection{Experimental Results}

\begin{figure}[h!]
  \centering
  \includegraphics[width=\columnwidth]{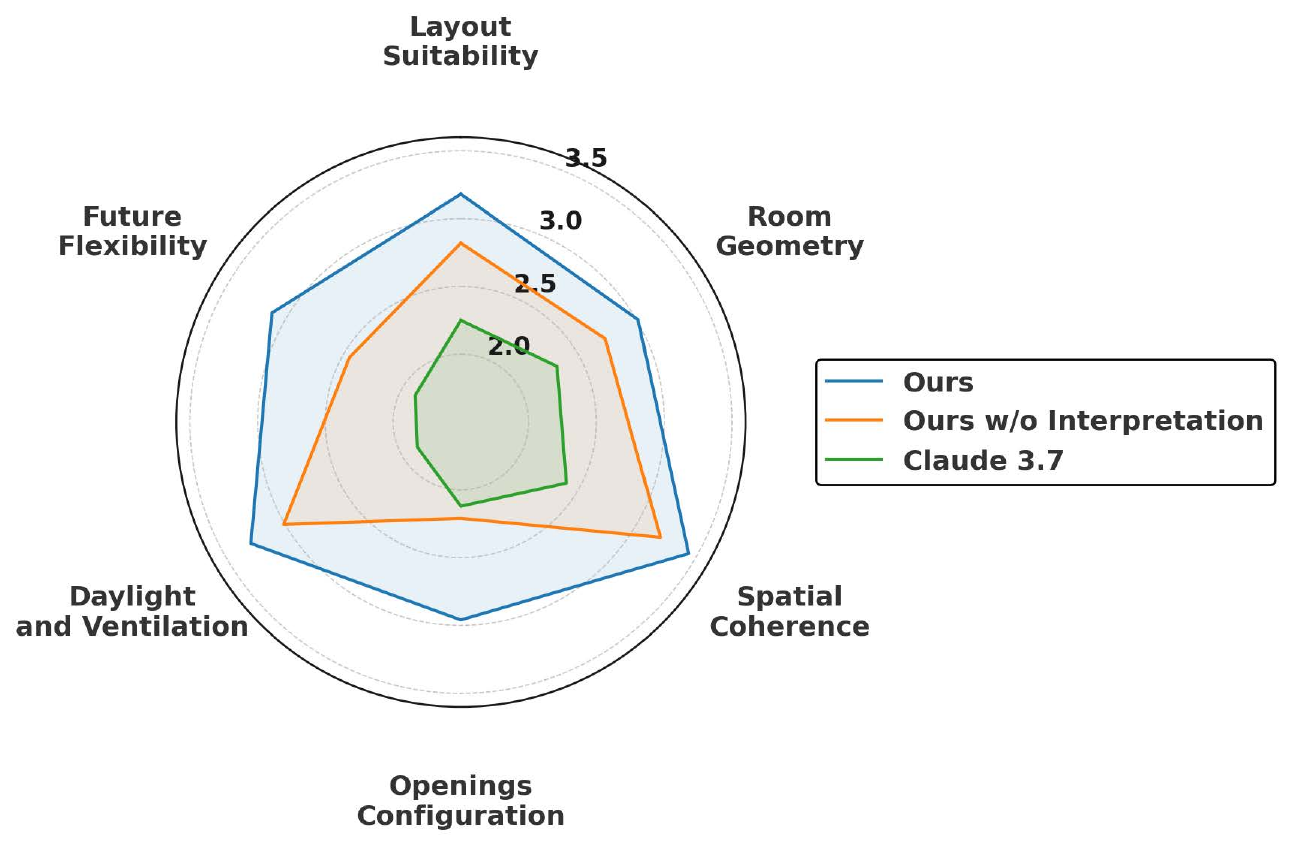}
  \caption{Human evaluation of the generated floorplan designs based on six criteria. Lower values indicate that the corresponding requirements were not clearly specified in the design instructions, while higher values reflect better alignment. The detailed evaluation process can be found in Appendix \ref{Criteria}.}
  \label{DESIGN}
\end{figure}

\textbf{Design Evaluation.} We asked human architects to grade the generated floorplans based on six design criteria. 
We then compared our full design layer with two baseline methods: floorplans generated by Claude 3.7 using SVG, and our design layer without the floorplan interpretation module. 
As illustrated in Figure~\ref{DESIGN}, the results show that our method produces the most reasonable and acceptable designs across all six criteria. Notably, it achieves scores above 3 out of 5 in every aspect, outperforming the baseline models and demonstrating a superior ability to handle open-ended design tasks. The detailed evaluation process can be found in Appendix \ref{Criteria}.

\textbf{Operation Evaluation.} As shown in Table~\ref{tab:sroftool}, the BIMgent achieves a 32\% end-to-end success rate on the proposed Mini Building Benchmark. The complexity and length of each task make it difficult to directly assess outcomes, as many steps can influence the final success. To enable a more granular evaluation of agent performance, in addition to 25 end-to-end modeling tasks, we also focused on subtasks involving the modeling of key architectural components. Specifically, we evaluate the creation of design layers (41 tasks), walls (82 tasks), slabs (41 tasks), openings (82 tasks), and roofs (25 tasks) based on whether all required architectural elements are successfully created and whether all parameters are properly configured within each subtask. BIMgent performs particularly well on component-related tasks such as wall and opening creation, with 86.58\% and 92.68\% success rates respectively. This performance is attributable to the availability of floorplan metadata and the relatively simple element creation action patterns that BIMgent can learn effectively. The strongest baseline (Claude 3.7) could not complete any end-to-end modeling task in our benchmark because of the heavy planning and extensive GUI operations that are required. Compared to subtasks, the baseline also performed poorly. BIMgent achieves a 46.34\% success rate on the Layer subtask, 60\% on Roof (both versus 0\% for the baseline), and significantly outperforms the baseline in Wall, Slab, and Openings creation. The results suggest that BIMgent has strong potential to assist in tedious operations by reliably replicating human behavior. Compared to the baseline experiments, our method achieves significantly higher performance, both in overall success rate and across individual modeling parts.

\subsection{Ablation Study}

\textbf{Visual Improvement.} As illustrated in Table~\ref{tab:sroftool}, with the proposed dynamic GUI grounding mechanism, the overall success rate increases by 18.67\%. Delving into the detailed decomposed components, we find that Layer and Roof, which require extensive grounding for name editing and parameter configuration, show significant improvement with 12.19\% and 21.54\%, respectively. This targeted visual attention greatly enhances task efficiency and accuracy.

\textbf{Reflection Ability.} The integration of supervision and self-reflection strategies enables the agent to analyze previously failed tasks and generate improved action plans. Statistically, we observe that the overall task success rate increases by 20.57\% with this mechanism. The primary reason is that Vision-Driven tasks often fail to generate accurate grounding on the first attempt. Through self-reflection, the performance of tasks such as Layer and Roof was improved by 22.13\% and 12\% respectively. In addition, the performance of other related components was slightly improved by approximately 3\%, primarily due to the correction of rare software errors.

\textbf{Hierarchical Planning Is Essential.} Without hierarchical planning to guide the general building modeling steps, the agent struggles to complete even a single task. Other decomposed tasks also perform poorly, particularly the Vision-Driven subtasks such as Layer and Roof. Given the hundreds of substeps involved and the complex usage patterns of the design software, it is challenging for a single LLM to accurately generate the entire sequence of building actions or to fully comprehend the software’s operational logic. The agent exhibits a limited ability to create elements. This demonstrates its learning potential under documentation guidance, though its overall performance remains low.

\subsection{Error Analysis }
\label{erroranaly}

\begin{figure}[h!]
  \centering
  \includegraphics[width=\columnwidth]{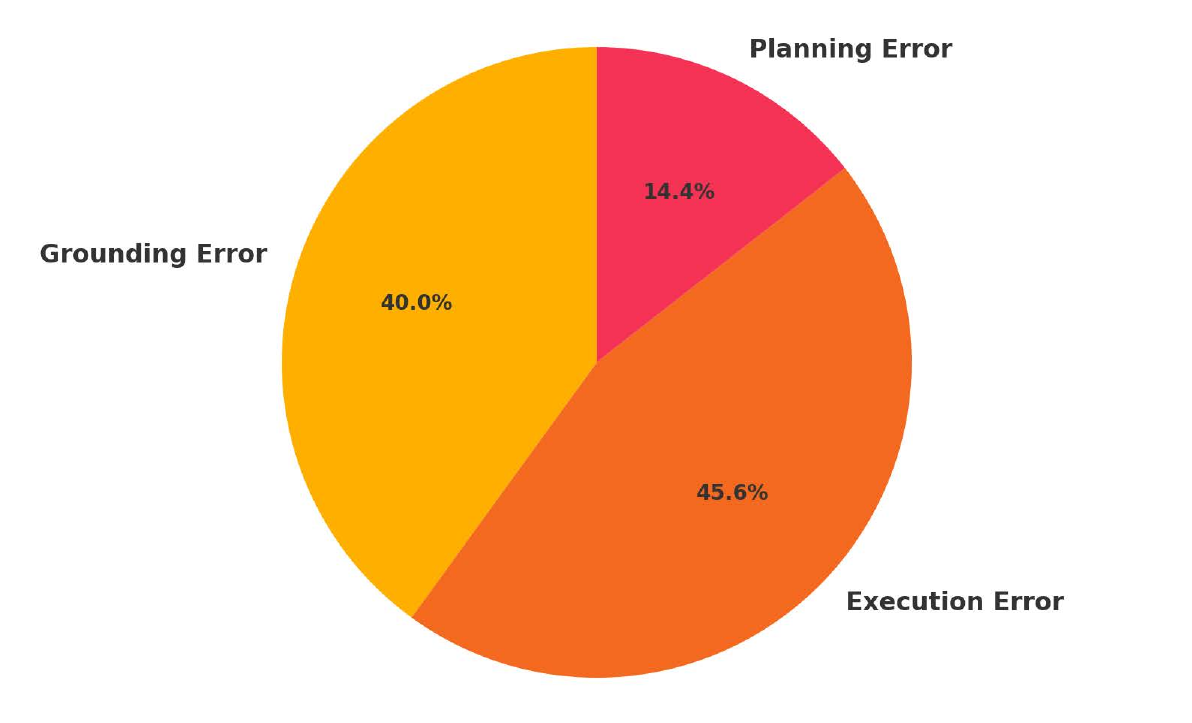}
   \caption{Distribution of the 92 errors across three types: planning errors, grounding errors, and execution errors.}
    
  \label{errorrate}
\end{figure}

To better visualize BIMgent’s performance, we draw inspiration from Agent S \cite{agashe2024agent} and conduct an error rate analysis across all 25 modeling tasks. We trace the BIMgent’s trajectory throughout the tasks and analyze all 92 erroneous action steps, categorizing them into three types: (1) Planning Errors -- incorrect plans that do not align with the current task; (2) Grounding Errors -- failures in accurately identifying or parsing the intended GUI targets; and (3) Execution Errors -- incorrect or unintended actions during operation. As shown in Figure~\ref{errorrate}, our findings reveal that, due to the complexity of the BIM authoring software's GUI and its intricate usage patterns, the most error-prone aspects are those associated with grounding and execution, which account for 40.0\% and 45.6\% of the total errors, respectively. Detailed visualizations of the three types of errors are presented in Appendix~\ref{trajectories}.

\subsection{Qualitative Analysis}
\label{qualitativeana}

In the qualitative analysis, we present a successful example from our benchmark. Given a hand-drawn floorplan and input prompt: 

\textit{Generate a building model based on a hand-drawn octagon floorplan, modifying the interior layout to include four rooms instead of three.} 

\begin{figure}[htbp]
  \centering
  %------------ First image ------------%
  \begin{subfigure}[b]{0.24\textwidth}
    \includegraphics[width=\linewidth]{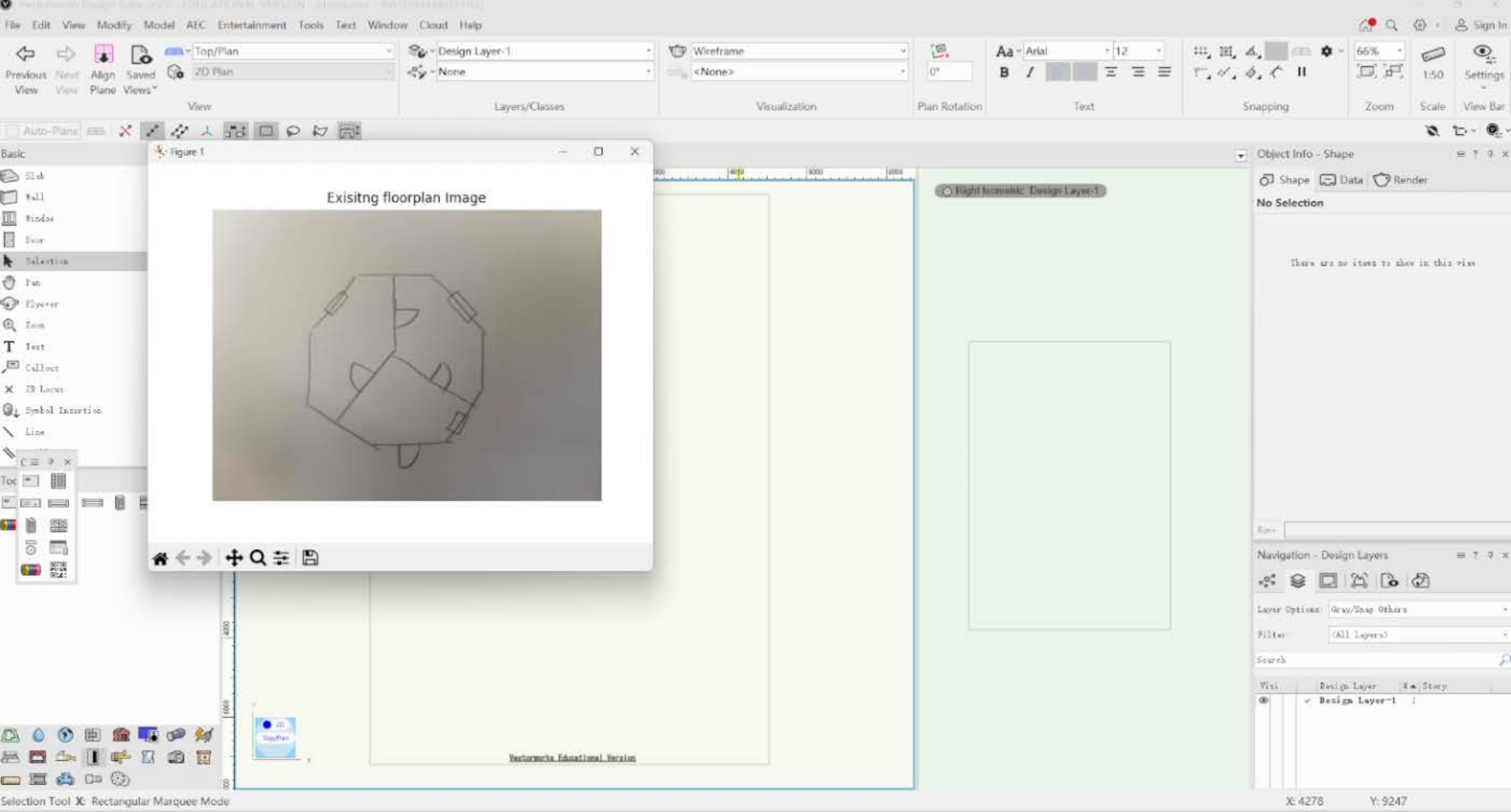}
    \caption{Existing Sketch}
  \end{subfigure}\hfill
  %------------ Second image -----------%
  \begin{subfigure}[b]{0.24\textwidth}
    \includegraphics[width=\linewidth]{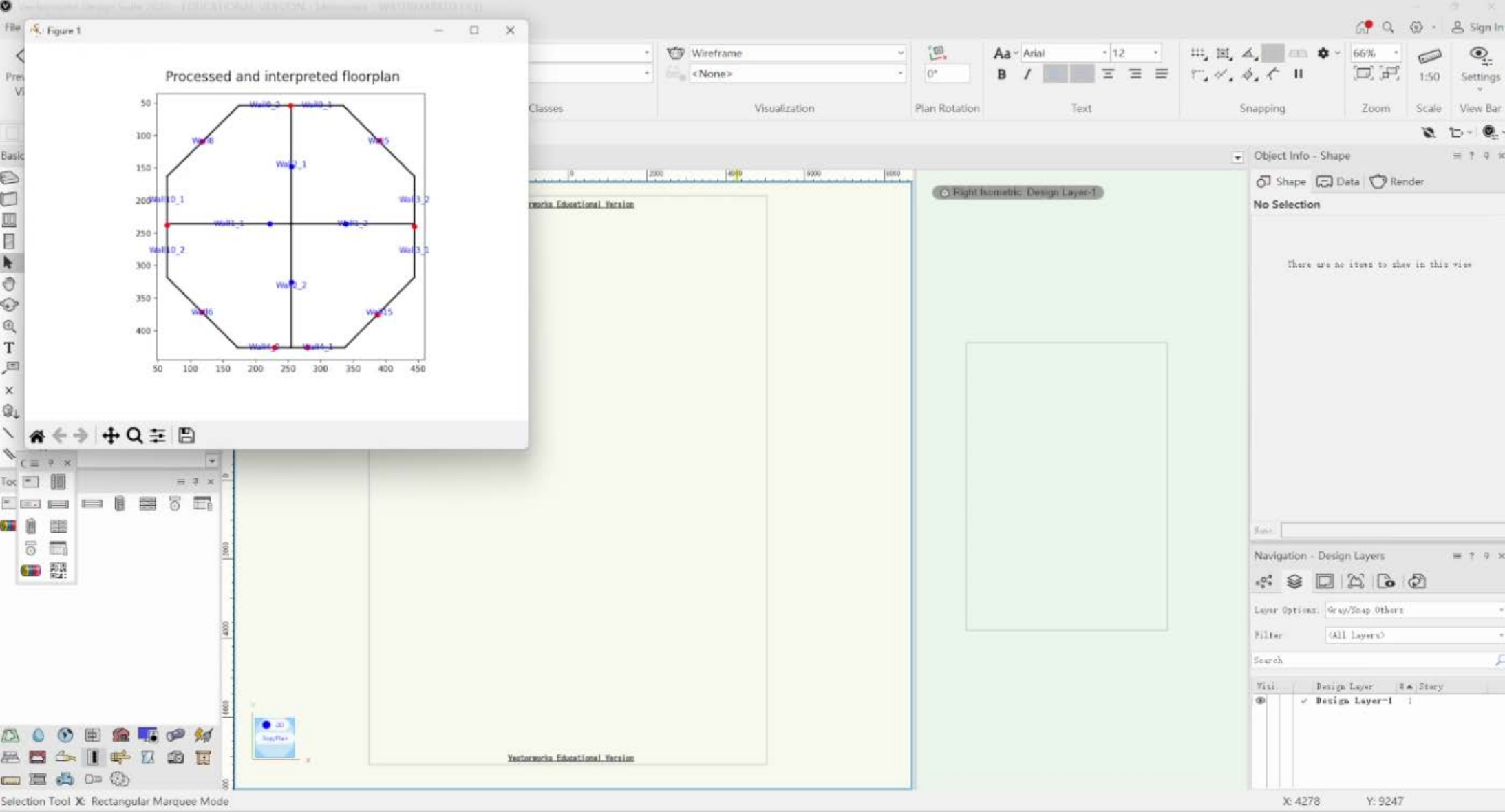}
    \caption{Redesign}
  \end{subfigure}\hfill
  %------------ Third image ------------%
  \begin{subfigure}[b]{0.24\textwidth}
    \includegraphics[width=\linewidth]{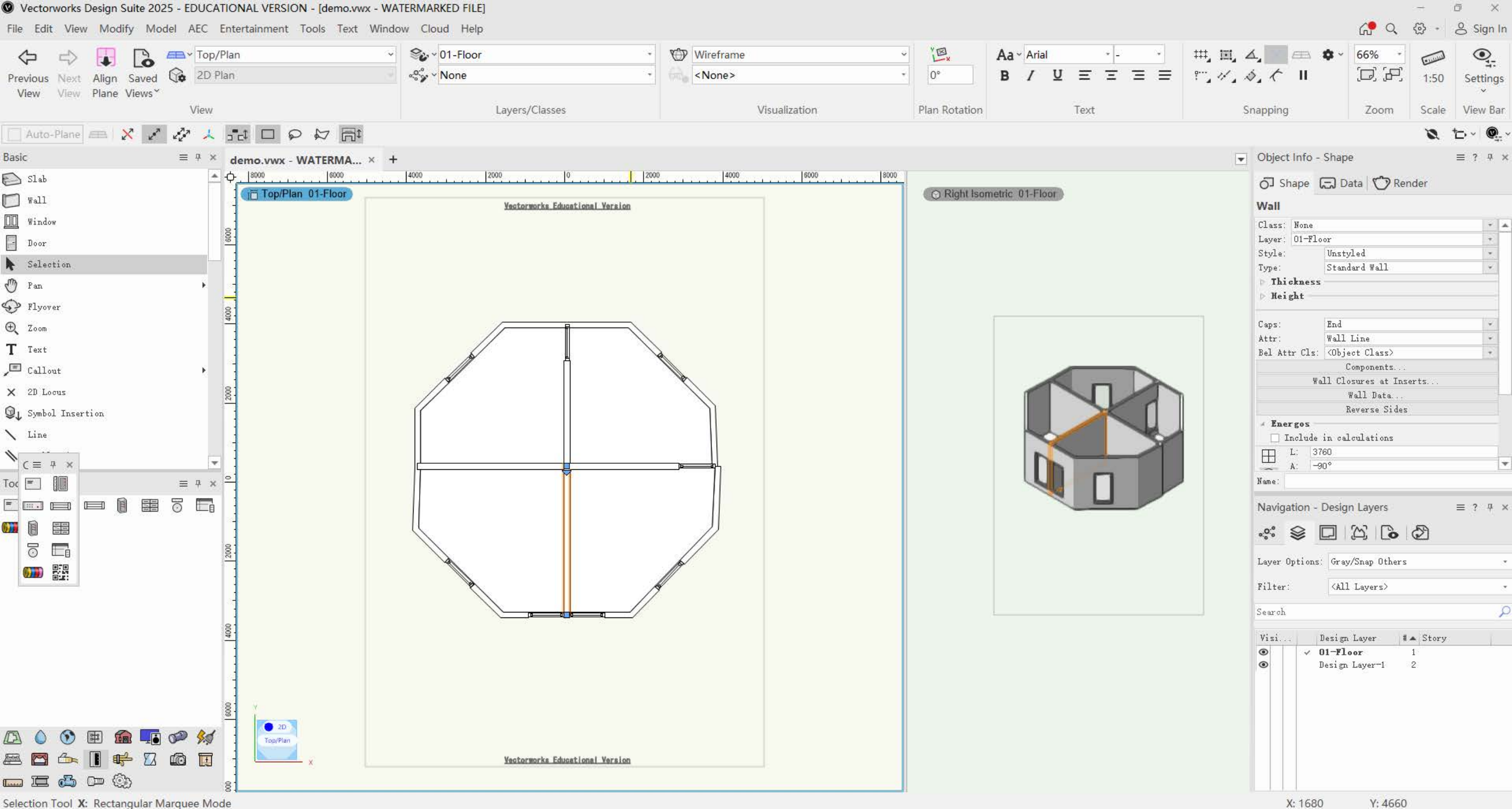}
    \caption{Wall creation}
  \end{subfigure}\hfill
  %------------ Fourth image -----------%
  \begin{subfigure}[b]{0.24\textwidth}
    \includegraphics[width=\linewidth]{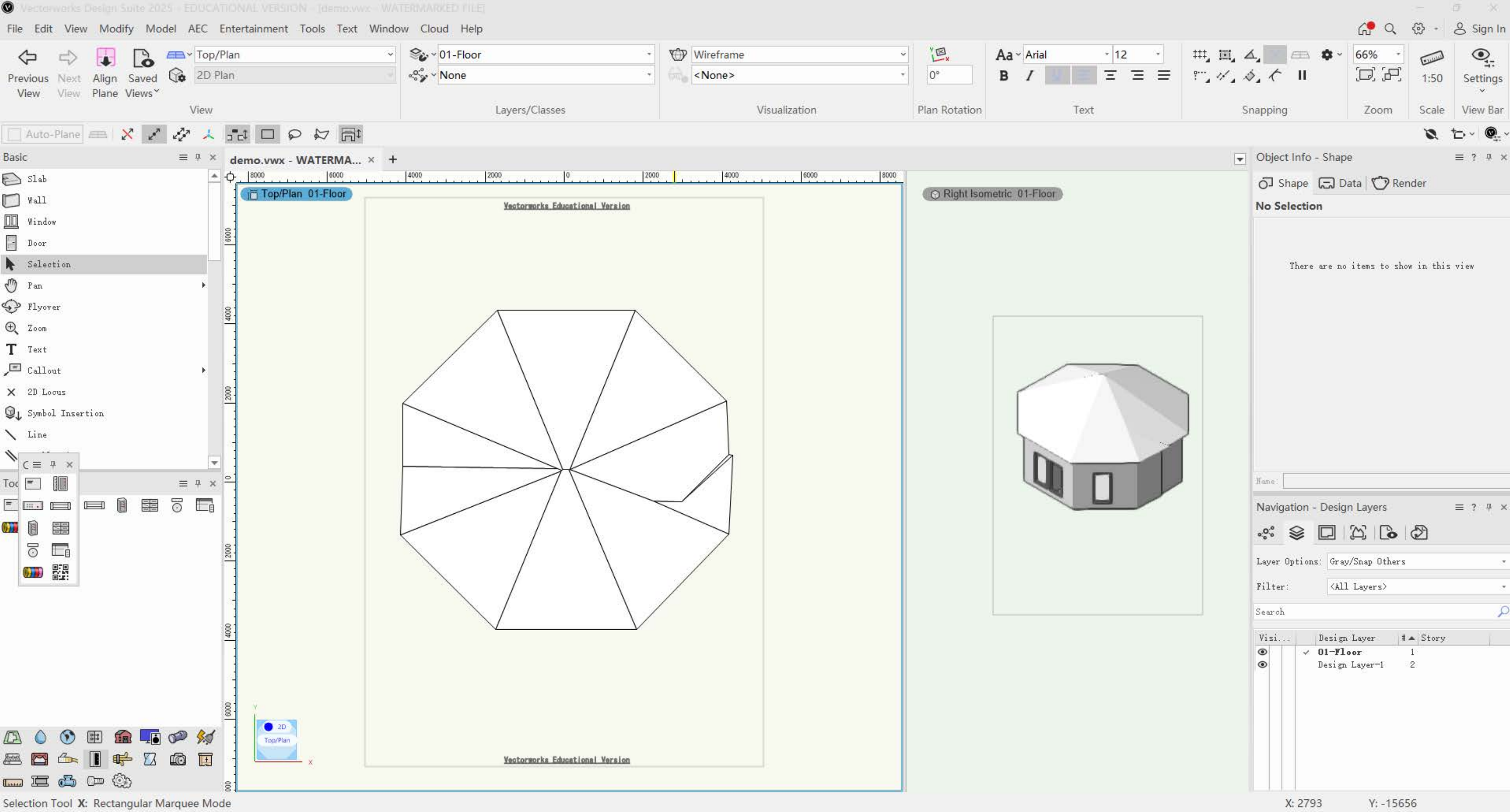}
    \caption{Generated 3D building model}
  \end{subfigure}

  \caption{Example: \textit{Generate a building model based on a hand-drawn octagon floorplan, modifying the interior layout to include four rooms instead of three.} (a)–(d) show the input, redesigned floorplan, action sequence for wall creation, and final 3D model, respectively. More examples can be found in Appendix \ref{trajectories} and \ref{app:visual}. }
  \label{fig:DESIGN}
\end{figure}

The creation of a 3D building model is illustrated in Figure~\ref{fig:DESIGN}. The process begins with the system ingesting both the textual description and the hand-drawn image, which it interprets, regenerates, segments, and processes to produce a 2D floorplan. This floorplan is then converted into a structured representation that downstream agents can parse and act upon. In the second stage, the agents use this structured data to sequentially generate actions and construct the model within the BIM software. Finally, a complete 3D building model is generated through automated computer control. More qualitative examples are in Appendix \ref{trajectories}.

\section{Conclusion}

In this paper, we introduce \textbf{BIMgent}, an agentic framework powered by multimodal LLMs, capable of autonomously generating building models within BIM authoring software through GUI operations. This paper makes the following key contributions: 

(1) Compared to existing GUI agents, \textbf{BIMgent} demonstrates the ability to handle open-ended design tasks, bridging the gap in applying GUI agents to professional design software.

(2) \textbf{BIMgent} addresses the limitations of domain-specific task handling by integrating software documentation into the planning process through a retrieval-augmented strategy.

(3) \textbf{BIMgent} significantly boosts performance by combining dynamic GUI grounding, reflective feedback, and hierarchical planning, thereby overcoming the challenges of complex GUI and the hundreds of steps required for building modeling tasks.

(4) By evaluating two stages of the framework, we show that \textbf{BIMgent} can complete the entire modeling process autonomously, particularly excelling in the most labor-intensive parts of the modeling process.

In the future, one primary direction is to extend the framework to other design software, exploring its potential for generalization across platforms. Another key challenge lies in optimizing the agent’s step count and execution time. Since our goal is to reduce the substantial manual effort involved in building model authoring, improving efficiency is crucial. Additionally, while we currently rely on existing pre-trained models, fine-tuning an open-source model could further enhance adaptability and performance. Finally, to address limitations in evaluation, we plan to develop a more automated evaluation method and introduce a dedicated benchmark for more consistent and scalable assessment.

\section{Acknowledgment}
This work is funded by Nemetschek Group, which is gratefully acknowledged. We sincerely appreciate the licensing support provided by Vectorworks, Inc.

\bibliography{example_paper}
\bibliographystyle{icml2025}

%%%%%%%%%%%%%%%%%%%%%%%%%%%%%%%%%%%%%%%%%%%%%%%%%%%%%%%%%%%%%%%%%%%%%%%%%%%%%%%
%%%%%%%%%%%%%%%%%%%%%%%%%%%%%%%%%%%%%%%%%%%%%%%%%%%%%%%%%%%%%%%%%%%%%%%%%%%%%%%
% APPENDIX
%%%%%%%%%%%%%%%%%%%%%%%%%%%%%%%%%%%%%%%%%%%%%%%%%%%%%%%%%%%%%%%%%%%%%%%%%%%%%%%
%%%%%%%%%%%%%%%%%%%%%%%%%%%%%%%%%%%%%%%%%%%%%%%%%%%%%%%%%%%%%%%%%%%%%%%%%%%%%%%
\newpage
\appendix
\onecolumn

\section{BIMgent Action Definitions}
\label{app:action_def}

We pre-define multiple low-level GUI actions. The action generator and low-level planner responsible for producing actions use these action definitions as references during execution. The detailed information is shown in Table~\ref{tab:actions}.

\begin{table}[H] % Changed from [t] to [H]
\caption{Defined low-level actions including their parameters and descriptions.}
\label{tab:actions}
\vskip 0.15in
\begin{center}
\begin{small}
\begin{sc}
\begin{tabular}{l c p{6cm}} % Set width of third column to enable line wrapping
\toprule
\textbf{Action} & \textbf{Parameter(s)} & \textbf{Description} \\
\midrule
\texttt{move\_mouse\_to(x: int, y: int)} & Pixel coordinates & Move the mouse cursor to the specified screen position. \\
\texttt{left\_click()} & -- & Perform a left-click using the mouse. \\
\texttt{type\_name(name: str)} & Name string & Type the given name using the keyboard. \\
\texttt{press\_escape()} & -- & Press the Escape key on the keyboard. \\
\texttt{press\_enter()} & -- & Press the Enter key on the keyboard. \\
\texttt{shortcut(combo: str)} & Key combination string & Execute a keyboard shortcut by pressing the specified key combination. \\
\texttt{select\_all()} & -- & Select all current components. \\
\bottomrule
\end{tabular}
\end{sc}
\end{small}
\end{center}
\vskip -0.1in
\end{table}

\section{Mini Building Benchmark}
\label{bench}
\subsection{Introduction}
\begin{figure}[h!]
  \centering
  \includegraphics[width=\columnwidth]{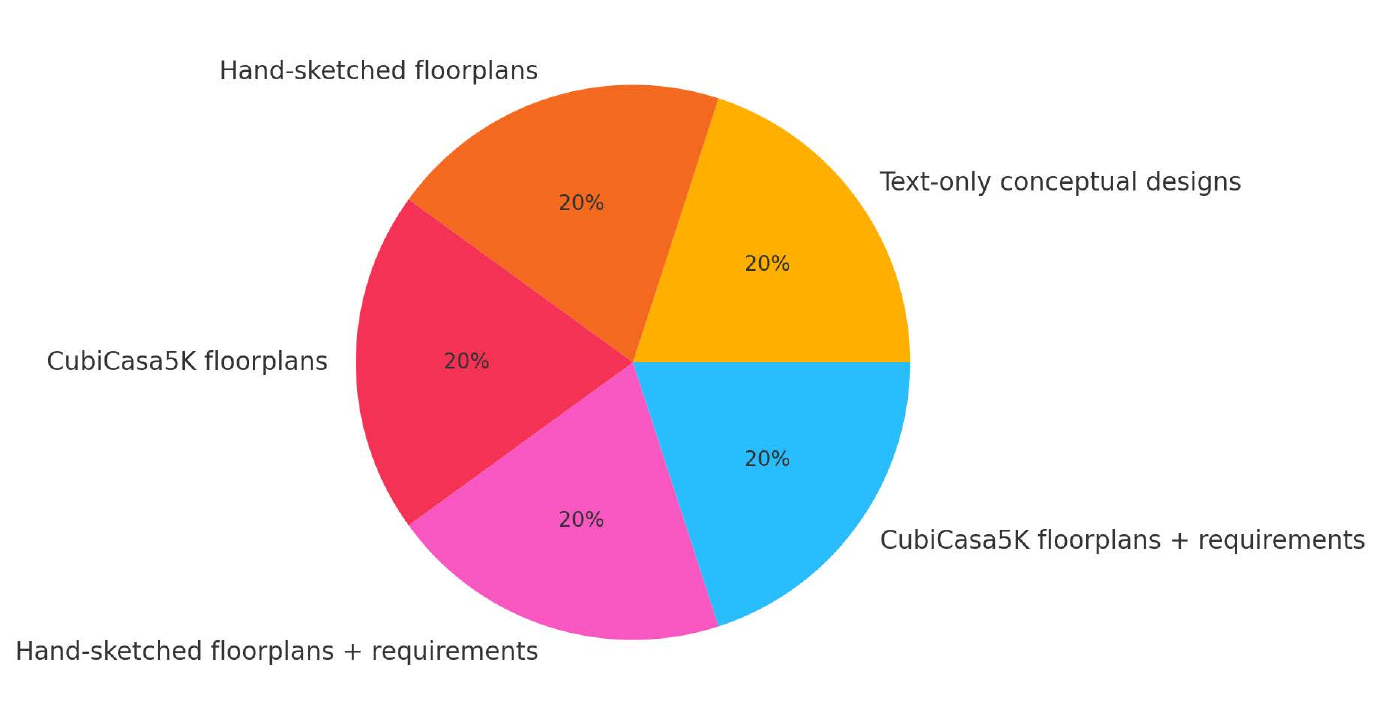}
   \caption{Task Distribution of the Mini Building Benchmark}
    
  \label{dis_builidngbench}
\end{figure}

We present a Mini Building Benchmark consisting of 25 real-world BIM modeling tasks. These tasks cover five input scenarios. Firstly, five text-only conceptual designs that describe key attributes such as the building type, number and types of rooms, and the number of floors (ranging from one to three storeys). Secondly, five hand-drawn sketch floorplan images depict both regular and irregular shapes, along with the number of floors. Thirdly, five randomly unmodified floorplans were selected from the CubiCasa5K dataset \cite{kalervo2019cubicasa5k}, which contains 5000 real estate floorplan images. Fourthly, the same five hand-drawn sketch floorplans were reused with additional explicit modification requirements (e.g., ‘add a room’). Fifthly, the same five CubiCasa5K floorplans with similar modification instructions. Figure~\ref{dis_builidngbench} illustrates the distribution.

\subsection{Tasks}
In this section, we present the detailed task contents, as shown in Table~\ref{tab:text} and Table~\ref{tab:images}.

\begin{longtable}{l p{10.5cm}}
\caption{Detailed contents of the tasks 1-5 from the Mini Building Benchmark} \label{tab:text} \\
\toprule
\parbox[c][1cm][c]{4cm}{\centering \textbf{Task Index}} & 
\parbox[c][1cm][c]{10.5cm}{\centering \textbf{Task content}} \\

\midrule
\endfirsthead

\toprule
\parbox[c][1cm][c]{4cm}{\centering \textbf{Task Index}} & 
\parbox[c][1cm][c]{10.5cm}{\centering \textbf{Task content}} \\

\midrule
\endhead

\parbox[c][2cm][c]{4cm}{\centering Task 1} & \parbox[c][2cm][c]{10.5cm}{\centering Generate a one-storey office building with a large open workspace occupying most of the floor area. The layout should also include two enclosed meeting rooms, a manager’s office, a small pantry, and two restrooms.} \\
\midrule

\parbox[c][2cm][c]{4cm}{\centering Task 2} & \parbox[c][2cm][c]{10.5cm}{\centering Generate a one-storey building with a regular hexagonal footprint. Inside the building, create four rooms of roughly equal size. Each room must include doors and one window.} \\
\midrule

\parbox[c][2cm][c]{4cm}{\centering Task 3} & \parbox[c][2cm][c]{10.5cm}{\centering Create a two-storey rectangular residential building designed for a single family. The layout must include a living room, kitchen, bathroom, master bedroom, and one additional bedroom.} \\
\midrule

\parbox[c][2cm][c]{4cm}{\centering Task 4} & \parbox[c][2cm][c]{10.5cm}{\centering Design a two-storey hospital building with eight distinct rooms. These must include a reception area, two consultation rooms, one minor surgery room, one waiting area, two patient rooms, and one staff room.} \\
\midrule

\parbox[c][2cm][c]{4cm}{\centering Task 5} & \parbox[c][2cm][c]{10.5cm}{\centering Create a three-storey commercial office building where each floor has the same layout. Each floor must include two large office rooms, a small meeting room, a restroom, and a central corridor. The entrance to the building is located on the ground floor and leads directly to the corridor.} \\
\bottomrule
\end{longtable}

\begin{longtable}{p{3cm} p{3.5cm} p{8.5cm}} % total = 15cm
\caption{Detailed contents of tasks 6–25 from the Mini Building Benchmark}
\label{tab:images} \\
\toprule
\parbox[t]{3cm}{\centering \textbf{Task Index}} & 
\parbox[t]{3.5cm}{\centering \textbf{Floorplan Image}} & 
\parbox[t]{8.5cm}{\centering \textbf{Task Content}} \\
\midrule
\endfirsthead

\toprule
\parbox[t]{3cm}{\centering \textbf{Task Index}} & 
\parbox[t]{3.5cm}{\centering \textbf{Floorplan Image}} & 
\parbox[t]{8.5cm}{\centering \textbf{Task Content}} \\
\midrule
\endhead

\parbox[t]{3cm}{\centering Task 6 \& 16} &
\raisebox{-0.5\height}{\includegraphics[width=3.3cm]{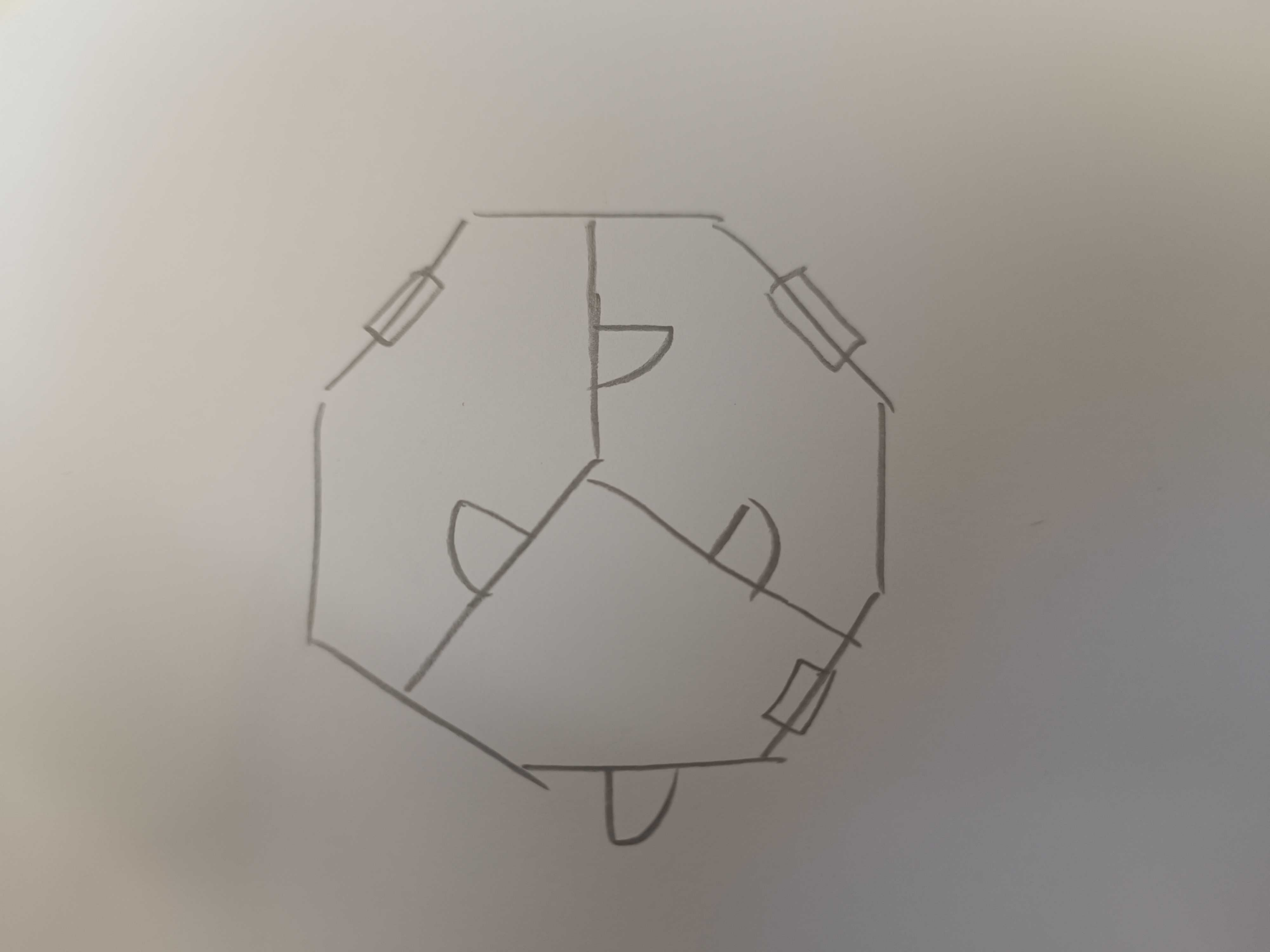}} &
\parbox[c]{8.5cm}{\centering
\vspace{0.3em} % optional spacing
\textbf{Task 6}: Generate a one-storey octagonal building based on the hand-drawn sketch. \newline
\textbf{Task 16}: Generate a building model based on a hand-drawn octagon floorplan, modifying the interior layout to include four rooms instead of three. % optional spacing
} \\
\midrule

\parbox[t]{3cm}{\centering Task 7 \& 17} &
\raisebox{-0.5\height}{\includegraphics[width=3.3cm]{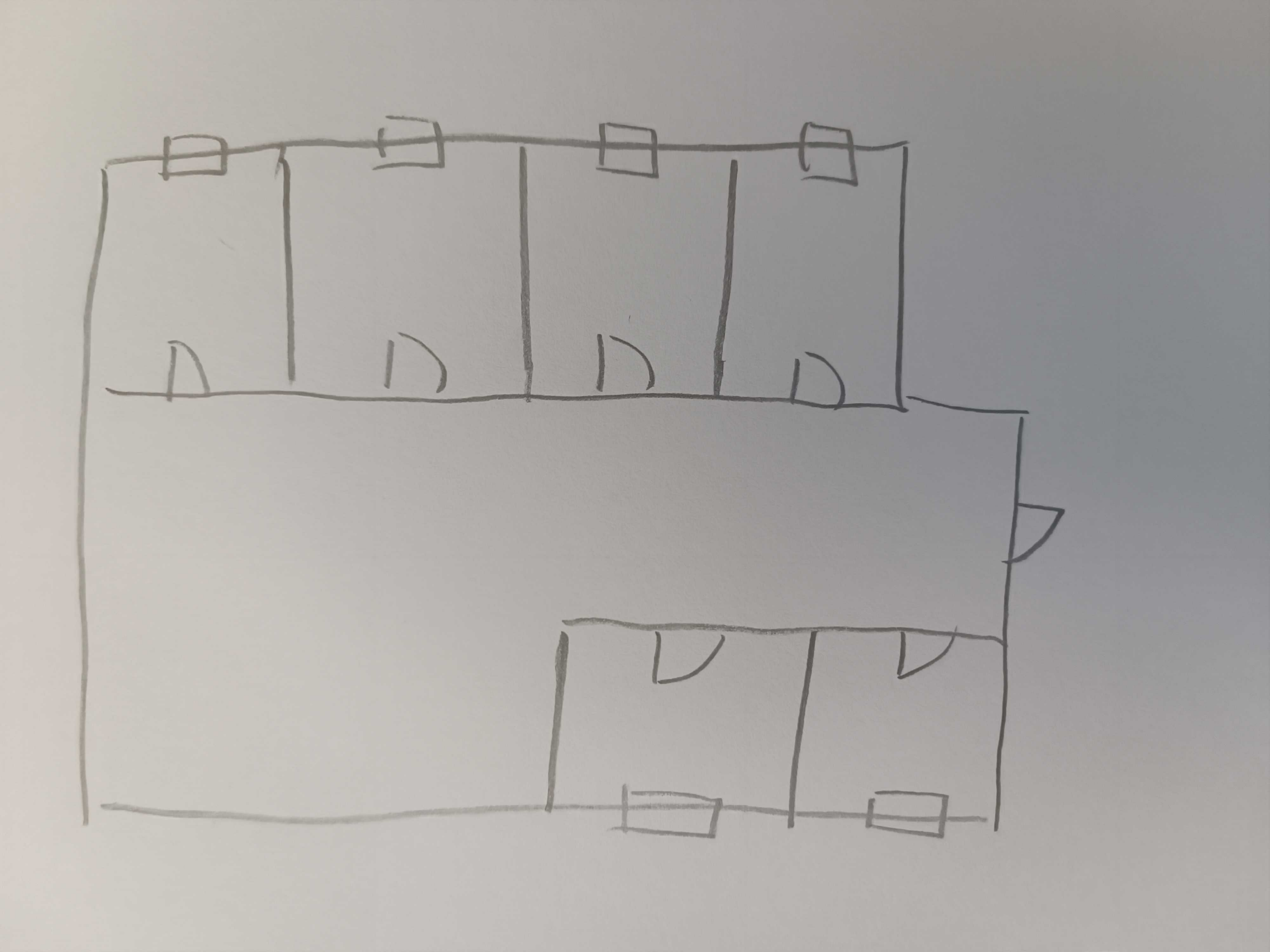}} &
\parbox[c]{8.5cm}{\centering \textbf{Task 7}: Generate a two-storey building based on the sketch. \newline
\textbf{Task 17}: Make a two-floor office building based on the sketch with changes. Split the biggest room in the middle into two rooms.
} \\
\midrule

\parbox[t]{3cm}{\centering Task 8 \& 18} &
\raisebox{-0.5\height}{\includegraphics[width=3.3cm]{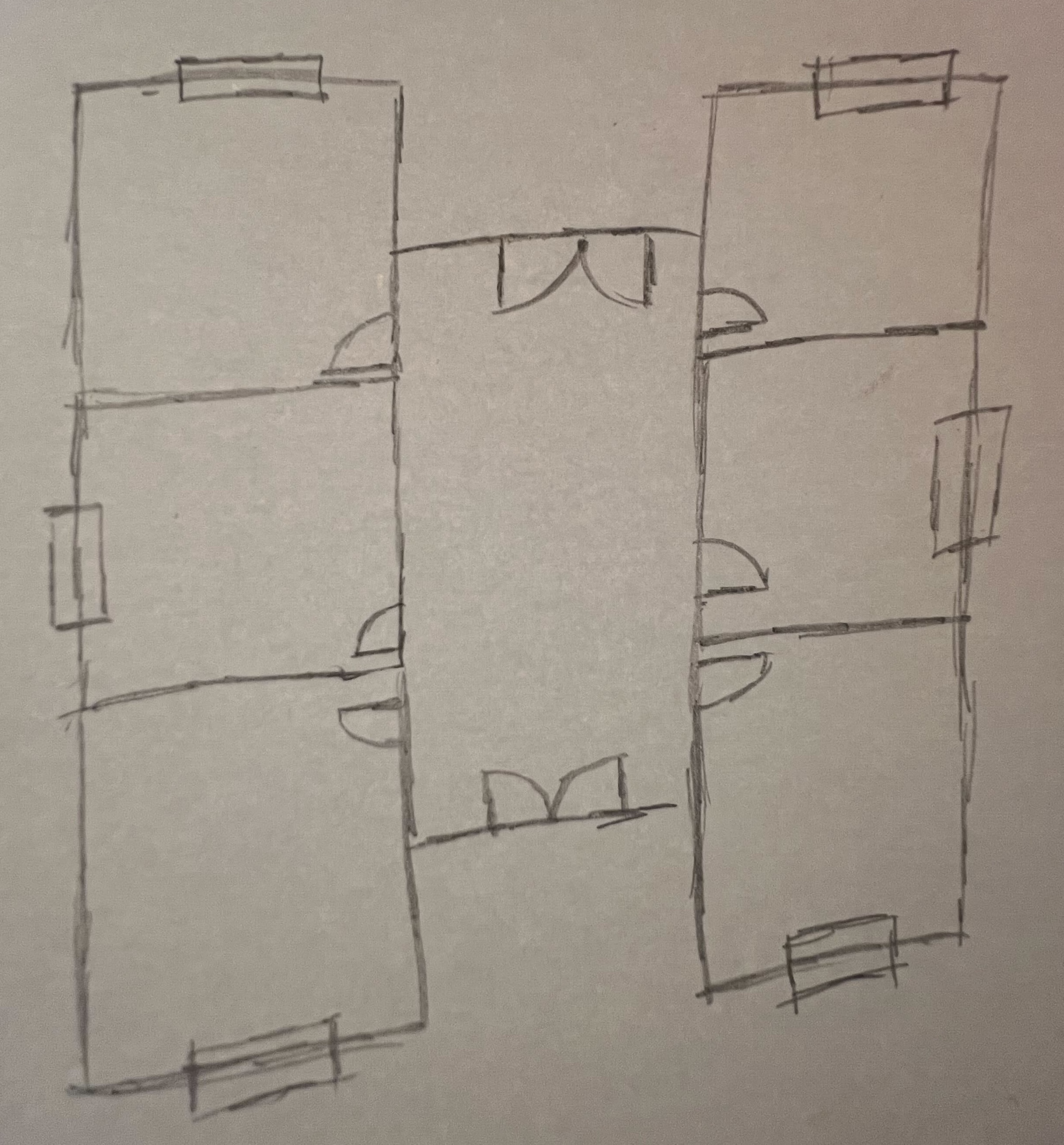}} &
\parbox[c]{8.5cm}{\centering \textbf{Task 8}: Generate a one-storey building based on the sketch. \newline \textbf{Task 18}:Make a one-floor building based on the sketch with updates. The building has an H-shape. Add one extra room to both blocks, so each block has four rooms instead of three.} \\
\midrule

\parbox[t]{3cm}{\centering Task 9 \& 19} &
\raisebox{-0.5\height}{\includegraphics[width=3.3cm]{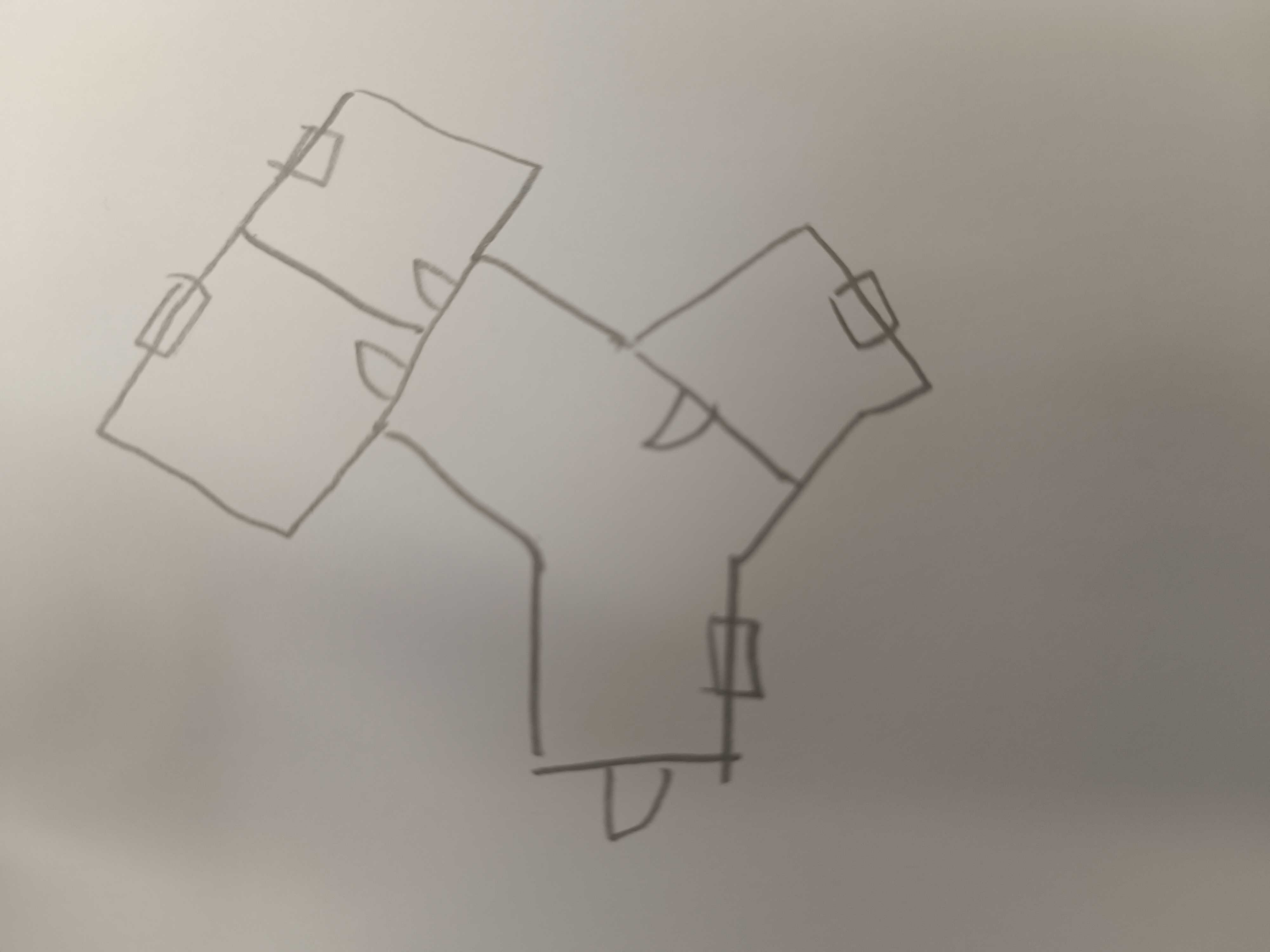}} &
\parbox[c]{8.5cm}{\centering \textbf{Task 9}: Generate a two-storey building based on the sketch. \newline \textbf{Task 19}: Make a two-floor building based on the sketch with updates. Add one more room to the left wing, so the building has five rooms total.} \\
\midrule

\parbox[t]{3cm}{\centering Task 10 \& 20} &
\raisebox{-0.5\height}{\includegraphics[width=3.3cm]{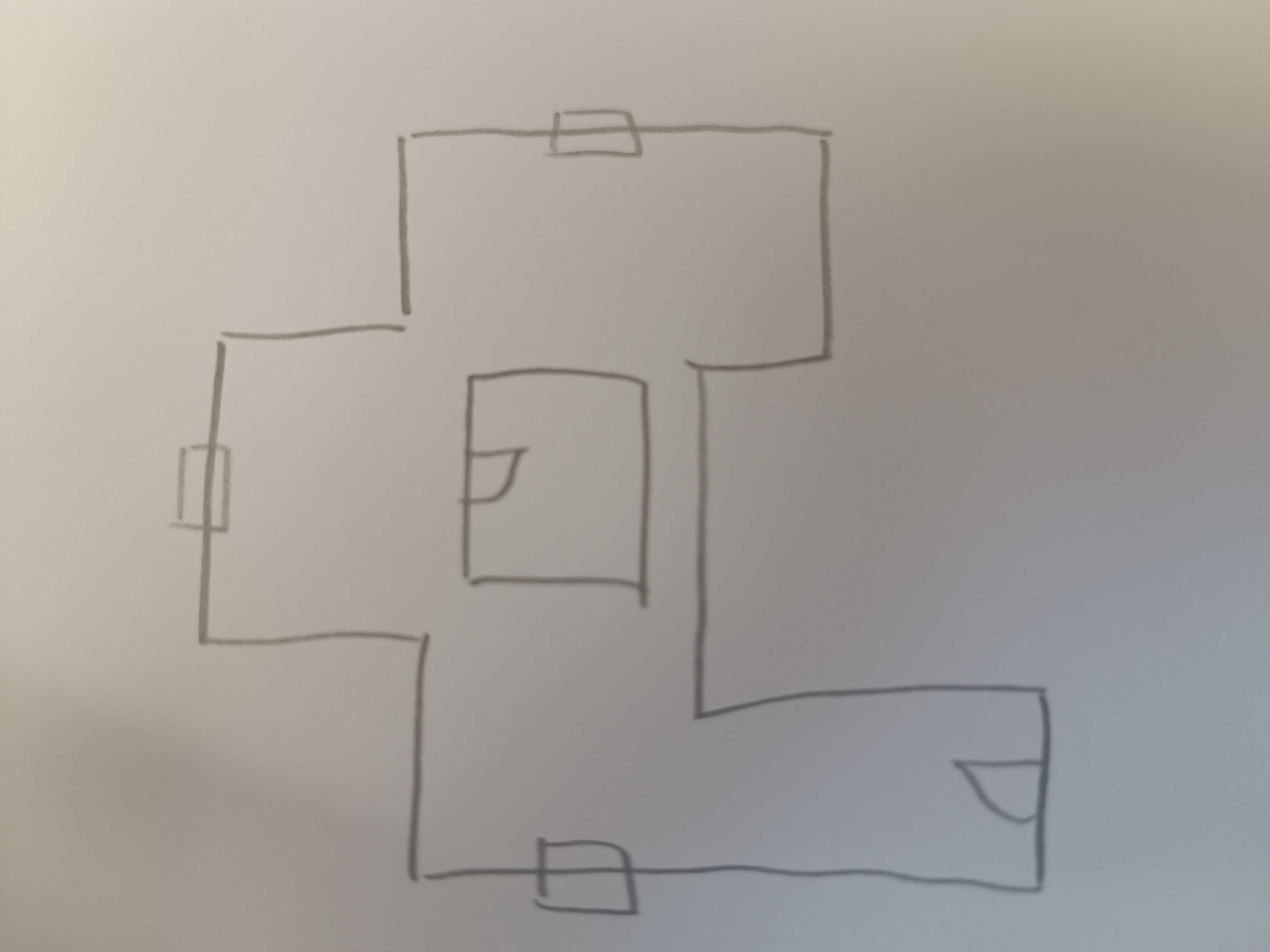}} &
\parbox[c]{8.5cm}{\centering \textbf{Task 10}: Generate a two-storey building based on the sketch. \newline \textbf{Task 20}: Generate a two-storey building based on the sketch, remove the small room in the middle.} \\

\midrule

\parbox[t]{3cm}{\centering Task 11 \& 21} &
\raisebox{-0.5\height}{\includegraphics[width=3.3cm]{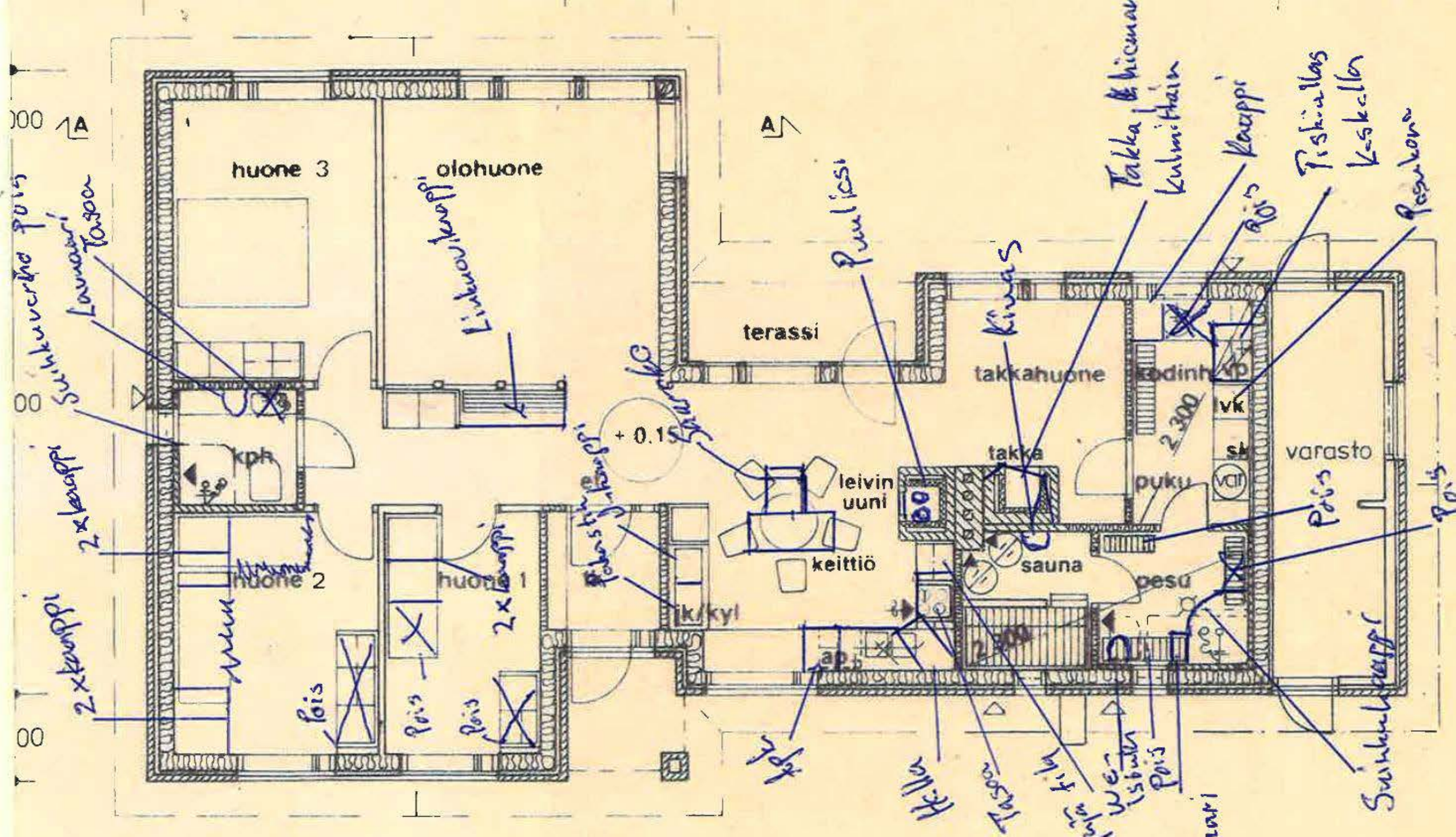}} &
\parbox[c]{8.5cm}{\centering \textbf{Task 11}: Generate a one-storey building based on the image. \newline \textbf{Task 21}: Make a one-floor house based on the floorplan image with updates. Add an extra room next to the top left room.} \\

\midrule

\parbox[t]{3cm}{\centering Task 12 \& 22} &
\raisebox{-0.5\height}{\includegraphics[width=3.3cm]{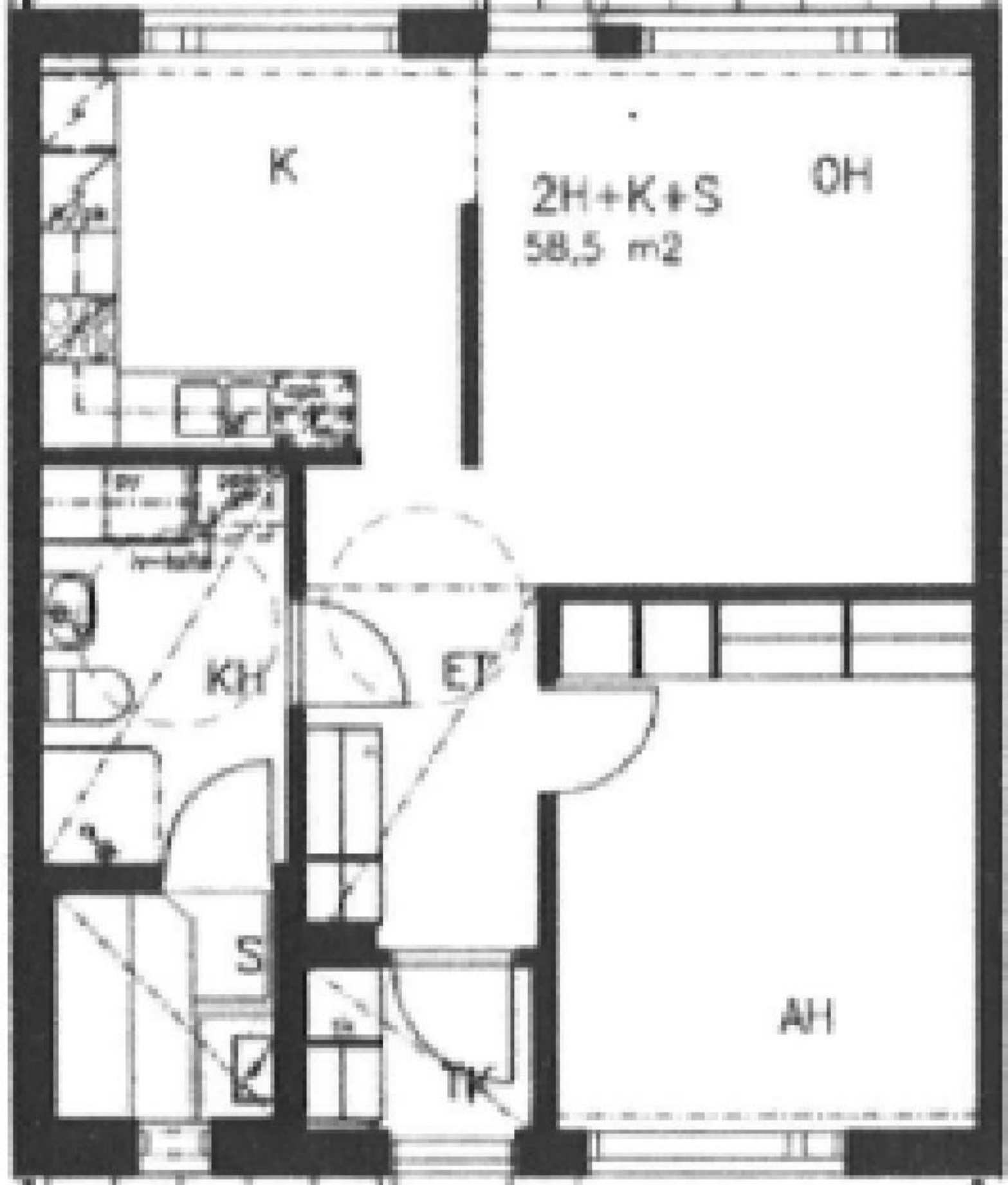}} &
\parbox[c]{8.5cm}{\centering \textbf{Task 12}: Generate a one-floor building based on the image. \newline 
\textbf{Task 22}: Make a one-floor apartment based on the image but with updates. Add an additional room in the bottom-right corner.} \\

\midrule

\parbox[t]{3cm}{\centering Task 13 \& 23} &
\raisebox{-0.5\height}{\includegraphics[width=3.3cm]{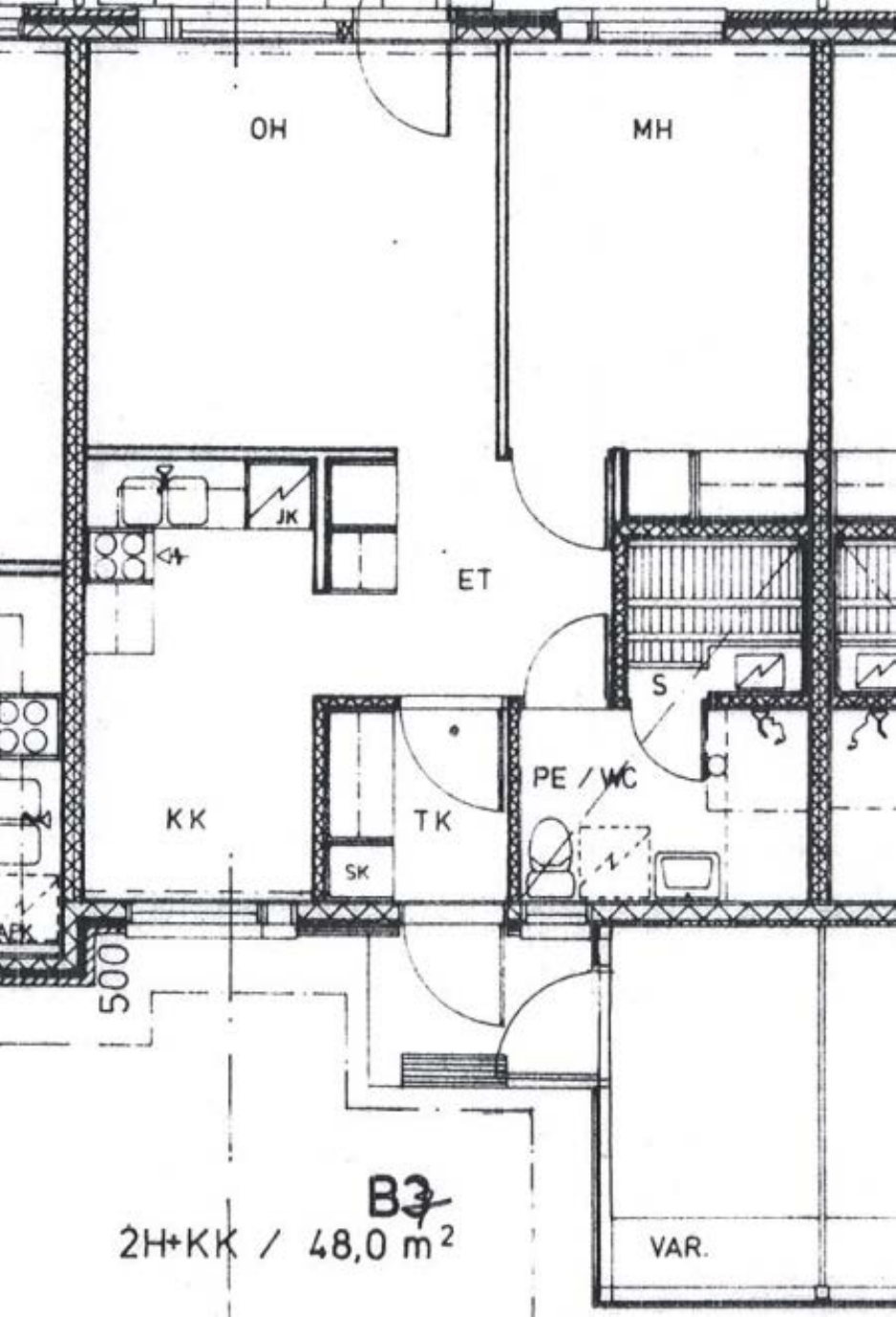}} &
\parbox[c]{8.5cm}{\centering \textbf{Task 13}: Generate a two-floor apartment based on the image. \newline \textbf{Task 23}: Generate a two-floor apartment based on the image with updates. Make the left bottom room smaller and add an additional space next to it.
} \\

\midrule

\parbox[t]{3cm}{\centering Task 14 \& 24} &
\raisebox{-0.5\height}{\includegraphics[width=3.3cm]{images/prompt_files/prompt_019.pdf}} &
\parbox[c]{8.5cm}{\centering \textbf{Task 14}: Generate a two-floor house based on the image. \newline \textbf{Task 24}: Generate a two-floor house based on the floorplan image with updates. Add a small room next to the top right room. 
} \\

\midrule

\parbox[t]{3cm}{\centering Task 15 \& 25} &
\raisebox{-0.5\height}{\includegraphics[width=3.3cm]{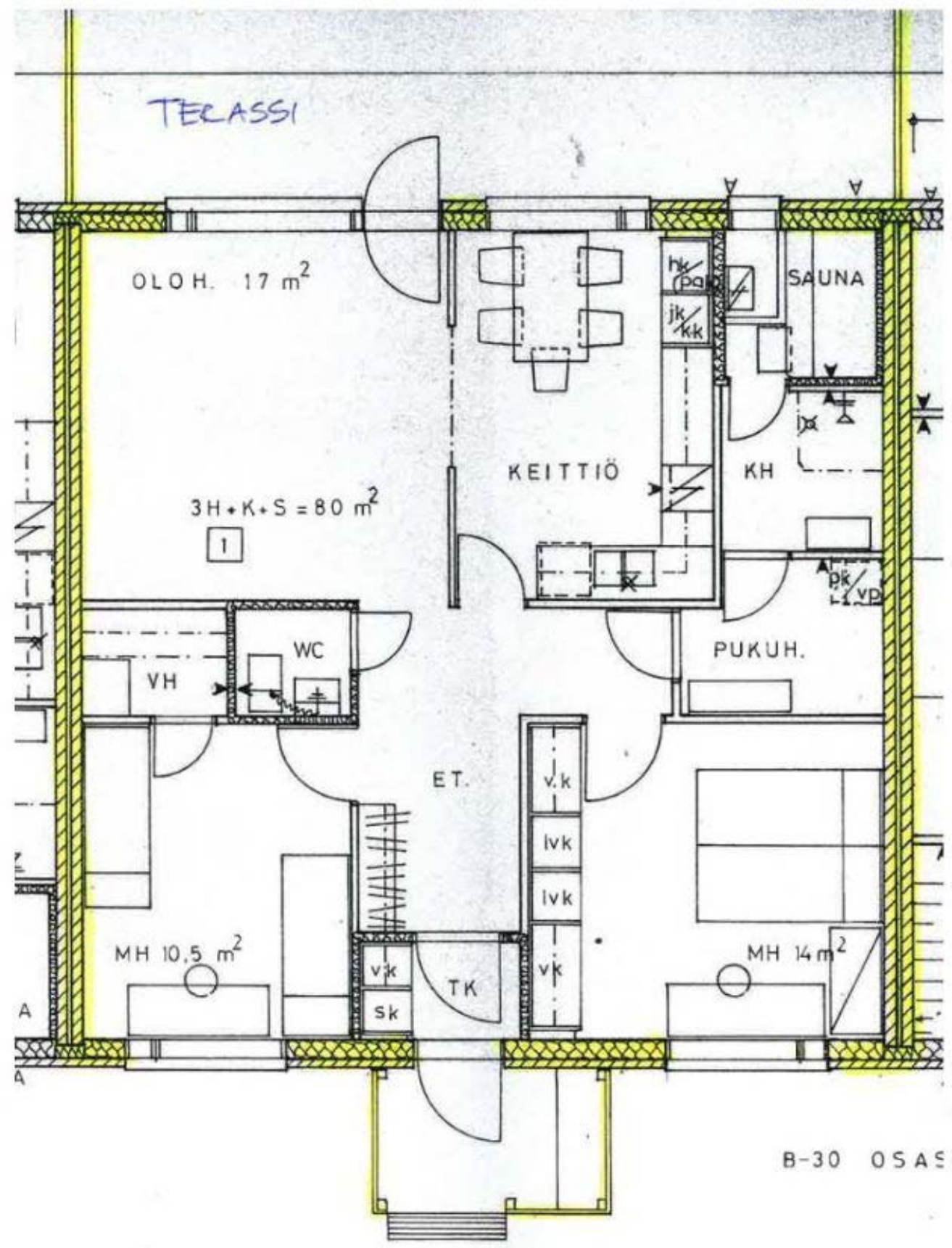}} &
\parbox[c]{8.5cm}{\centering \textbf{Task 15}: Generate a two-floor building based on the image. \newline \textbf{Task 25}: Generate a two-floor house based on the image with some updates. Add a small room in the bottom.
} \\
\bottomrule
\end{longtable}

\section{Evaluation Criteria}
\label{Criteria}
We divided the evaluation process into two phases: design evaluation and operation evaluation. In this section, we introduce the detailed evaluation criteria for these two phases.

\subsection{Design Evaluation}

\textbf{Evaluation Criteria}. Since the design task is open-ended, there is no definitive signal to automatically determine whether the design requirements have been fulfilled. Therefore, we rely on human evaluation at this stage. We define six criteria to assess the quality of the generated floorplans, which are shown in Table \ref{tab:critia}.
\begin{longtable}{p{4cm} p{10.5cm}}
\caption{Detailed descriptions of the criteria for design evaluation} \label{tab:critia} \\
\toprule
\parbox[t]{4cm}{\centering \textbf{Criteria}} & 
\parbox[t]{10.5cm}{\centering \textbf{Description}} \\
\midrule
\endfirsthead

\toprule
\parbox[t]{4cm}{\centering \textbf{Criteria}} & 
\parbox[t]{10.5cm}{\centering \textbf{Description}} \\
\midrule
\endhead

\parbox[t]{4cm}{\centering Layout Suitability} & 
Evaluate whether the overall layout, including the number and configuration of rooms, meets the specified design requirements. \\
\midrule

\parbox[t]{4cm}{\centering Room Geometry} & 
Check for geometric validity of rooms, ensuring there are no isolated, distorted, or nonsensical room shapes. \\
\midrule

\parbox[t]{4cm}{\centering Spatial Coherence} & 
Assess the spatial connectivity between rooms and ensure there are no disconnected or inaccessible areas. \\
\midrule

\parbox[t]{4cm}{\centering Openings Configuration} & 
Verify that openings such as doors and windows are placed correctly on walls and follow typical architectural conventions. \\
\midrule

\parbox[t]{4cm}{\centering Daylight and Ventilation} & 
Confirm that each room has access to openings that enable natural light and ventilation. \\
\midrule

\parbox[t]{4cm}{\centering Future Flexibility} & 
Evaluate whether the floorplan allows for future functional changes, adaptability, or expansions in use. \\
\bottomrule
\end{longtable}

\begin{figure}[htbp]
  \centering
  %------------ First image ------------%
  \begin{subfigure}[b]{0.45\textwidth}
    \centering
    \includegraphics[height=8cm, keepaspectratio]{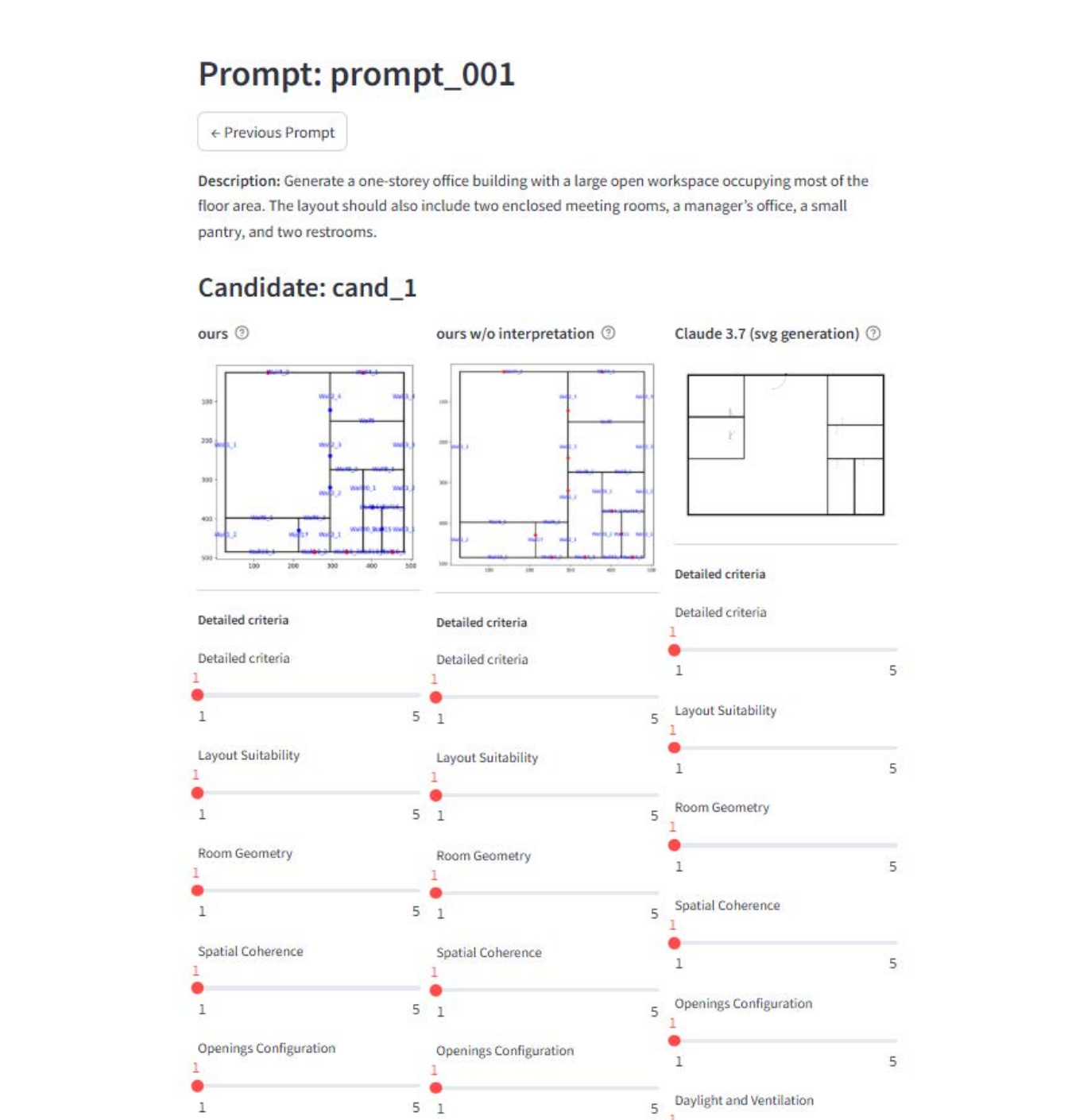}
    \caption{Example survey interface for Task 1}
    \label{fig:survy1png}
  \end{subfigure}
  \hspace{0.05\textwidth}
  %------------ Second image -----------%
  \begin{subfigure}[b]{0.45\textwidth}
    \centering
    \includegraphics[height=8cm, keepaspectratio]{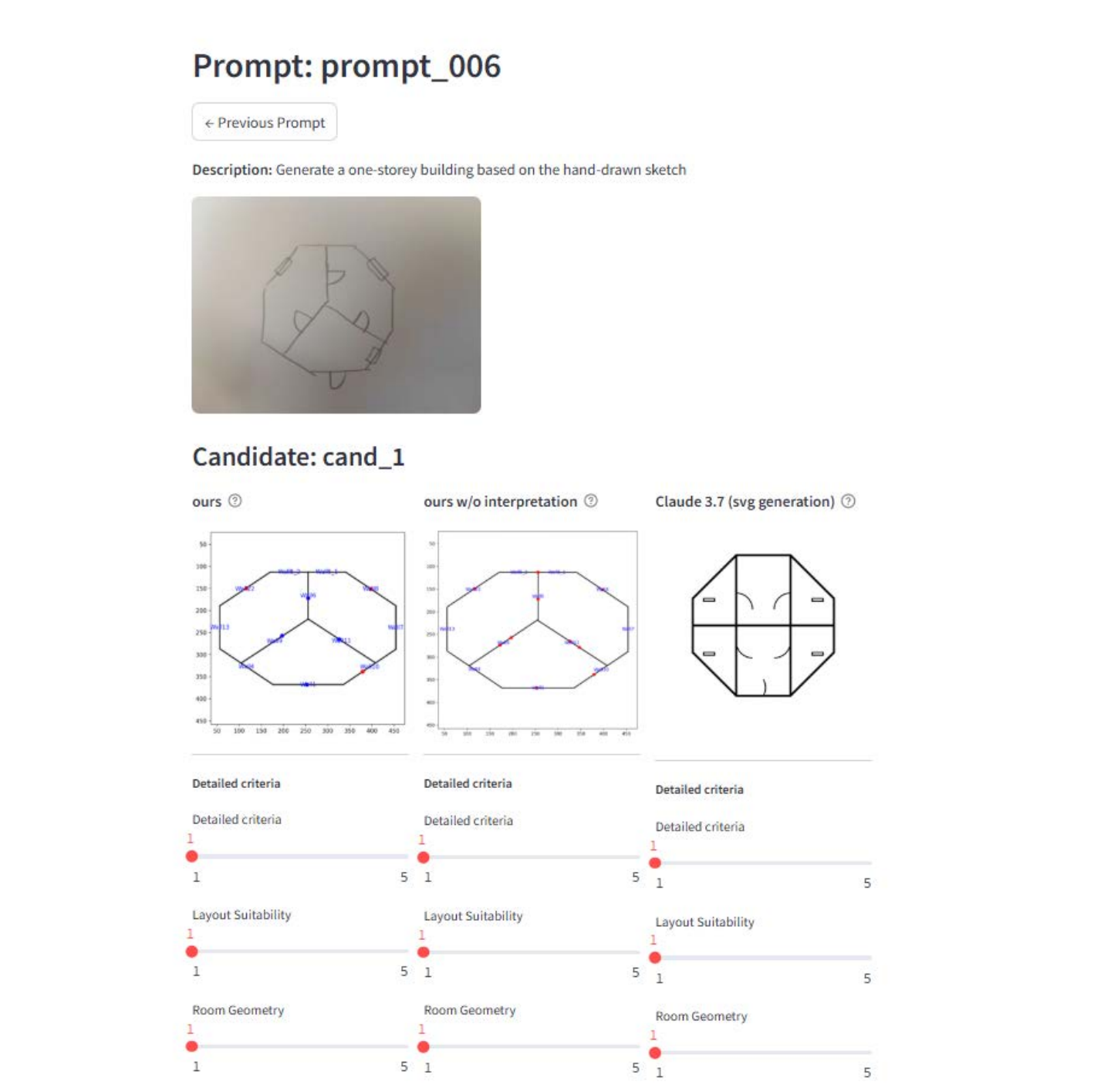}
    \caption{Example survey interface for Task 6}
    \label{fig:design-b}
  \end{subfigure}

  \caption{User interface for the design evaluation survey.}
  \label{fig:survey}
\end{figure}

\textbf{Evaluation Method} We use Claude 3.7 (SVG generation) as the baseline. Additionally, we conduct an ablation study by removing the interpretation part from our design method. To make the evaluation more robust, we expanded the dataset by running the design component of our framework three times for each task in the Mini Building Benchmark, resulting in 75 generated floorplan images per method. 
For each evaluation criterion, a score of 5 represents full satisfaction of the requirement. The closer a score is to 5, the better the design fulfills the criterion, and vice versa. We developed a web-based survey platform where the generated images were presented alongside rating questions based on the defined criteria. We invited architects to participate in the evaluation. A snapshot of the survey interface is shown in Figure~\ref{fig:survey}.

\subsection{Operation Evaluation}

For the operation evaluation, we first selected the best-performing design (based on the design evaluation) and treated it as the ground truth reference. For end-to-end operation evaluation, we exported the final output files and calculated the number of each architecture component (e.g., walls, openings, and slabs). These are compared against the ground truth to identify any missing or incorrect elements. We further conducted a human evaluation on each project file. Experts manually inspected the design layer settings, wall heights, roof configurations, and parameter settings to assess correctness and completeness.

For subtask-level evaluation, we captured screenshots throughout the modeling process, labeling each screenshot according to its corresponding subtask. These screenshots were manually reviewed to identify the specific subtask stage at which any failures occurred. In addition, the final output file was examined to verify whether the component assigned to each subtask was successfully created. If the required component is missing, both the end-to-end task and the corresponding subtask are marked as failed.

\section{BIMgent Execution Trajectories on the Mini Building Benchmark}
\label{trajectories}

We present the execution trajectories of 2 tasks from the Mini Building Benchmark that were successfully completed by BIMgent, providing a supplementary perspective to the qualitative analysis discussed in Section~\ref{qualitativeana}. Additionally, we include failure cases referenced in Section~\ref{erroranaly}, along with corresponding screenshots and sequences of the generated actions. As tasks involve a large number of action steps, the execution process is broken down into detailed segments. 

\subsection{Task 1 Action Examples}
\begin{figure}[htbp]
  \centering

  % 第一行
  \begin{subfigure}[b]{0.32\textwidth}
    \centering
    \includegraphics[width=\linewidth,keepaspectratio]{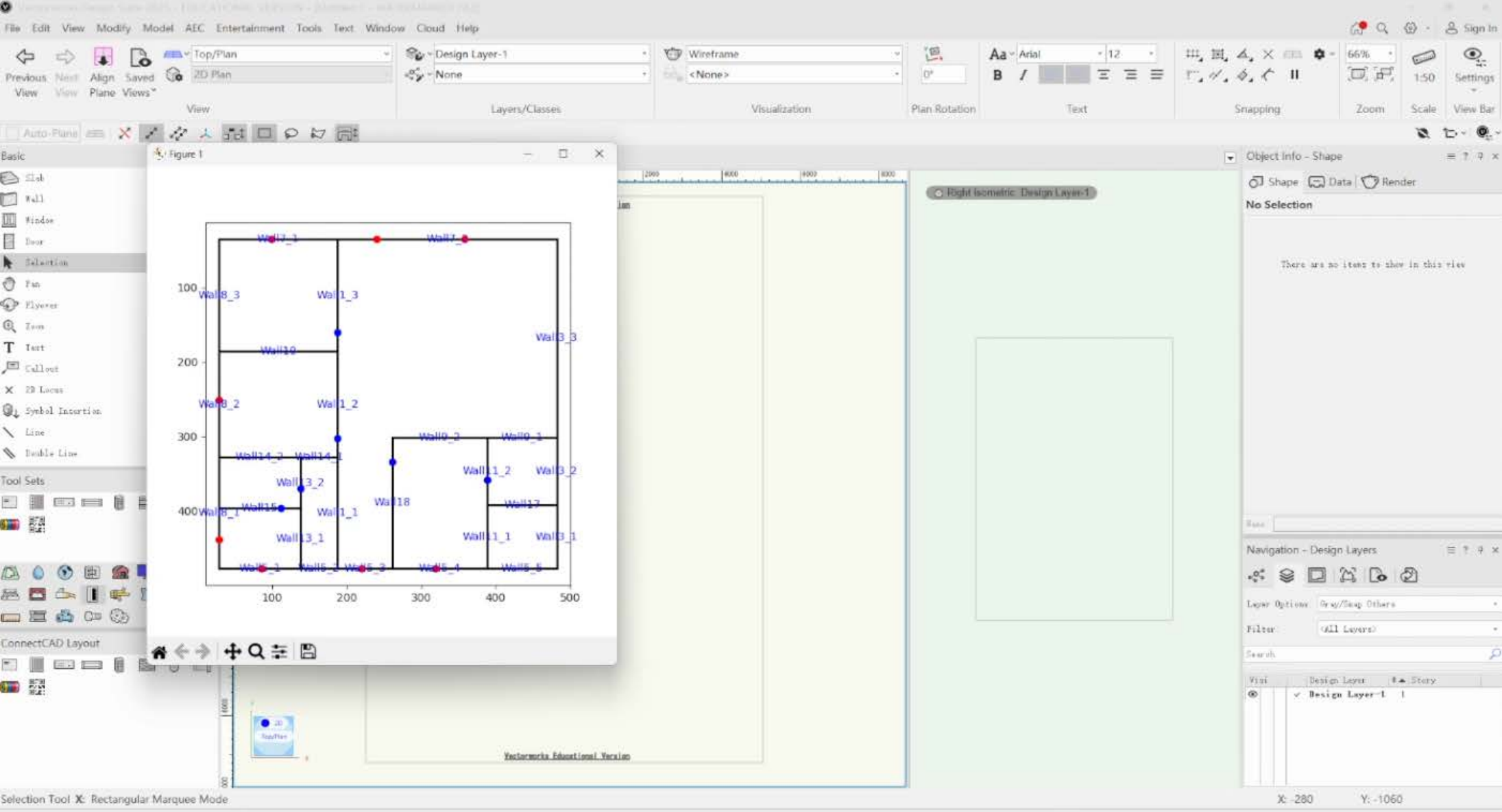}
    \caption{%
      \parbox[c][4\baselineskip][c]{\linewidth}{%
        \centering
        Floorplan design\\
         --\\
          --
      }%
    }
    \label{fig:trajectory-a}
  \end{subfigure}\hfill
  \begin{subfigure}[b]{0.32\textwidth}
    \centering
    \includegraphics[width=\linewidth,keepaspectratio]{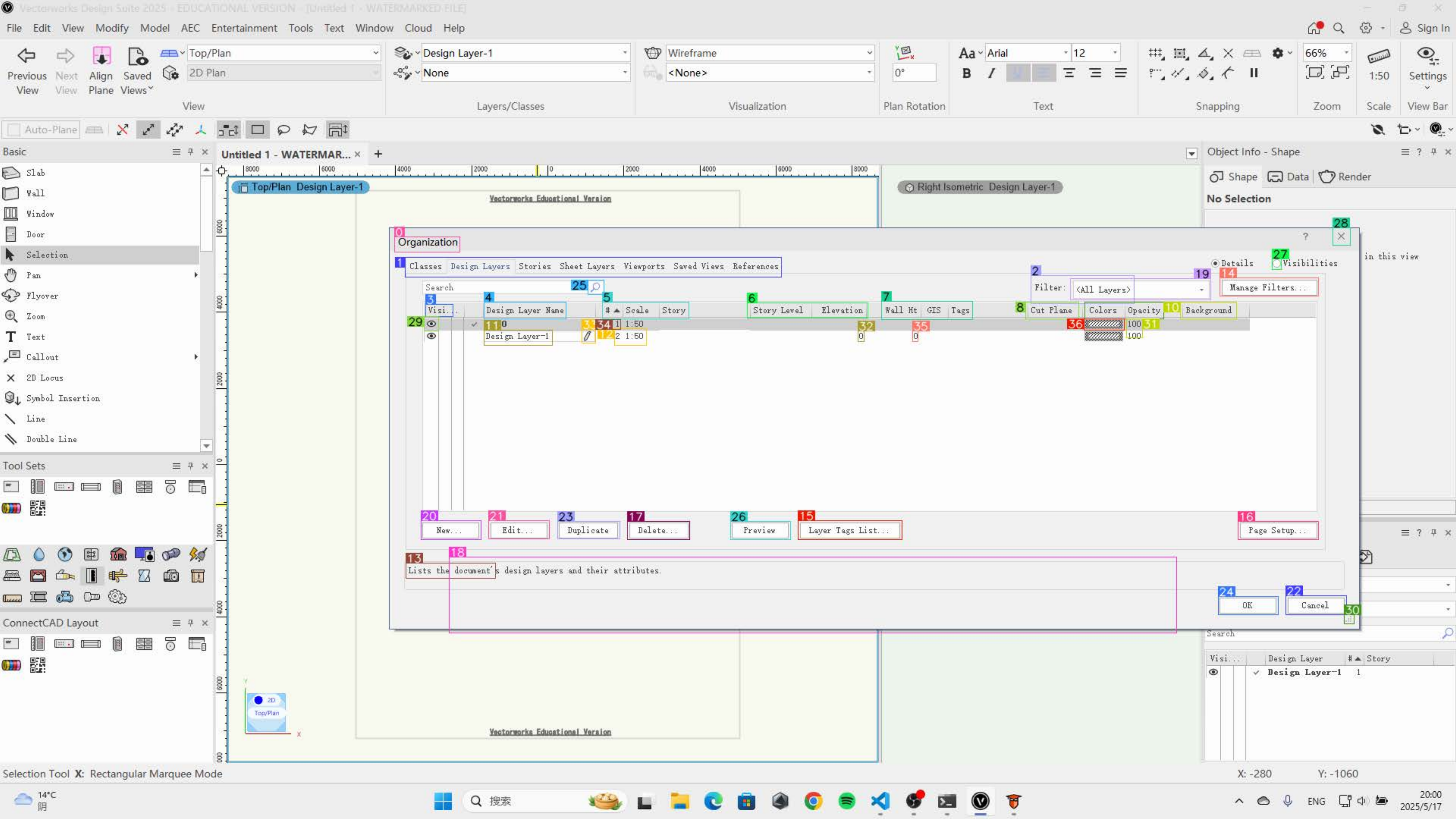}
    \caption{%
      \parbox[c][4\baselineskip][c]{\linewidth}{%
        \centering
        Open organization dialog\\
        \texttt{shortcut(ctrl + shift + O)}\\
         --
      }%
    }
    \label{fig:trajectory-b}
  \end{subfigure}\hfill
  \begin{subfigure}[b]{0.32\textwidth}
    \centering
    \includegraphics[width=\linewidth,keepaspectratio]{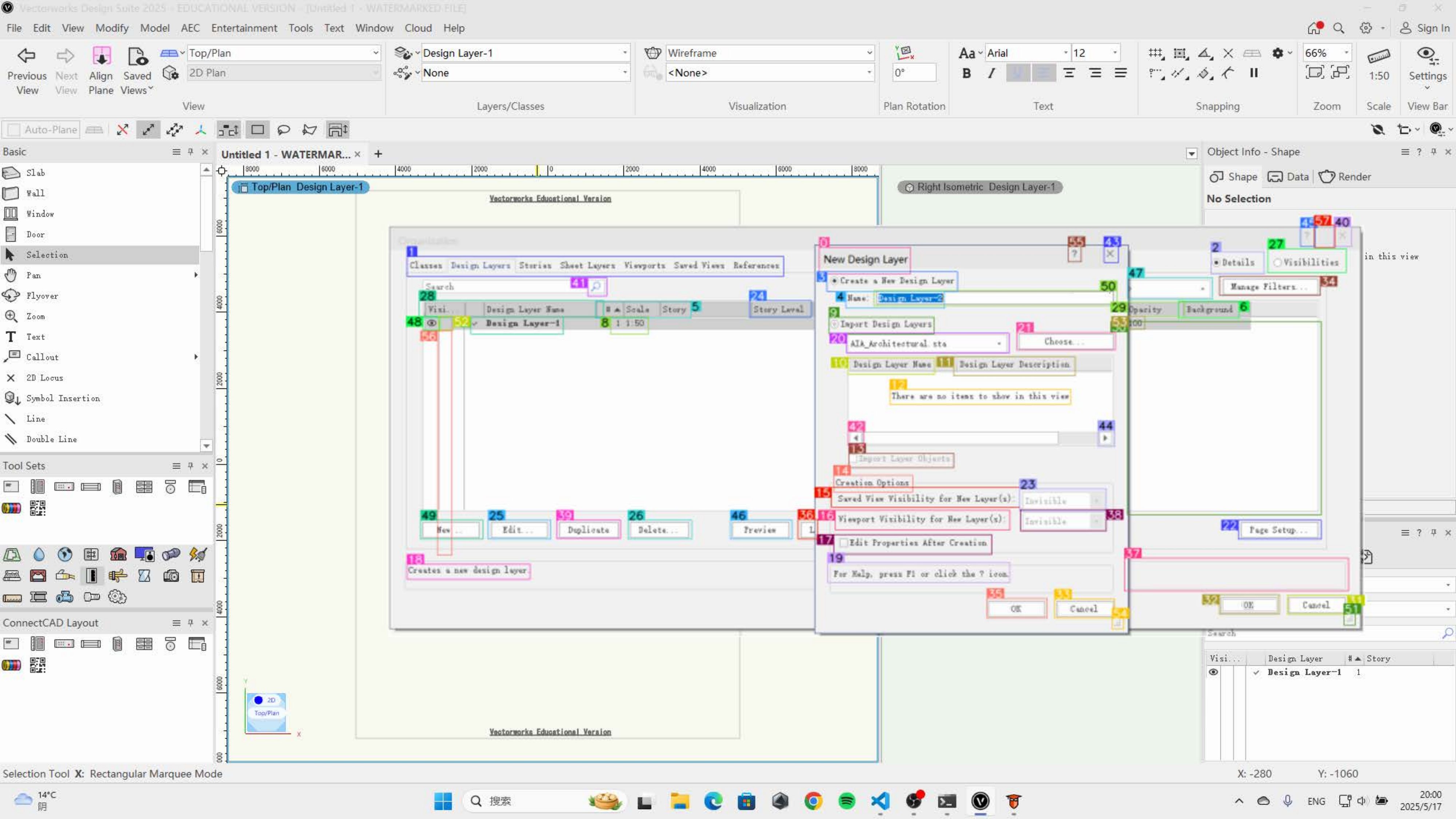}
    \caption{%
      \parbox[c][4\baselineskip][c]{\linewidth}{%
        \centering
        Click ‘New…’\\
        \texttt{move\_mouse\_to(590,699)}, \texttt{left\_click()}
      }%
    }
    \label{fig:trajectory-c}
  \end{subfigure}

  \vspace{0.5em}

  % 第二行
  \begin{subfigure}[b]{0.32\textwidth}
    \centering
    \includegraphics[width=\linewidth,keepaspectratio]{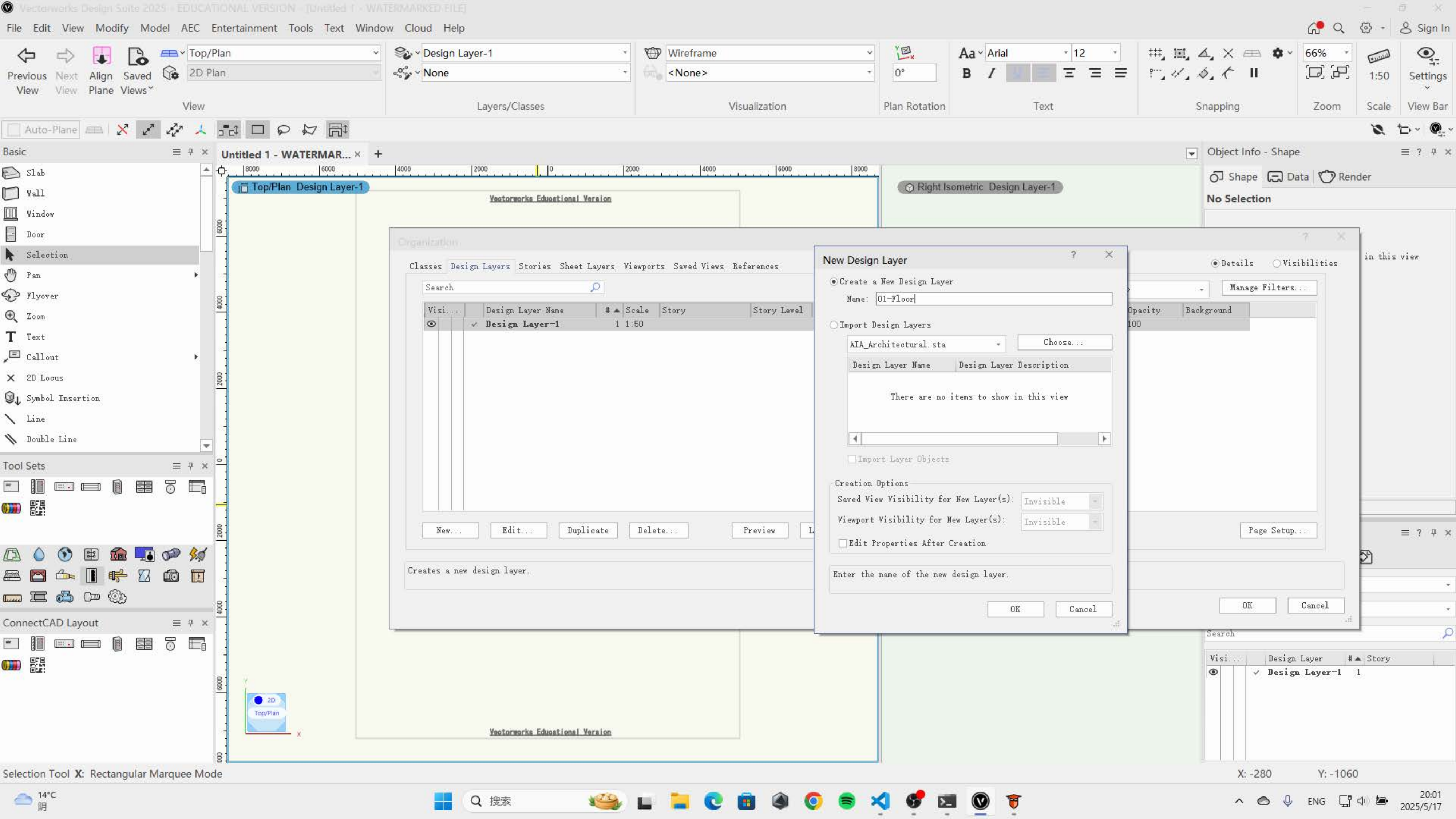}
    \caption{%
      \parbox[c][4\baselineskip][c]{\linewidth}{%
        \centering
        Type name\\
        \texttt{move\_mouse\_to(1179,396)}, \texttt{left\_click()},
        \texttt{select\_all()}, \texttt{type\_name("01-Floor")}
      }%
    }
    \label{fig:trajectory-d}
  \end{subfigure}\hfill
  \begin{subfigure}[b]{0.32\textwidth}
    \centering
    \includegraphics[width=\linewidth,keepaspectratio]{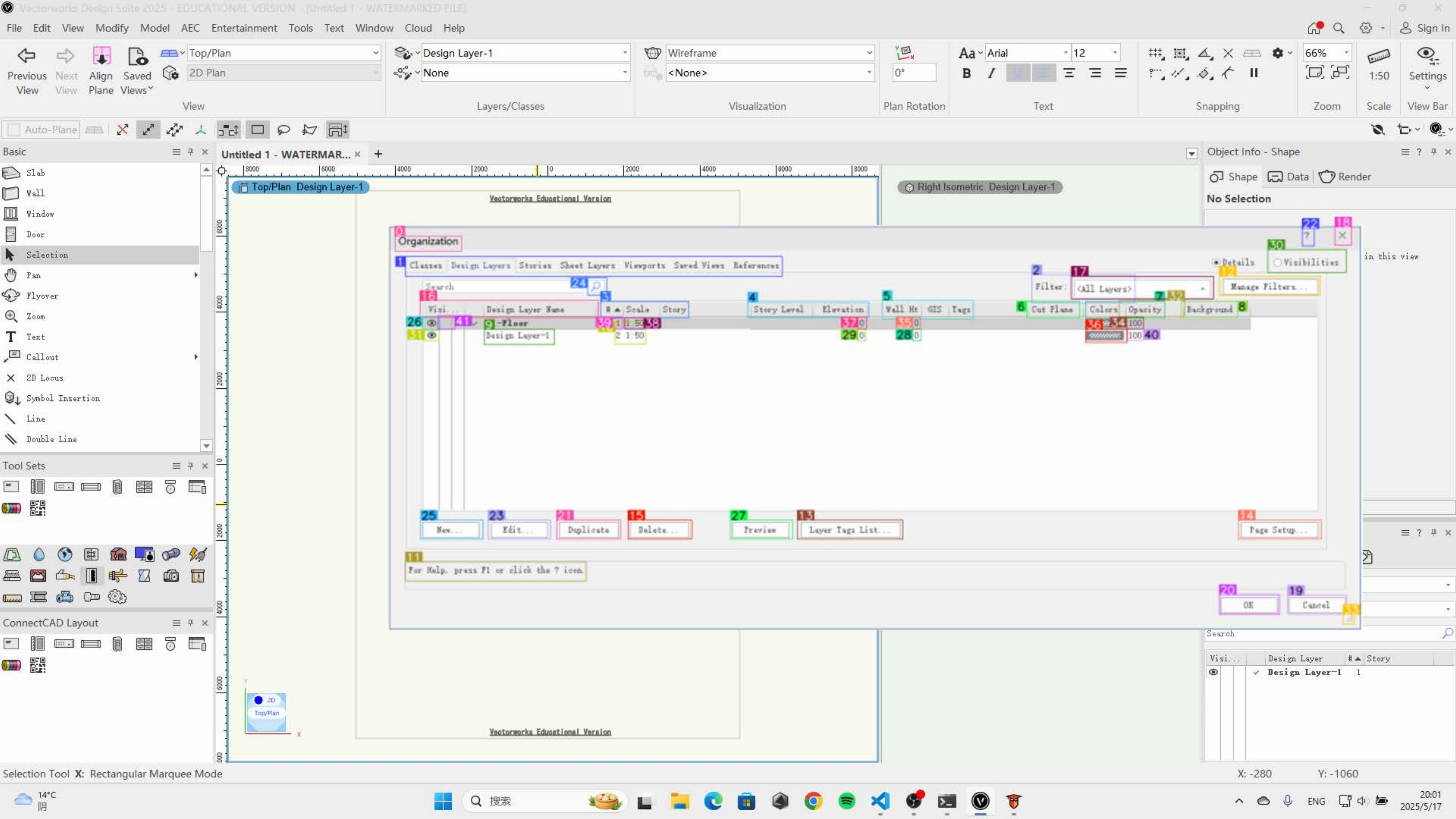}
    \caption{%
      \parbox[c][4\baselineskip][c]{\linewidth}{%
        \centering
        Confirm\\
        \texttt{press\_enter()}\\
         --
      }%
    }
    \label{fig:trajectory-e}
  \end{subfigure}\hfill
  \begin{subfigure}[b]{0.32\textwidth}
    \centering
    \includegraphics[width=\linewidth,keepaspectratio]{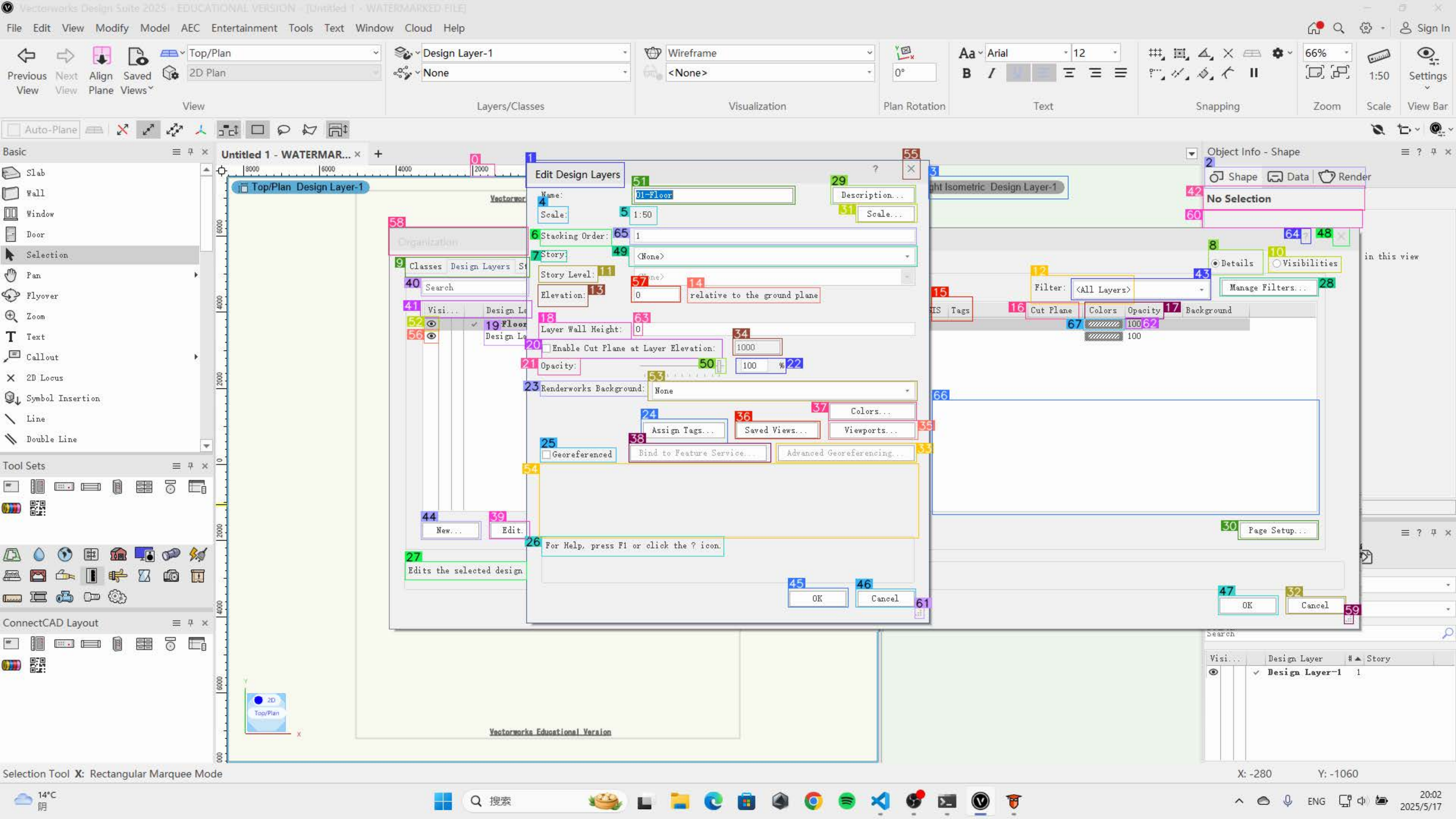}
    \caption{%
      \parbox[c][4\baselineskip][c]{\linewidth}{%
        \centering
        Click ‘Edit…’\\
        \texttt{move\_mouse\_to(684,698)}, \texttt{left\_click()}
      }%
    }
    \label{fig:trajectory-f}
  \end{subfigure}

  \vspace{0.5em}

  % 第三行
  \begin{subfigure}[b]{0.32\textwidth}
    \centering
    \includegraphics[width=\linewidth,keepaspectratio]{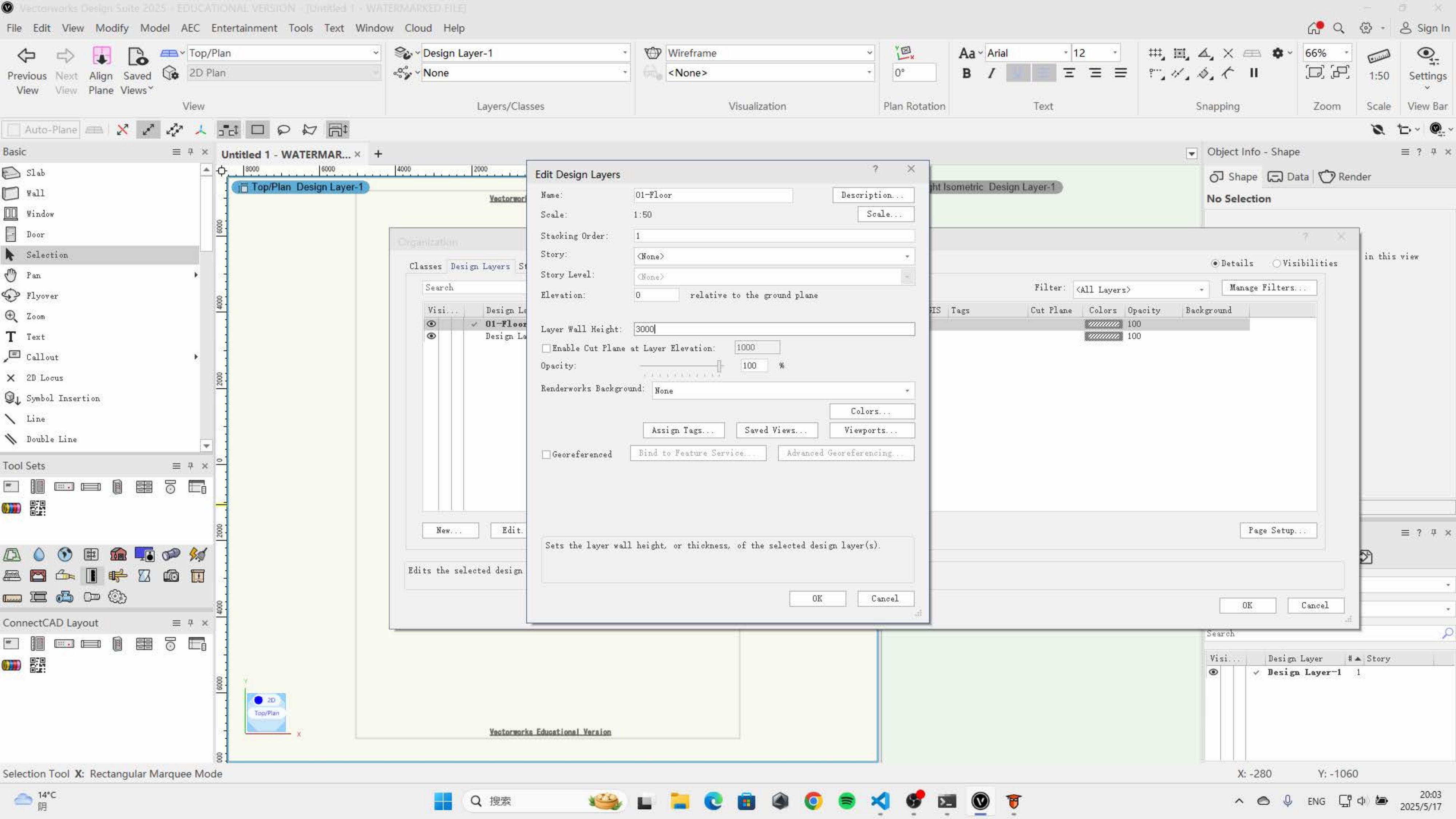}
    \caption{%
      \parbox[c][4\baselineskip][c]{\linewidth}{%
        \centering
        Edit elevation\\
        \texttt{move\_mouse\_to(870,435)}, \texttt{left\_click()},\\
        \texttt{select\_all()}, \texttt{type\_name("3000")}
      }%
    }
    \label{fig:trajectory-g}
  \end{subfigure}\hfill
  \begin{subfigure}[b]{0.32\textwidth}
    \centering
    \includegraphics[width=\linewidth,keepaspectratio]{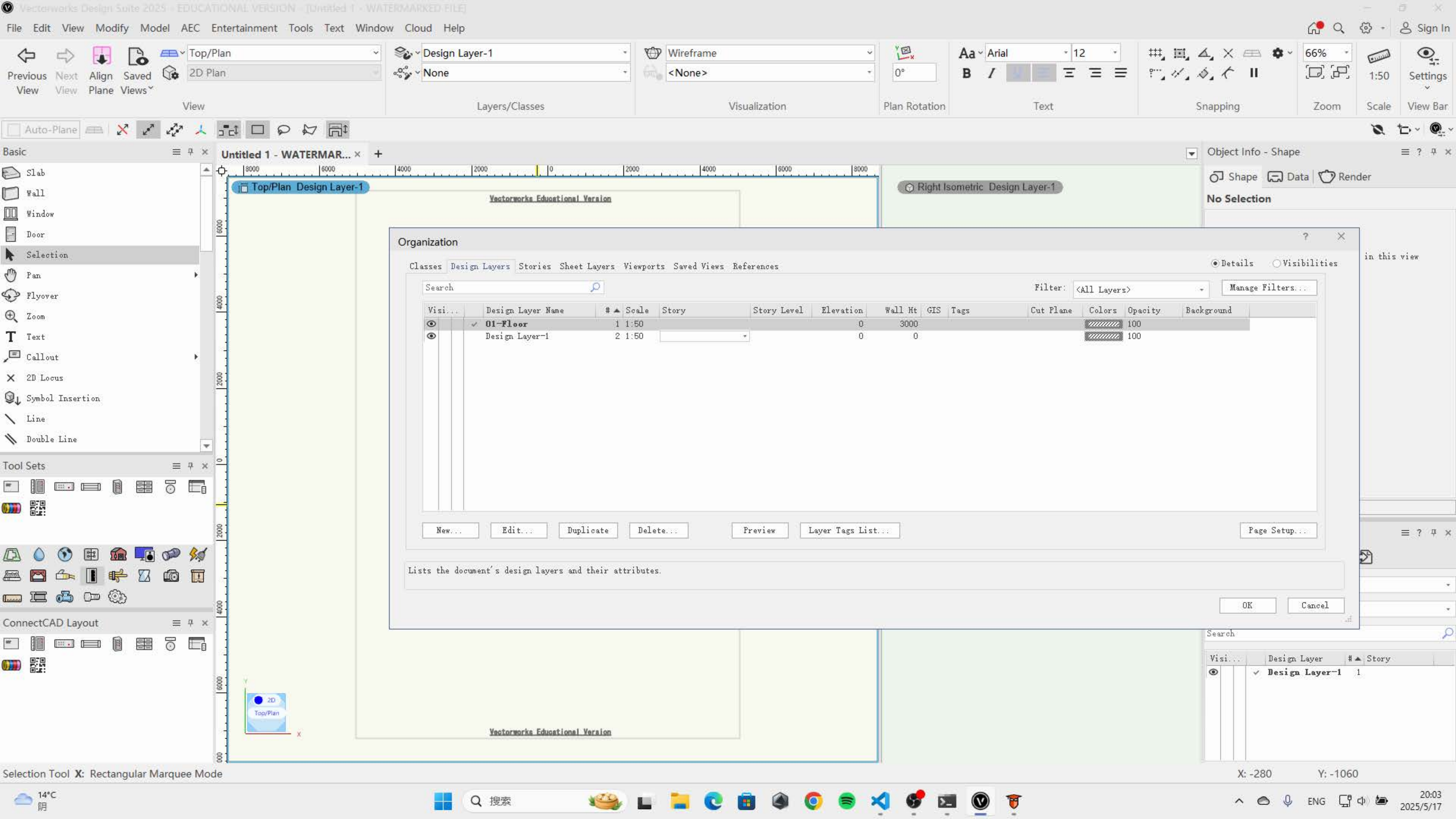}
    \caption{%
      \parbox[c][4\baselineskip][c]{\linewidth}{%
        \centering
        Confirm settings\\
        \texttt{press\_enter()}\\
         --\\
          --
      }%
    }
    \label{fig:trajectory-h}
  \end{subfigure}\hfill
  \begin{subfigure}[b]{0.32\textwidth}
    \centering
    \includegraphics[width=\linewidth,keepaspectratio]{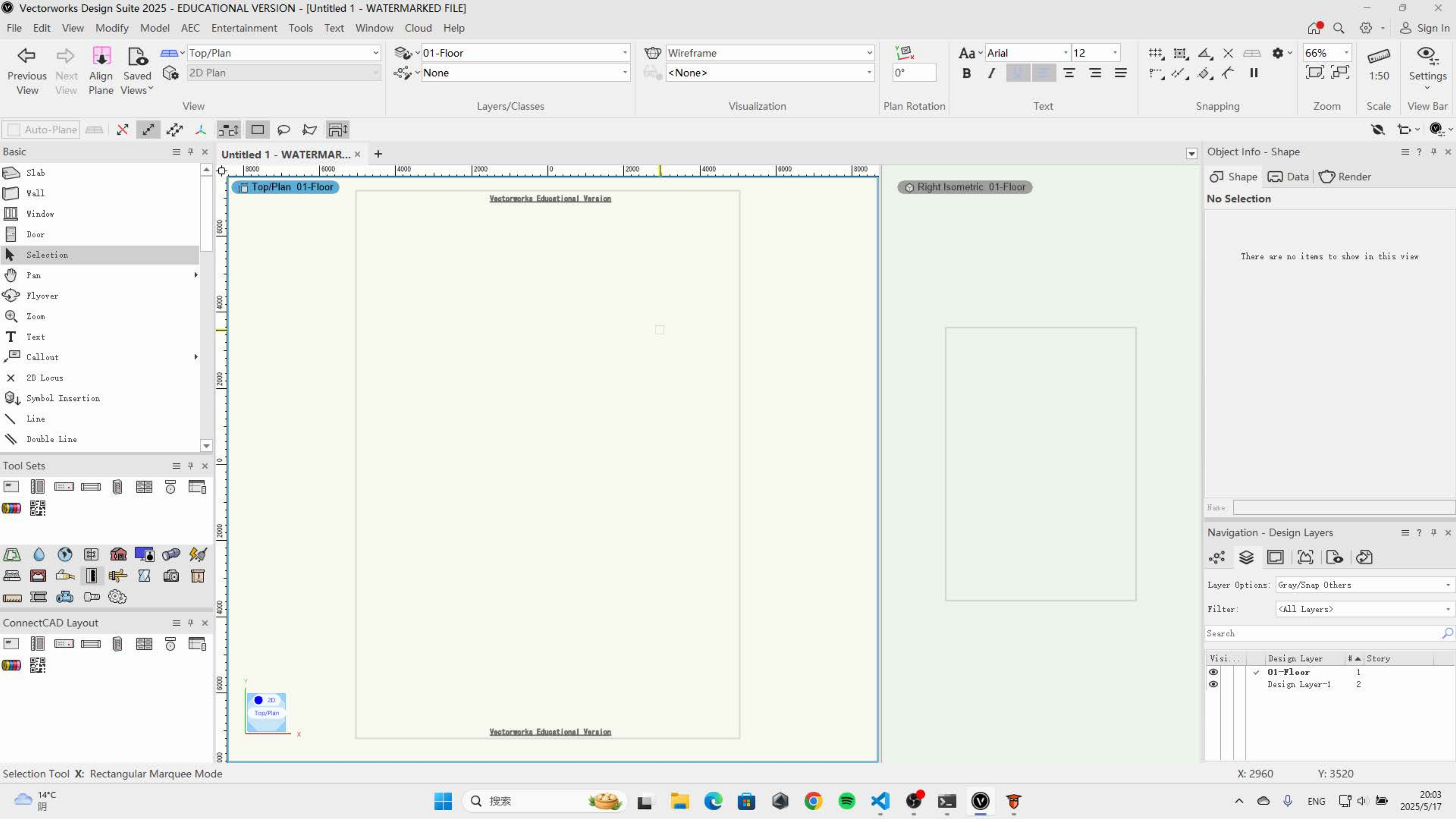}
    \caption{%
      \parbox[c][4\baselineskip][c]{\linewidth}{%
        \centering
        Final confirmation\\
        \texttt{press\_enter()}\\
         --\\
          --
      }%
    }
    \label{fig:trajectory-i}
  \end{subfigure}

  \caption{Screenshots of Floorplan Design and Design Layer creation actions.}
  \label{fig:desingandlayer}
\end{figure}

\begin{figure}[htbp]
  \centering

  % 第一行
  \begin{subfigure}[b]{0.32\textwidth}
    \centering
    \includegraphics[width=\linewidth,keepaspectratio]{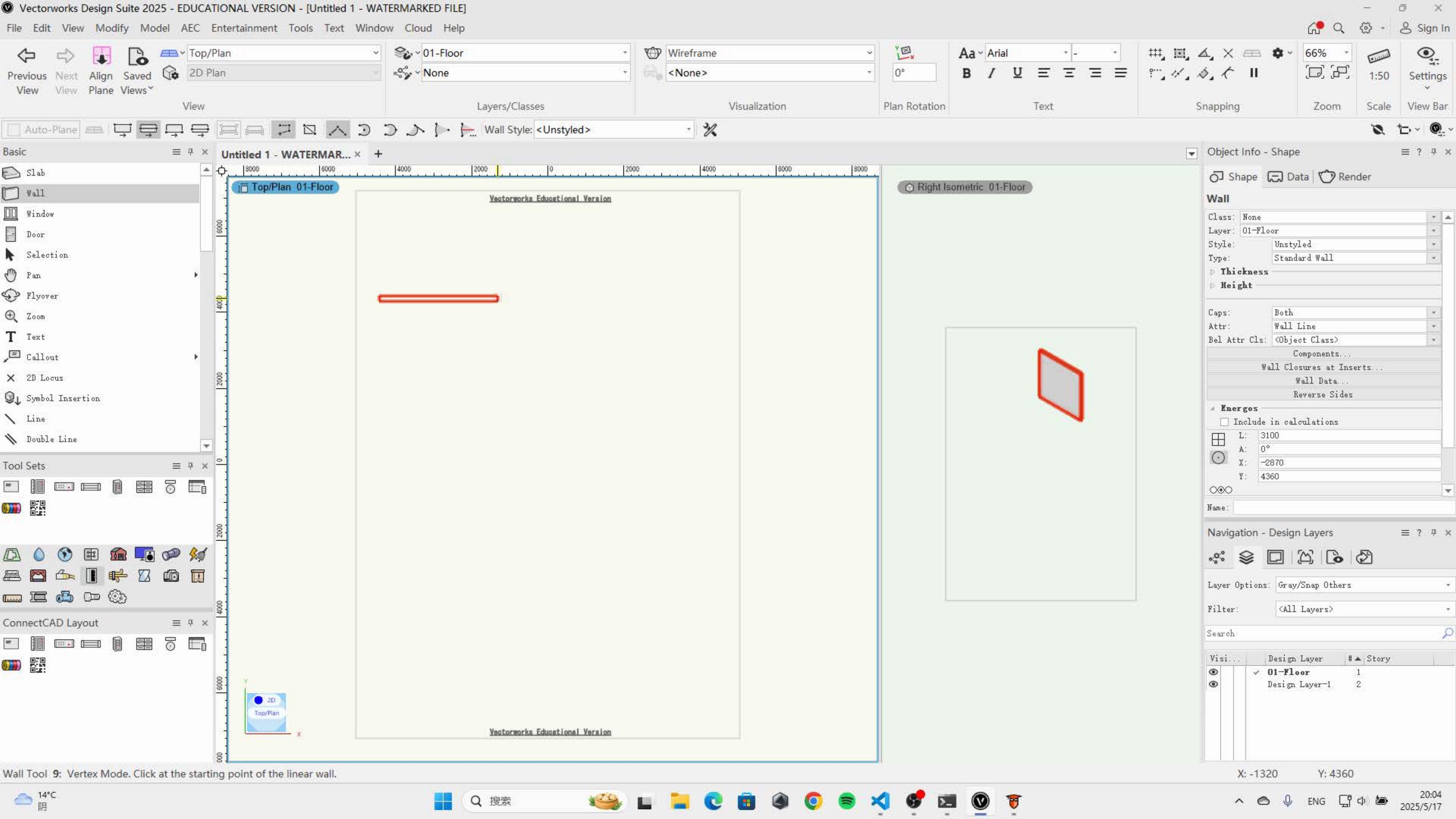}
    \caption{%
      \parbox[c][7\baselineskip][c]{\linewidth}{%
        \centering
        First external wall\\
        \texttt{shortcut(combo='9')}\\
        \texttt{move\_mouse\_to(x=500,y=393)}\\
        \texttt{left\_click()}\\
        \texttt{move\_mouse\_to(x=656,y=393)}\\
        \texttt{left\_click()}\\
        \texttt{press\_enter()}
      }%
    }
    \label{fig:trajectory-a}
  \end{subfigure}\hfill
  \begin{subfigure}[b]{0.32\textwidth}
    \centering
    \includegraphics[width=\linewidth,keepaspectratio]{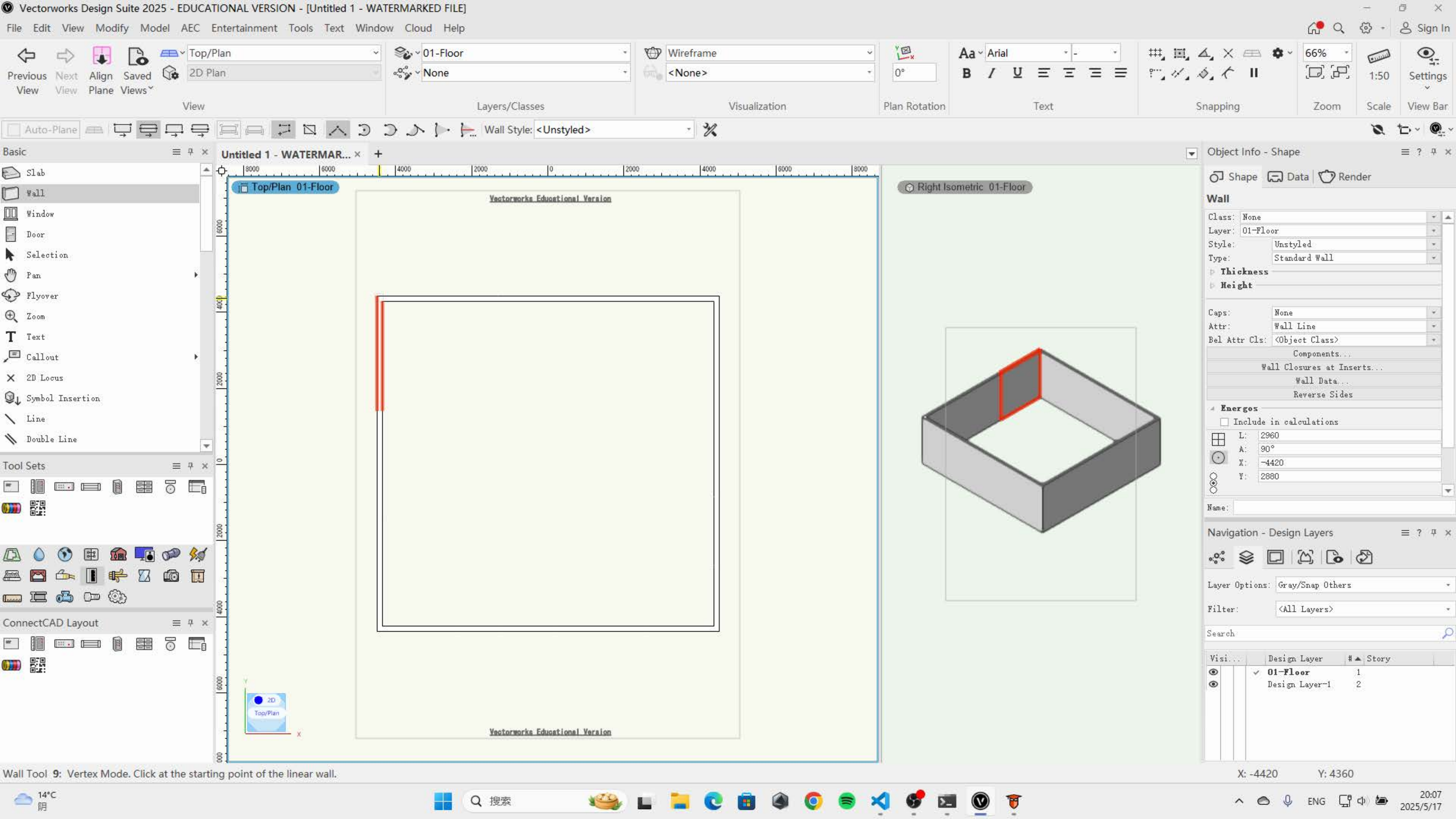}
    \caption{%
      \parbox[c][7\baselineskip][c]{\linewidth}{%
        \centering
        Last external wall\\
        \texttt{shortcut(combo='9')}\\
        \texttt{move\_mouse\_to(x=500,y=542)}\\
        \texttt{left\_click()}\\
        \texttt{move\_mouse\_to(x=500,y=393)}\\
        \texttt{left\_click()}\\
        \texttt{press\_enter()}
      }%
    }
    \label{fig:trajectory-b}
  \end{subfigure}\hfill
  \begin{subfigure}[b]{0.32\textwidth}
    \centering
    \includegraphics[width=\linewidth,keepaspectratio]{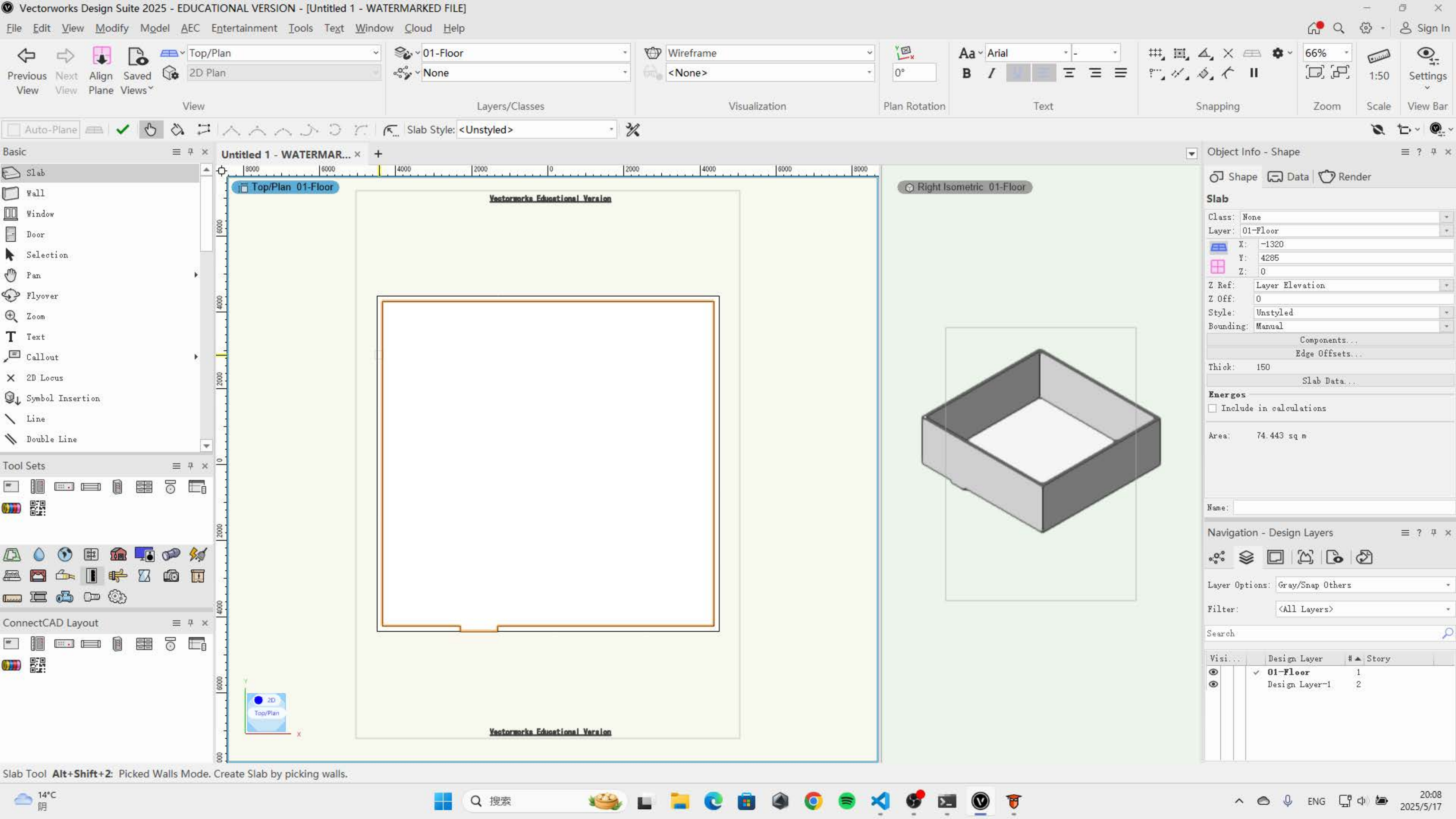}
    \caption{%
      \parbox[c][7\baselineskip][c]{\linewidth}{%
        \centering
        Create slab by picking external walls\\
        \texttt{shortcut(combo='alt+shift+2')}\\
        \texttt{move\_mouse\_to(x=578,y=393)}\\
        \texttt{left\_click()}\\
        …\\
        \texttt{left\_click()}\\
        \texttt{press\_enter()}
      }%
    }
    \label{fig:trajectory-c}
  \end{subfigure}

  \vspace{0.5em}

  % 第二行
  \begin{subfigure}[b]{0.32\textwidth}
    \centering
    \includegraphics[width=\linewidth,keepaspectratio]{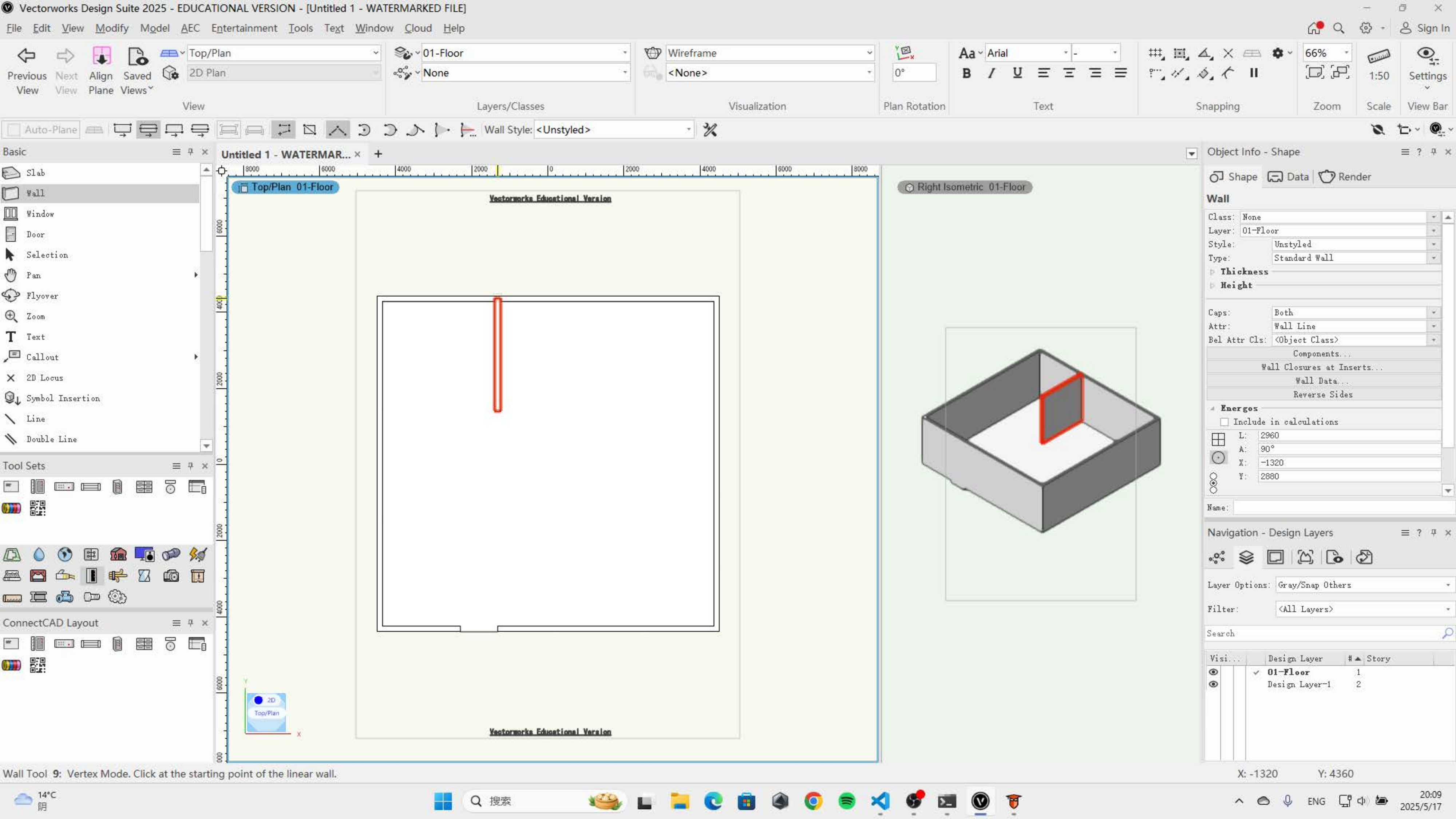}
    \caption{%
      \parbox[c][7\baselineskip][c]{\linewidth}{%
        \centering
        Create first internal wall\\
        \texttt{shortcut(combo='9')}\\
        \texttt{move\_mouse\_to(x=656,y=542)}\\
        \texttt{left\_click()}\\
        \texttt{move\_mouse\_to(x=656,y=393)}\\
        \texttt{left\_click()}\\
        \texttt{press\_enter()}
      }%
    }
    \label{fig:trajectory-d}
  \end{subfigure}\hfill
  \begin{subfigure}[b]{0.32\textwidth}
    \centering
    \includegraphics[width=\linewidth,keepaspectratio]{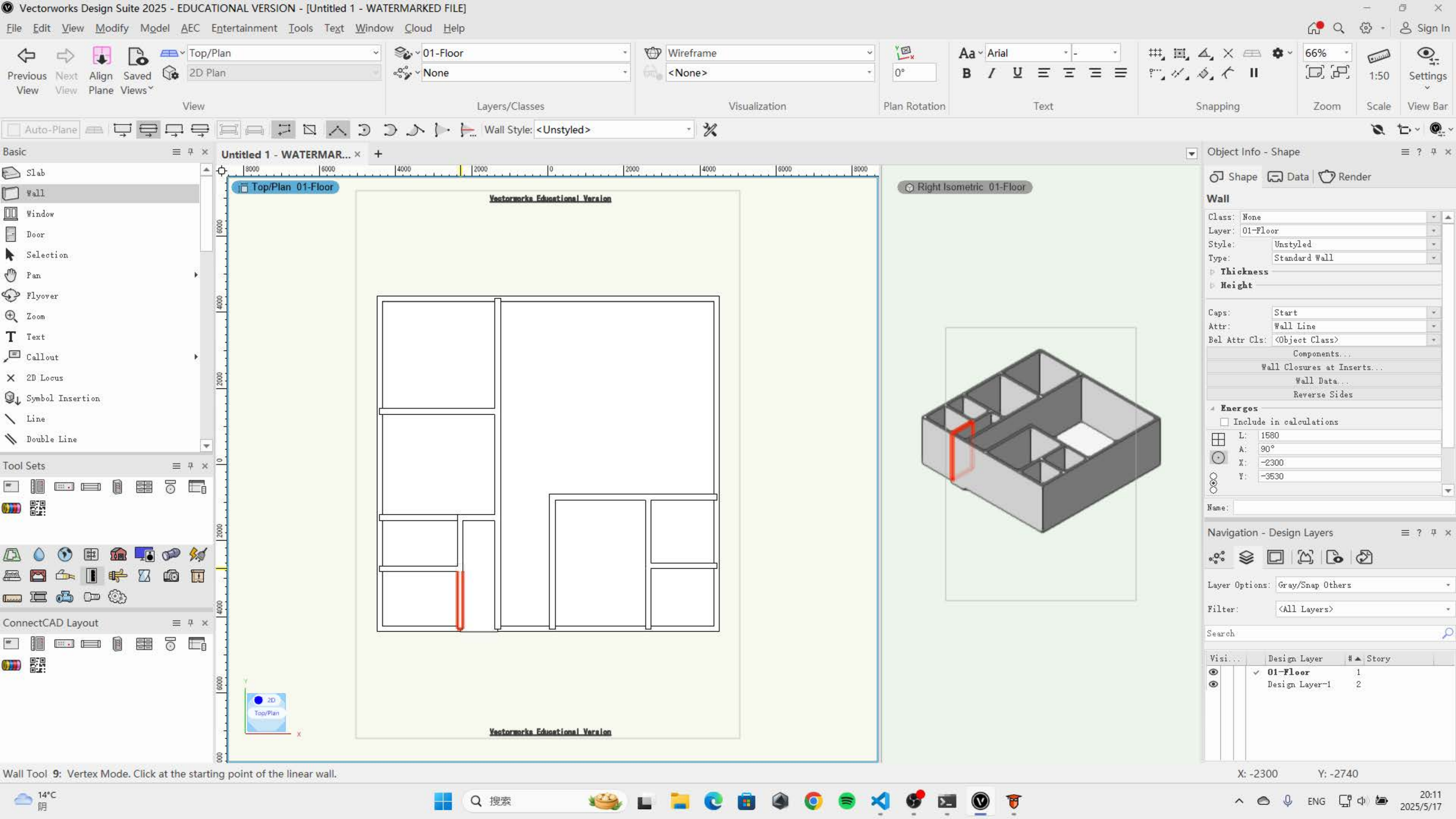}
    \caption{%
      \parbox[c][7\baselineskip][c]{\linewidth}{%
        \centering
        Final internal wall\\
        \texttt{shortcut(combo='9')}\\
        \texttt{move\_mouse\_to(x=607,y=829)}\\
        \texttt{left\_click()}\\
        \texttt{move\_mouse\_to(x=607,y=749)}\\
        \texttt{left\_click()}\\
        \texttt{press\_enter()}
      }%
    }
    \label{fig:trajectory-e}
  \end{subfigure}\hfill
  \begin{subfigure}[b]{0.32\textwidth}
    \centering
    \includegraphics[width=\linewidth,keepaspectratio]{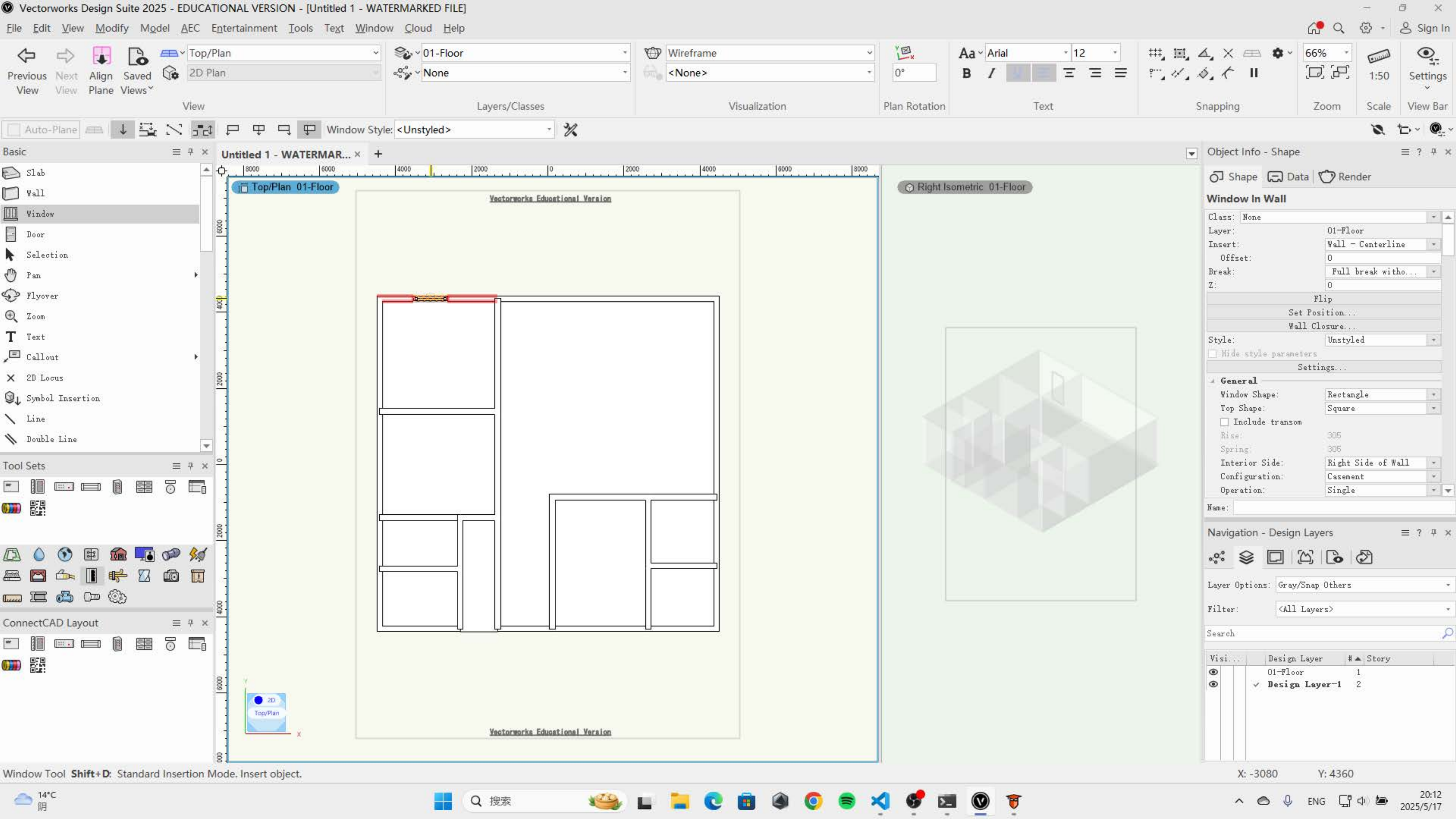}
    \caption{%
      \parbox[c][7\baselineskip][c]{\linewidth}{%
        \centering
        Insert first window\\
        \texttt{shortcut(combo='shift + d')}\\
        \texttt{move\_mouse\_to(x=568,y=393)}\\
        \texttt{left\_click()}\\
        \texttt{press\_enter()}\\
        --\\
        --
      }%
    }
    \label{fig:trajectory-f}
  \end{subfigure}

  \vspace{0.5em}

  % 第三行
  \begin{subfigure}[b]{0.32\textwidth}
    \centering
    \includegraphics[width=\linewidth,keepaspectratio]{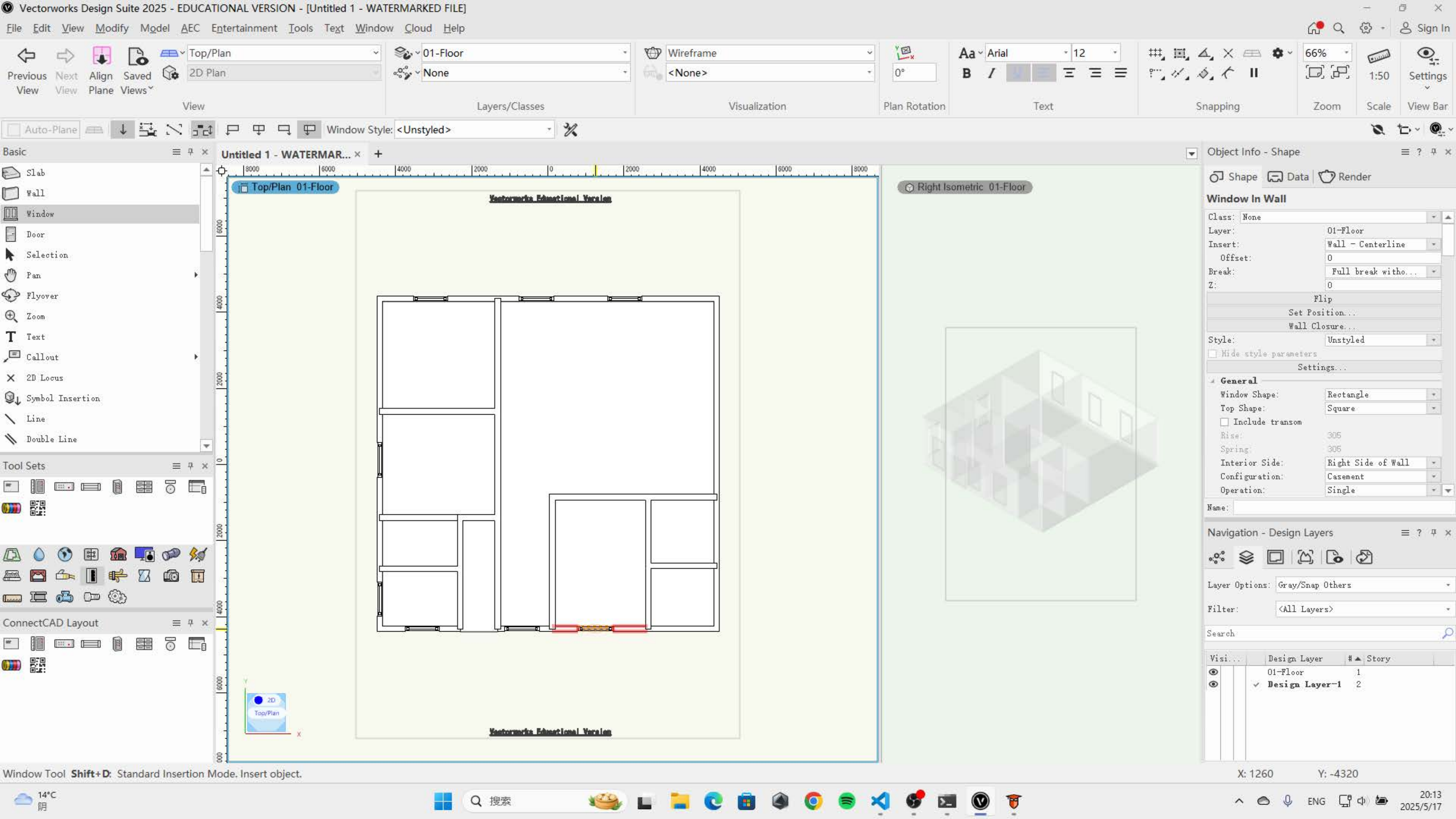}
    \caption{%
      \parbox[c][7\baselineskip][c]{\linewidth}{%
        \centering
        Insert final window\\
        \texttt{shortcut(combo='shift + d')}\\
        \texttt{move\_mouse\_to(x=785,y=829)}\\
        \texttt{left\_click()}\\
        \texttt{press\_enter()}\\
        --\\
        --
      }%
    }
    \label{fig:trajectory-g}
  \end{subfigure}\hfill
  \begin{subfigure}[b]{0.32\textwidth}
    \centering
    \includegraphics[width=\linewidth,keepaspectratio]{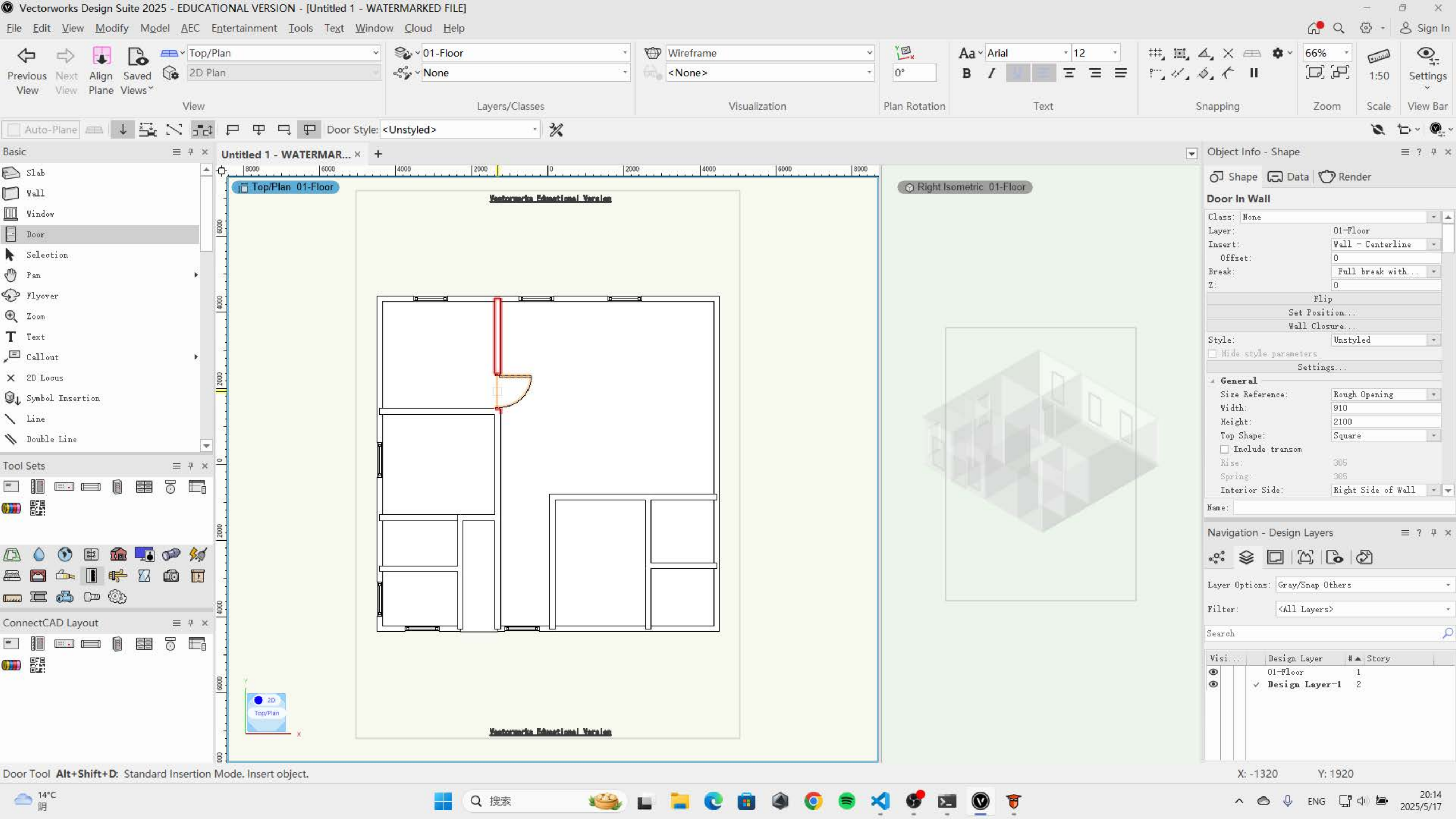}
    \caption{%
      \parbox[c][7\baselineskip][c]{\linewidth}{%
        \centering
        Insert first door\\
        \texttt{shortcut(combo='alt+shift+d')}\\
        \texttt{move\_mouse\_to(x=133,y=309)}\\
        \texttt{left\_click()}\\
        \texttt{press\_enter()}\\
        --\\
        --
      }%
    }
    \label{fig:trajectory-h}
  \end{subfigure}\hfill
  \begin{subfigure}[b]{0.32\textwidth}
    \centering
    \includegraphics[width=\linewidth,keepaspectratio]{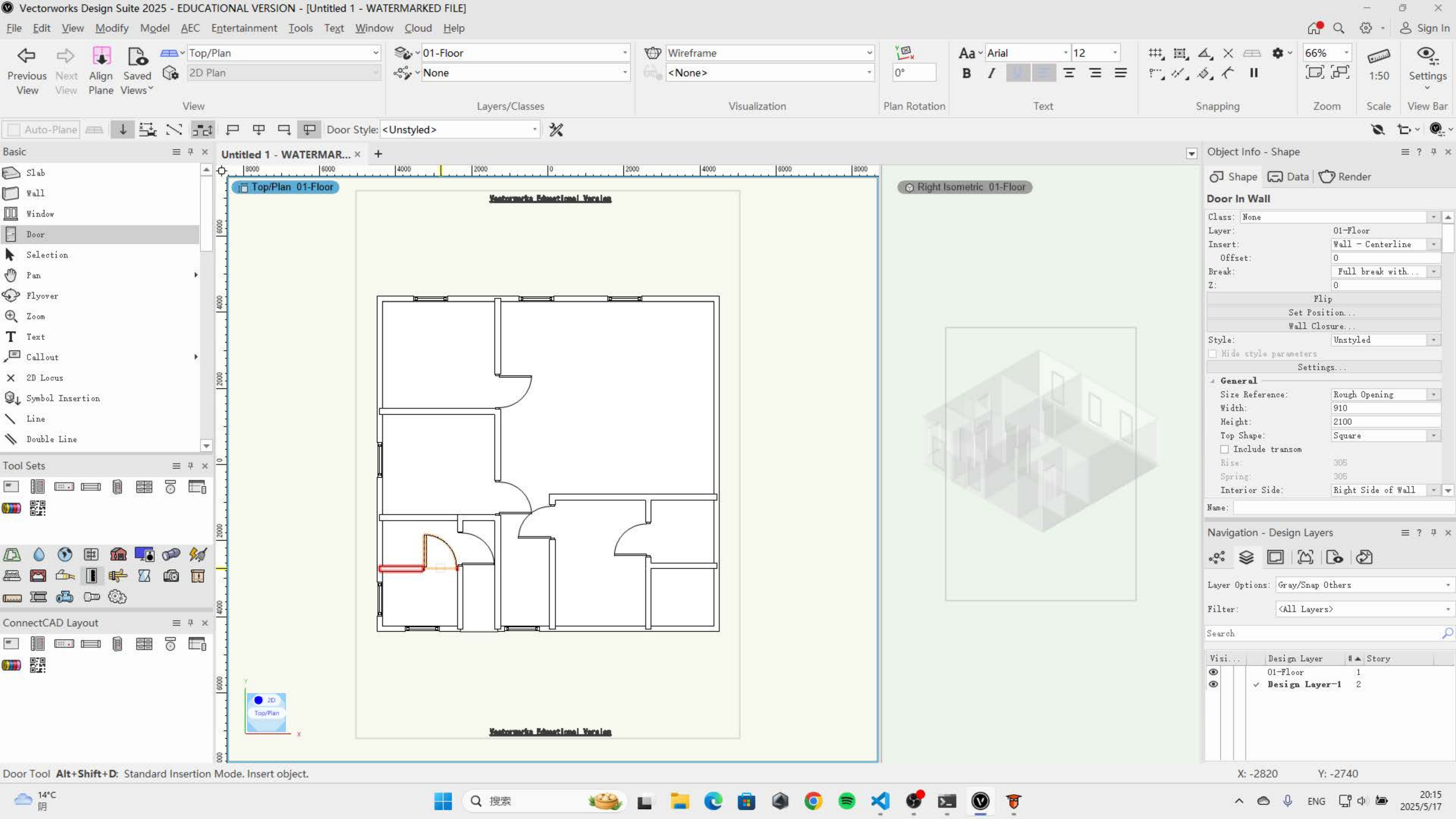}
    \caption{%
      \parbox[c][7\baselineskip][c]{\linewidth}{%
        \centering
        Insert final door\\
        \texttt{move\_mouse\_to(x=581,y=749)}\\
        \texttt{left\_click()}\\
        \texttt{press\_enter()}\\
        --\\
        --\\
        --
      }%
    }
    \label{fig:trajectory-i}
  \end{subfigure}

  \caption{Screenshots of element creations.}
  \label{fig:elementcreation}
\end{figure}

\begin{figure}[htbp]
  \centering

  % 第一行：两张图
  \begin{subfigure}[b]{0.45\textwidth}
    \centering
    \includegraphics[width=\linewidth,keepaspectratio]{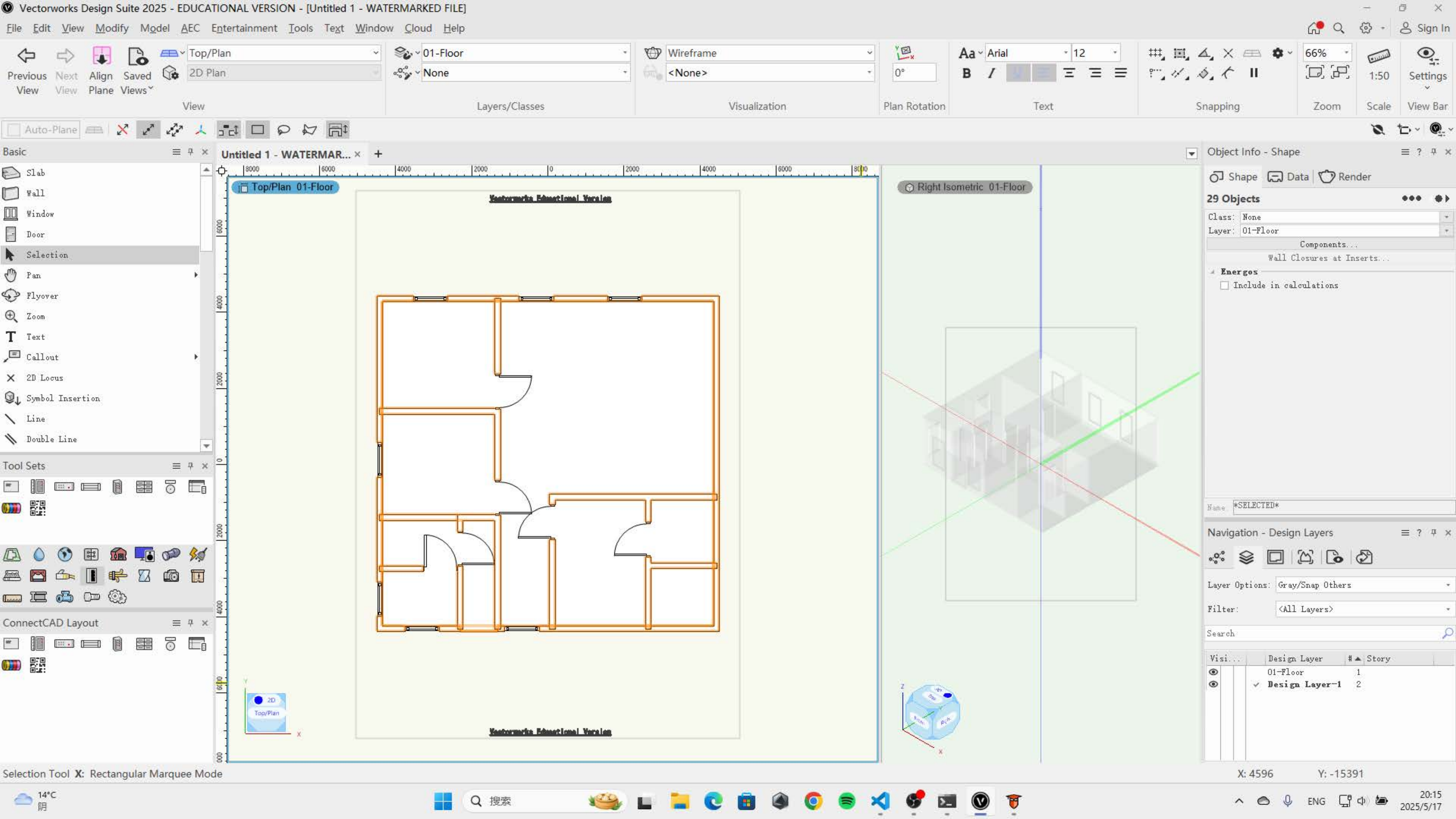}
    \caption{%
      \parbox[c][7\baselineskip][c]{\linewidth}{%
        \centering
        Select all components\\
        \texttt{select\_all()} \\
        --\\
        --
      }%
    }
    \label{fig:trajectory-a}
  \end{subfigure}\hfill
  \begin{subfigure}[b]{0.45\textwidth}
    \centering
    \includegraphics[width=\linewidth,keepaspectratio]{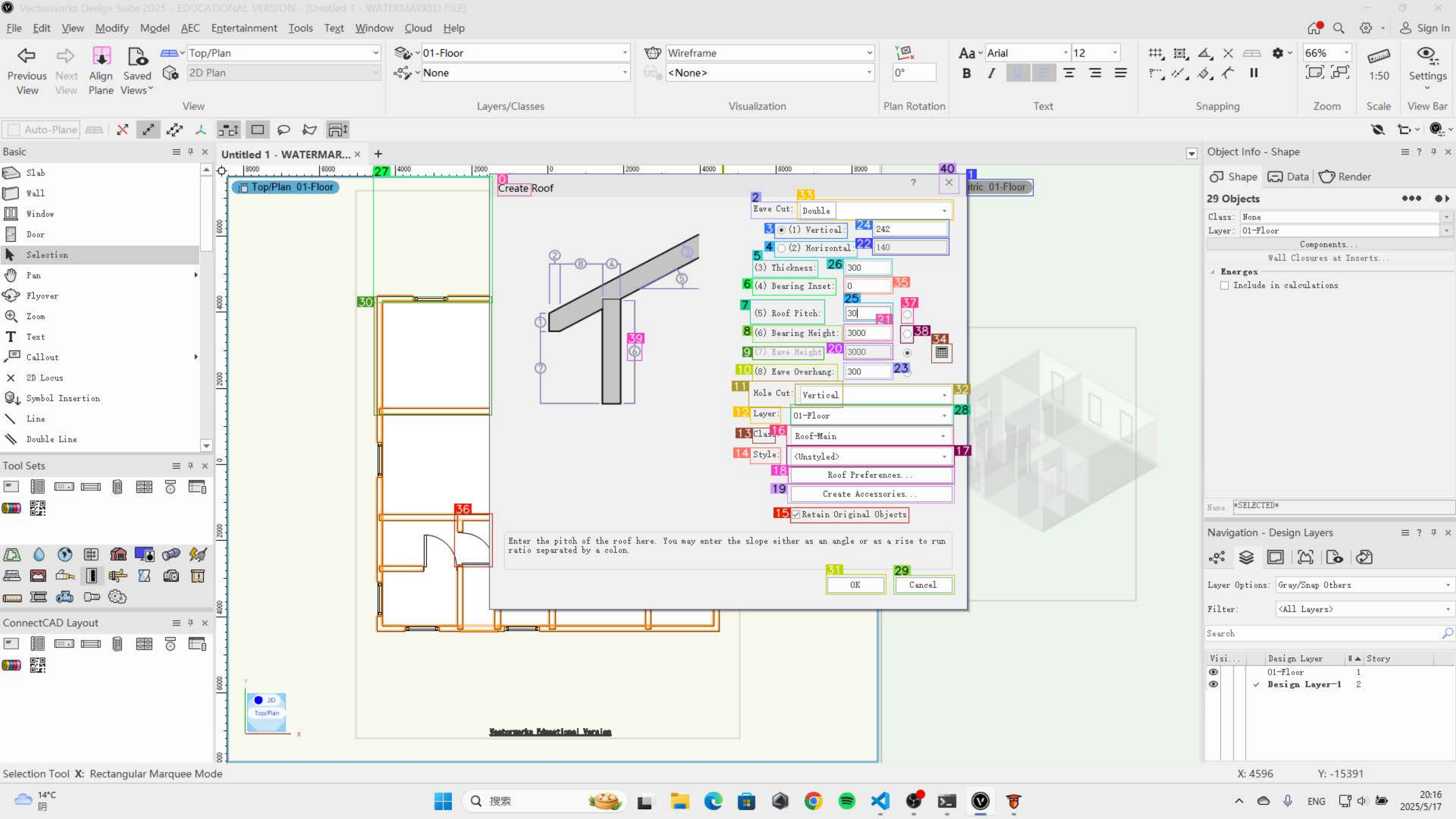}
    \caption{%
      \parbox[c][7\baselineskip][c]{\linewidth}{%
        \centering
        Active roof tool\\
        \texttt{shortcut(combo="ctrl + alt + shift + 1")}\\
        --\\
        --
      }%
    }
    \label{fig:trajectory-b}
  \end{subfigure}

  \vspace{0.5em}

  % 第二行：两张图
  \begin{subfigure}[b]{0.45\textwidth}
    \centering
    \includegraphics[width=\linewidth,keepaspectratio]{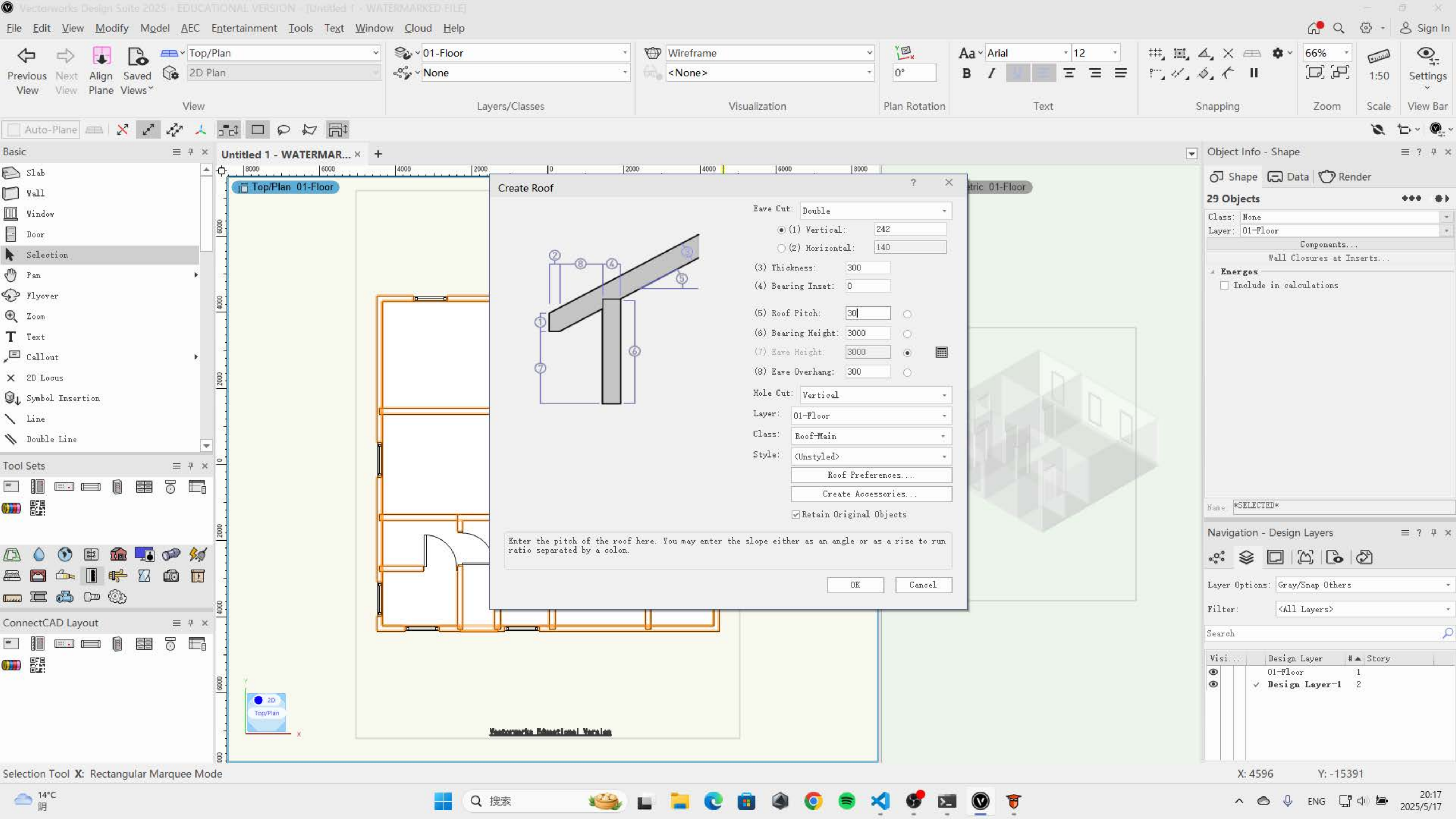}
    \caption{%
      \parbox[c][7\baselineskip][c]{\linewidth}{%
        \centering
        Set parameters\\
        \texttt{move\_mouse\_to(1145,353)}\\
        \texttt{left\_click()}\\
        \dots\\
        \texttt{type\_name("30")}
      }%
    }
    \label{fig:trajectory-c}
  \end{subfigure}\hfill
  \begin{subfigure}[b]{0.45\textwidth}
    \centering
    \includegraphics[width=\linewidth,keepaspectratio]{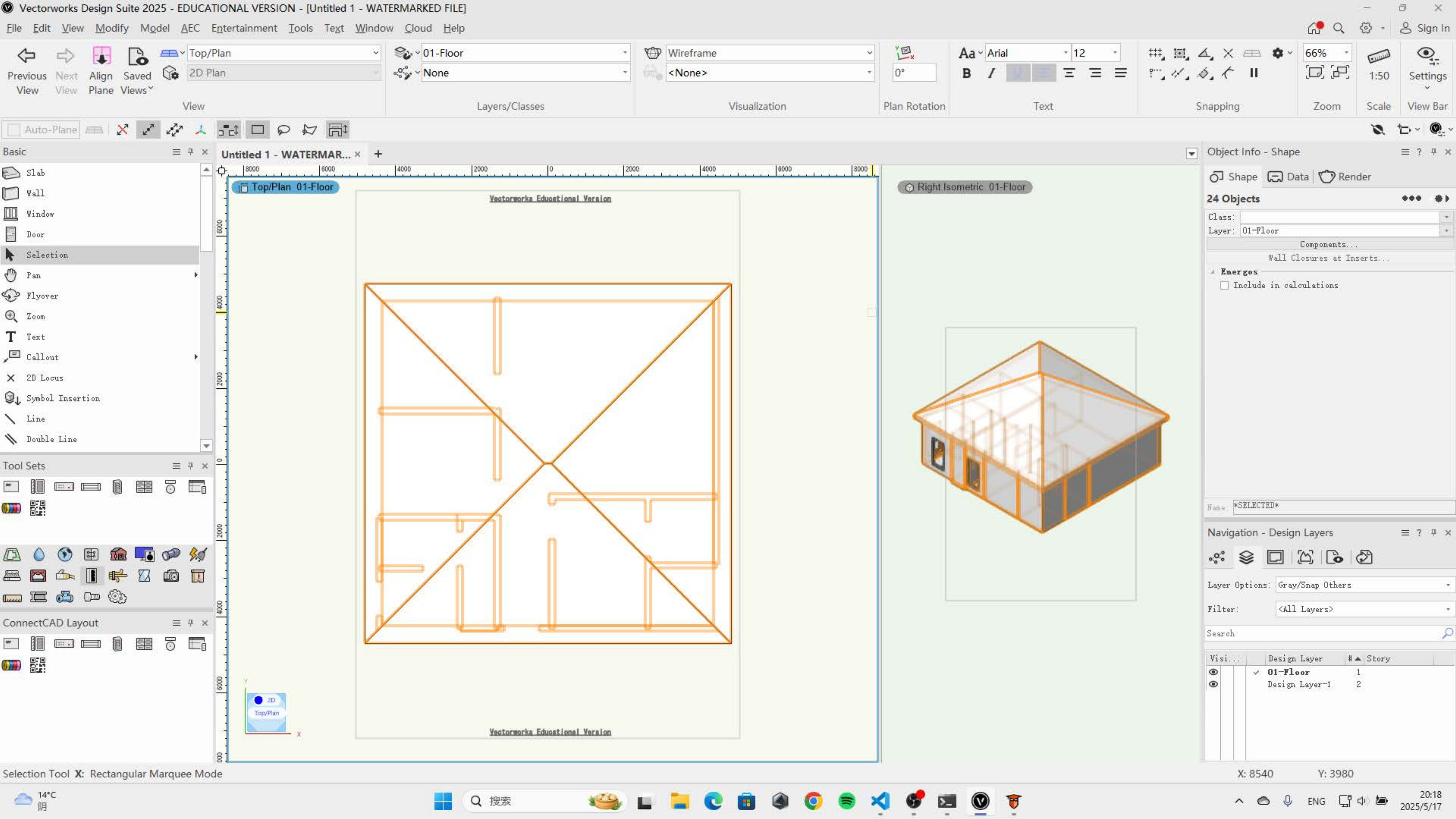}
    \caption{%
      \parbox[c][7\baselineskip][c]{\linewidth}{%
        \centering
        Confirm roof\\
        \texttt{press\_enter()}\\
        --\\
        --
      }%
    }
    \label{fig:trajectory-d}
  \end{subfigure}

  \caption{Screenshots of roof creation steps.}
  \label{fig:roof-creations}
\end{figure}

Given the task: \textit{Generate a one-storey office building with a large open workspace
occupying most of the floor area. The layout should also include two enclosed meeting rooms, a manager’s office, a small pantry, and two restrooms.} The floorplan design and new design layer creation are illustrated in Figure~\ref{fig:desingandlayer}, element creation is shown in Figure~\ref{fig:elementcreation}, and roof creation is depicted in Figure~\ref{fig:roof-creations}.

\subsection{Task 16 Action Examples}

\begin{figure}[htbp]
  \centering

  % 第一行
  \begin{subfigure}[b]{0.32\textwidth}
    \centering
    \includegraphics[width=\linewidth,keepaspectratio]{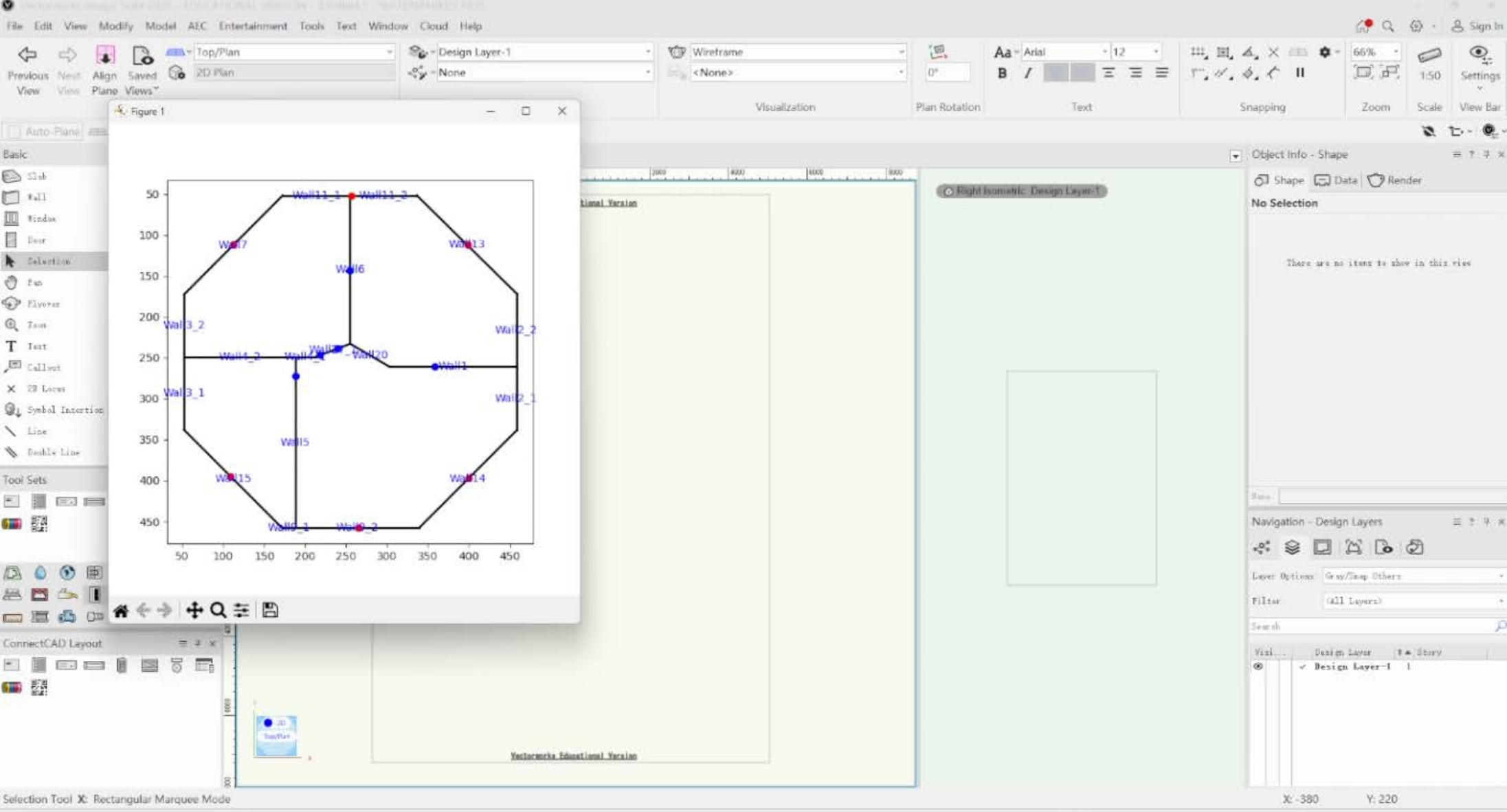}
    \caption{%
      \parbox[c][4\baselineskip][c]{\linewidth}{%
        \centering
        Floorplan design\\
         --\\
          --
      }%
    }
    \label{fig:trajectory-a}
  \end{subfigure}\hfill
  \begin{subfigure}[b]{0.32\textwidth}
    \centering
    \includegraphics[width=\linewidth,keepaspectratio]{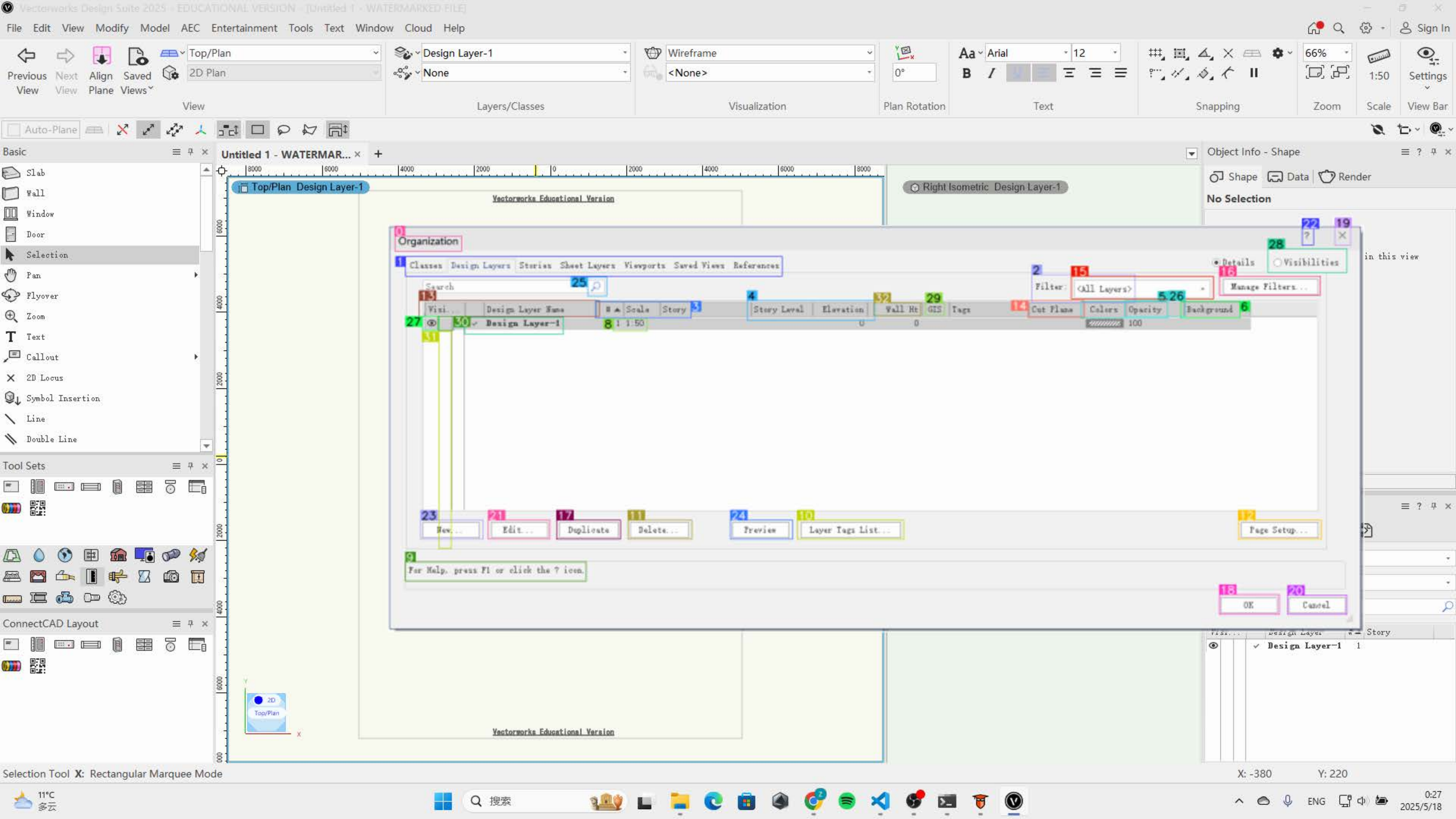}
    \caption{%
      \parbox[c][4\baselineskip][c]{\linewidth}{%
        \centering
        Open organization dialog\\
        \texttt{shortcut(ctrl + shift + O)}\\
         --
      }%
    }
    \label{fig:trajectory-b}
  \end{subfigure}\hfill
  \begin{subfigure}[b]{0.32\textwidth}
    \centering
    \includegraphics[width=\linewidth,keepaspectratio]{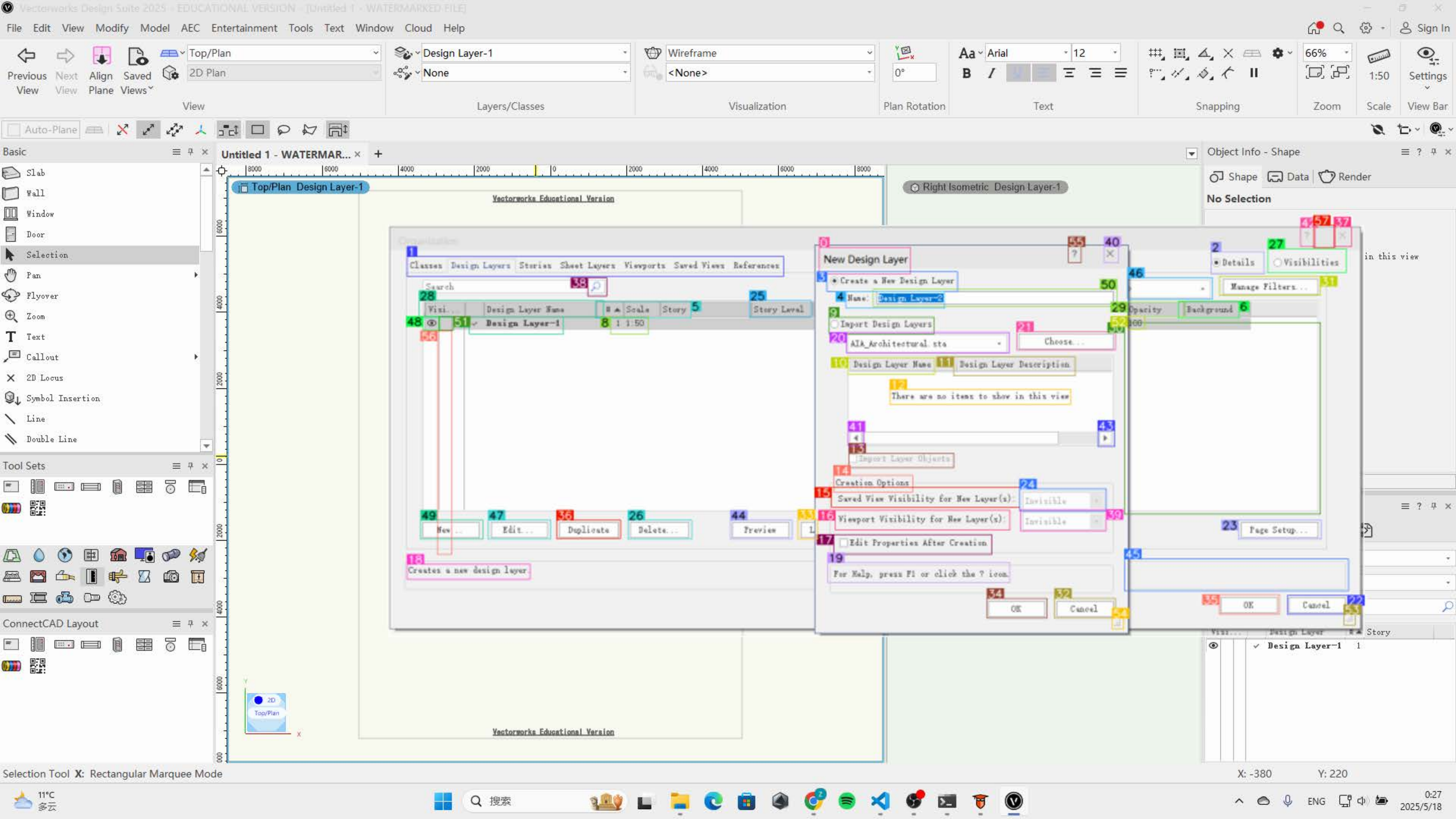}
    \caption{%
      \parbox[c][4\baselineskip][c]{\linewidth}{%
        \centering
        Click ‘New…’\\
        \texttt{move\_mouse\_to(595,699)}, \texttt{left\_click()}
      }%
    }
    \label{fig:trajectory-c}
  \end{subfigure}

  \vspace{0.5em}

  % 第二行
  \begin{subfigure}[b]{0.32\textwidth}
    \centering
    \includegraphics[width=\linewidth,keepaspectratio]{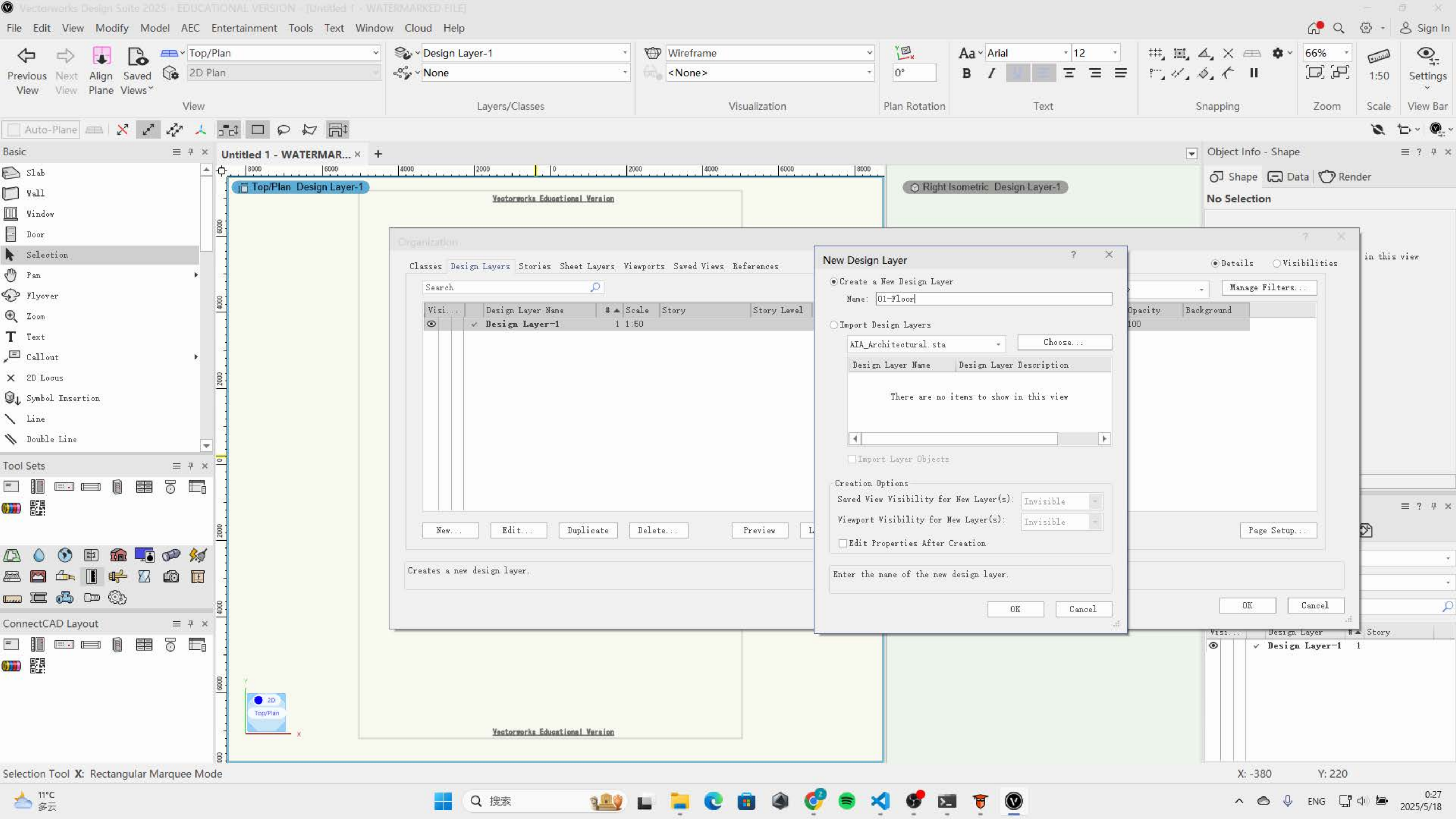}
    \caption{%
      \parbox[c][4\baselineskip][c]{\linewidth}{%
        \centering
        Type name\\
        \texttt{move\_mouse\_to(1179,396)}, \texttt{left\_click()},
        \texttt{select\_all()}, \texttt{type\_name("01-Floor")}
      }%
    }
    \label{fig:trajectory-d}
  \end{subfigure}\hfill
  \begin{subfigure}[b]{0.32\textwidth}
    \centering
    \includegraphics[width=\linewidth,keepaspectratio]{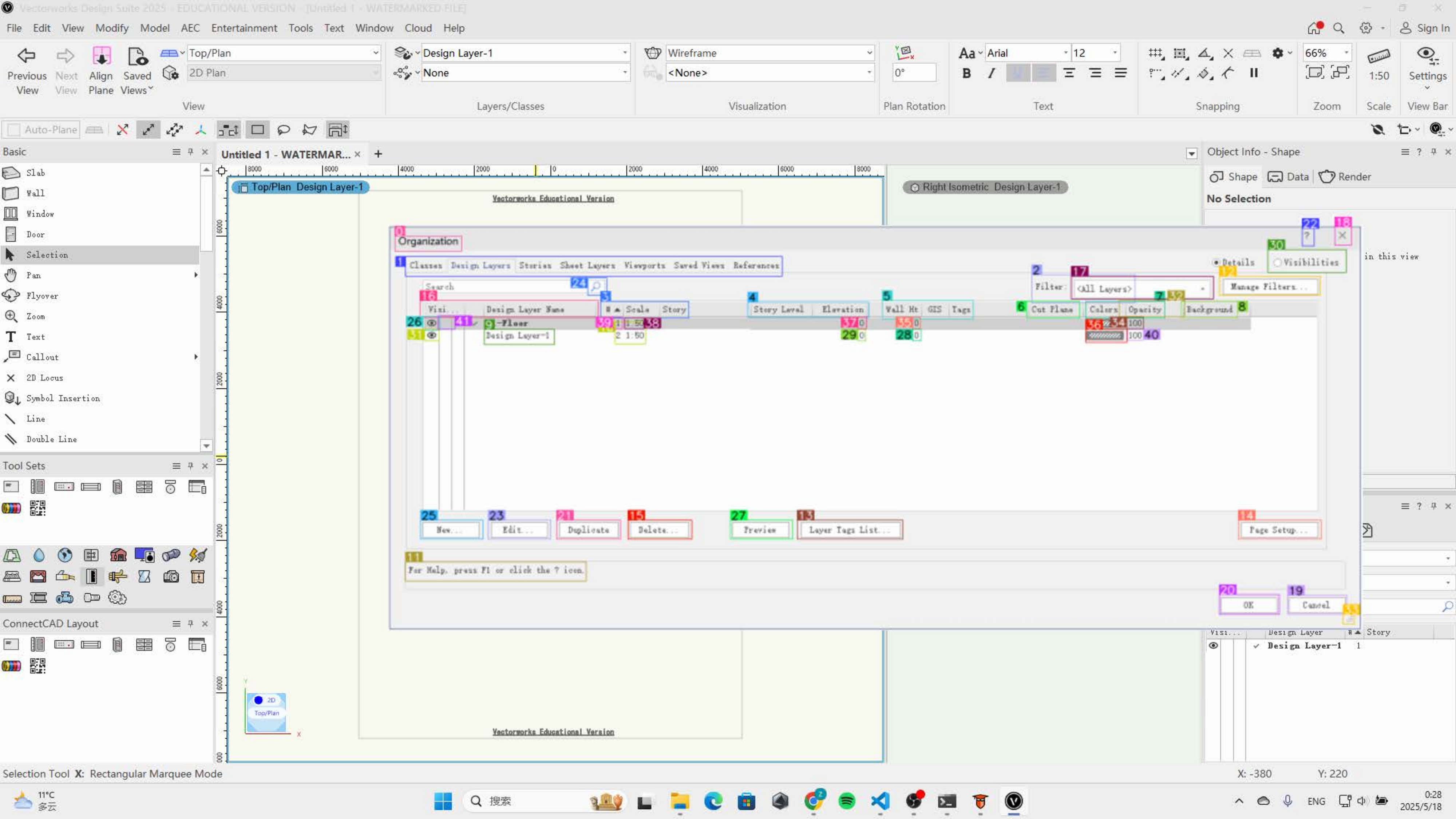}
    \caption{%
      \parbox[c][4\baselineskip][c]{\linewidth}{%
        \centering
        Confirm\\
        \texttt{press\_enter()}\\
         --
      }%
    }
    \label{fig:trajectory-e}
  \end{subfigure}\hfill
  \begin{subfigure}[b]{0.32\textwidth}
    \centering
    \includegraphics[width=\linewidth,keepaspectratio]{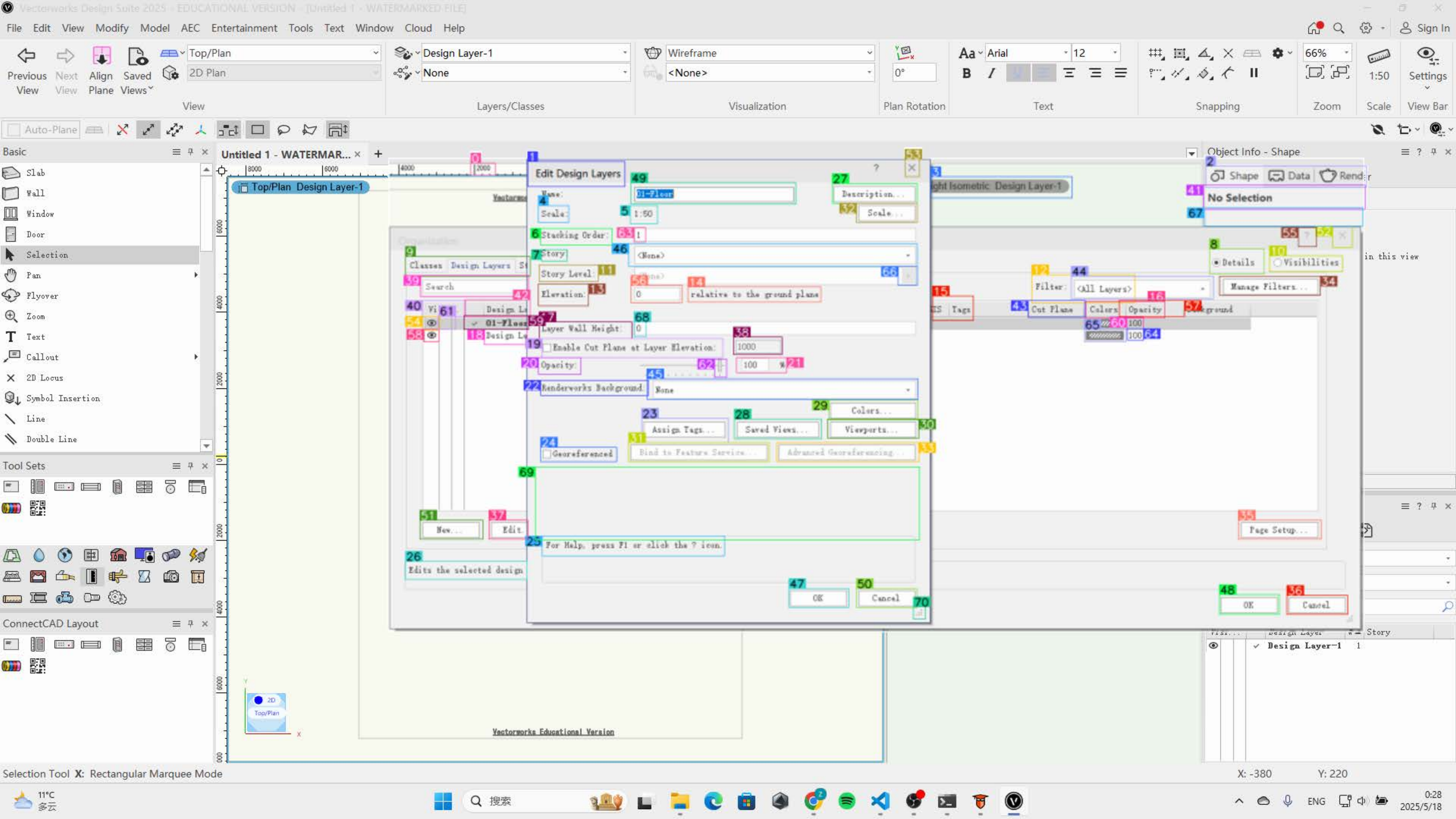}
    \caption{%
      \parbox[c][4\baselineskip][c]{\linewidth}{%
        \centering
        Click ‘Edit…’\\
        \texttt{move\_mouse\_to(688,698)}, \texttt{left\_click()}
      }%
    }
    \label{fig:trajectory-f}
  \end{subfigure}

  \vspace{0.5em}

  % 第三行
  \begin{subfigure}[b]{0.32\textwidth}
    \centering
    \includegraphics[width=\linewidth,keepaspectratio]{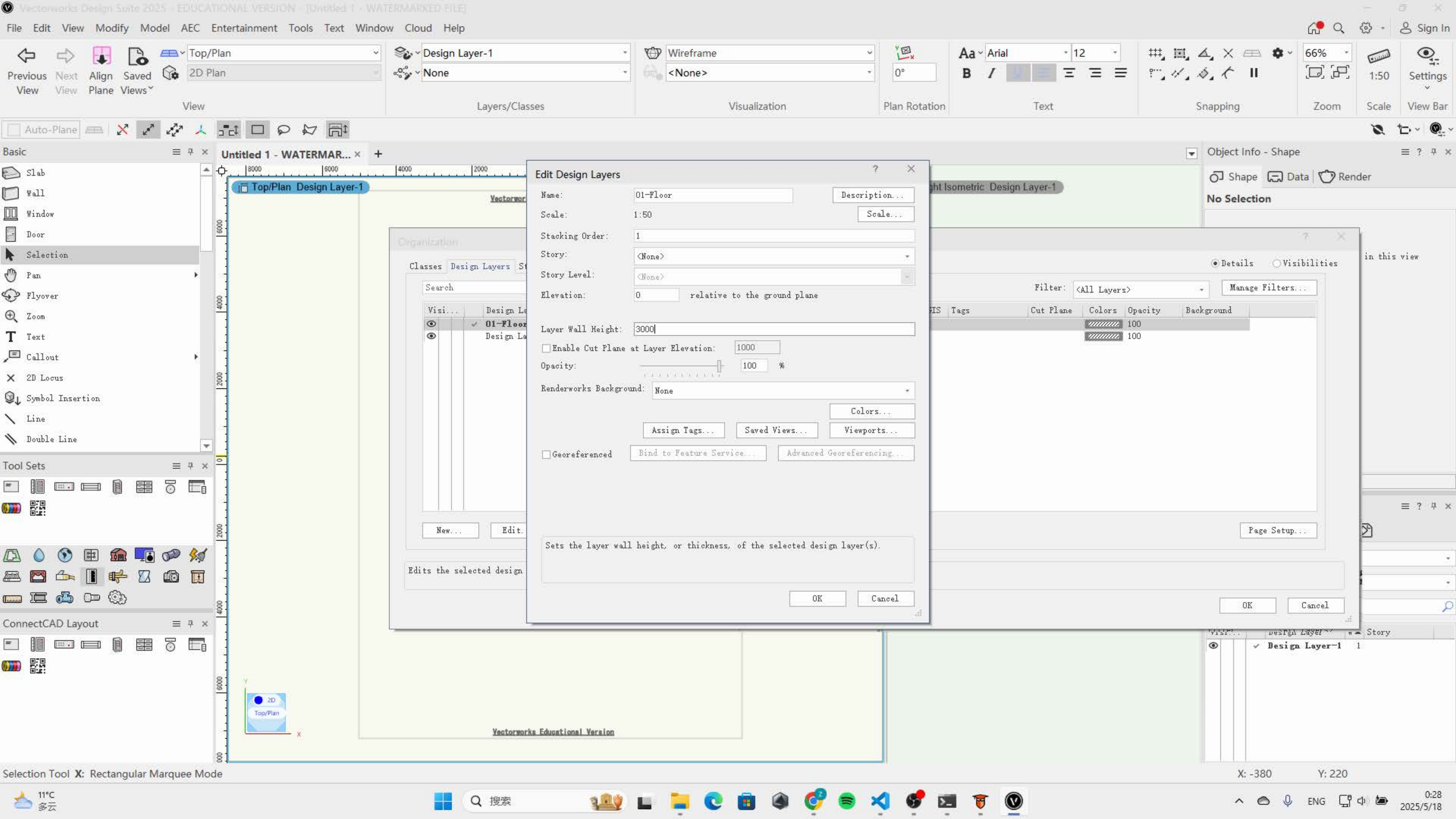}
    \caption{%
      \parbox[c][4\baselineskip][c]{\linewidth}{%
        \centering
        Edit elevation\\
        \texttt{move\_mouse\_to(866,436)}, \texttt{left\_click()},\\
        \texttt{select\_all()}, \texttt{type\_name("3000")}
      }%
    }
    \label{fig:trajectory-g}
  \end{subfigure}\hfill
  \begin{subfigure}[b]{0.32\textwidth}
    \centering
    \includegraphics[width=\linewidth,keepaspectratio]{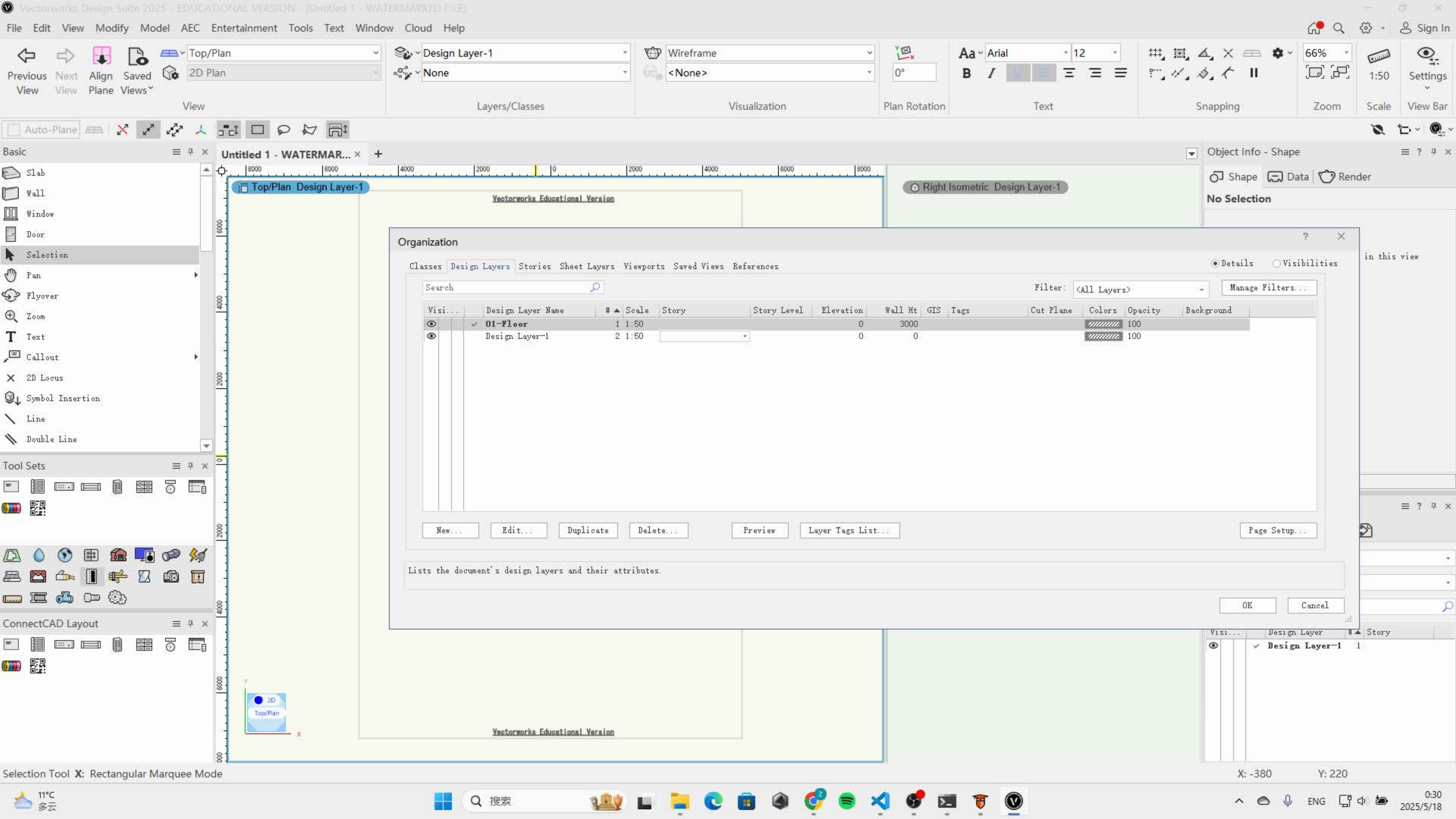}
    \caption{%
      \parbox[c][4\baselineskip][c]{\linewidth}{%
        \centering
        Confirm settings\\
        \texttt{press\_enter()}\\
         --\\
          --
      }%
    }
    \label{fig:trajectory-h}
  \end{subfigure}\hfill
  \begin{subfigure}[b]{0.32\textwidth}
    \centering
    \includegraphics[width=\linewidth,keepaspectratio]{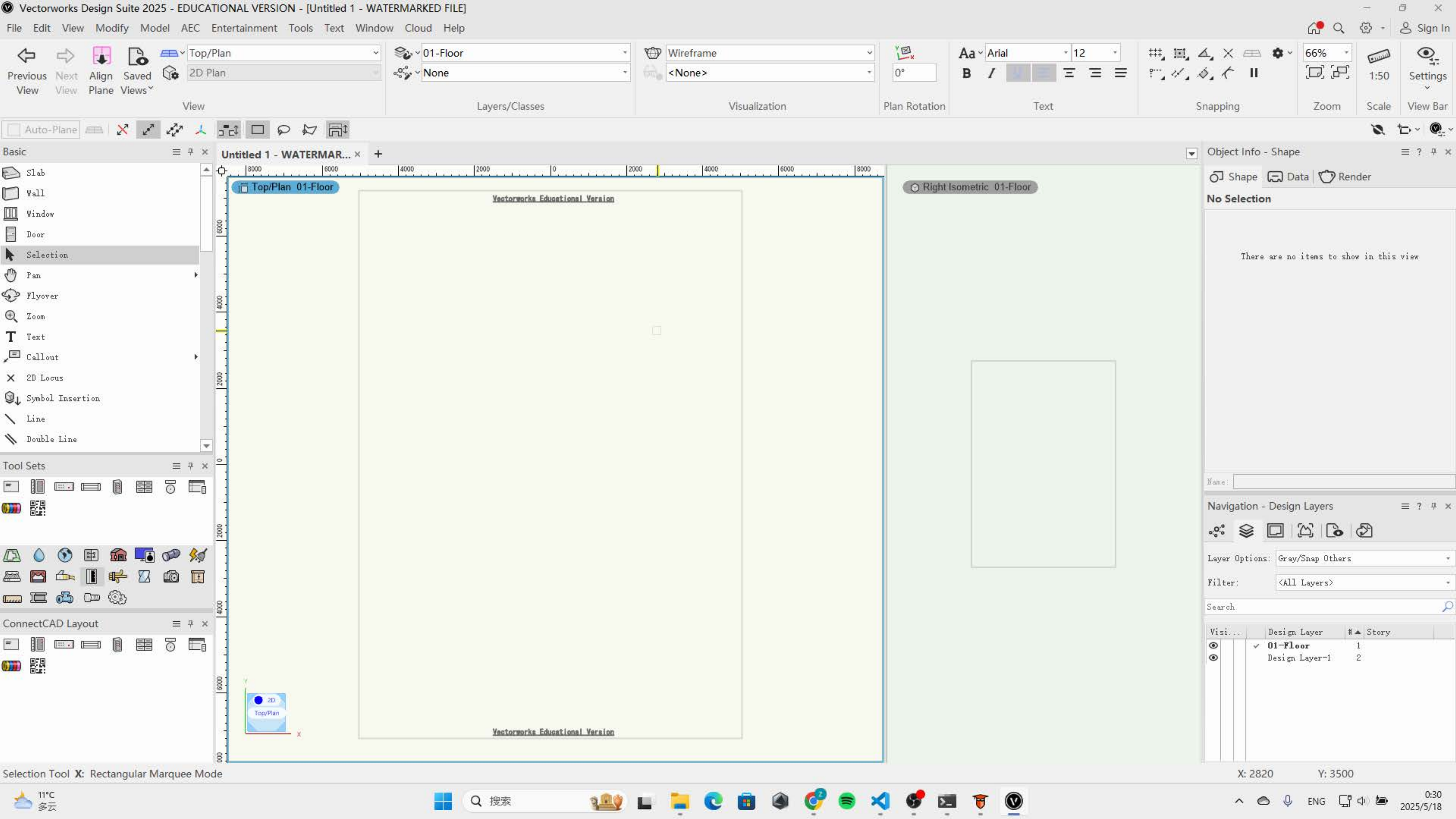}
    \caption{%
      \parbox[c][4\baselineskip][c]{\linewidth}{%
        \centering
        Final confirmation\\
        \texttt{press\_enter()}\\
         --\\
          --
      }%
    }
    \label{fig:trajectory-i}
  \end{subfigure}

  \caption{Screenshots of Floorplan Design and Design Layer creation actions.}
  \label{fig:desingandlayer111}
\end{figure}

\begin{figure}[htbp]
  \centering

  % 第一行
  \begin{subfigure}[b]{0.32\textwidth}
    \centering
    \includegraphics[width=\linewidth,keepaspectratio]{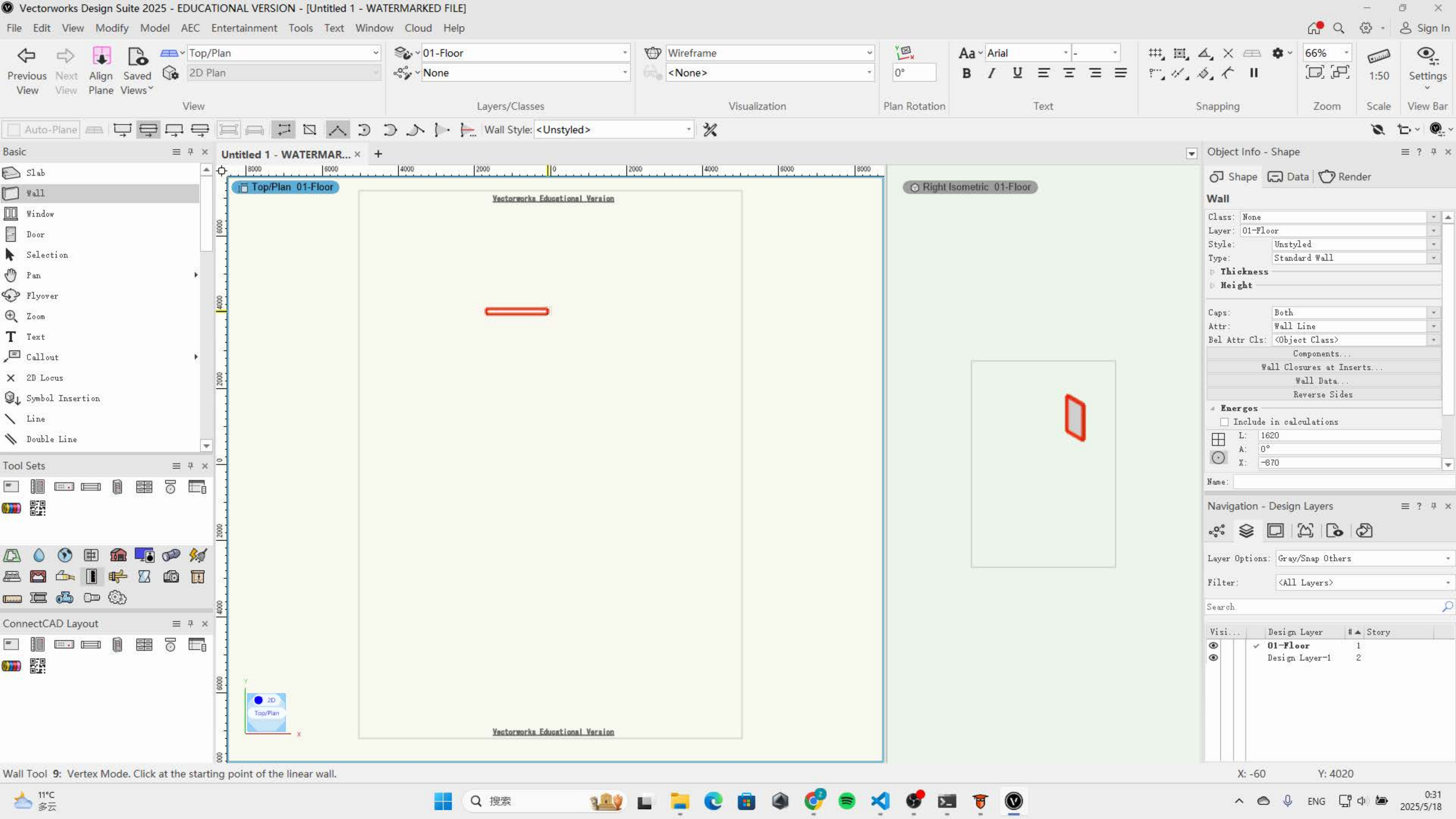}
    \caption{%
      \parbox[c][7\baselineskip][c]{\linewidth}{%
        \centering
        First external wall\\
        \texttt{shortcut(combo='9')}\\
        \texttt{move\_mouse\_to(x=641,y=410)}\\
        \texttt{left\_click()}\\
        \texttt{move\_mouse\_to(x=722,y=410)}\\
        \texttt{left\_click()}\\
        \texttt{press\_enter()}
      }%
    }
    \label{fig:trajectory-a}
  \end{subfigure}\hfill
  \begin{subfigure}[b]{0.32\textwidth}
    \centering
    \includegraphics[width=\linewidth,keepaspectratio]{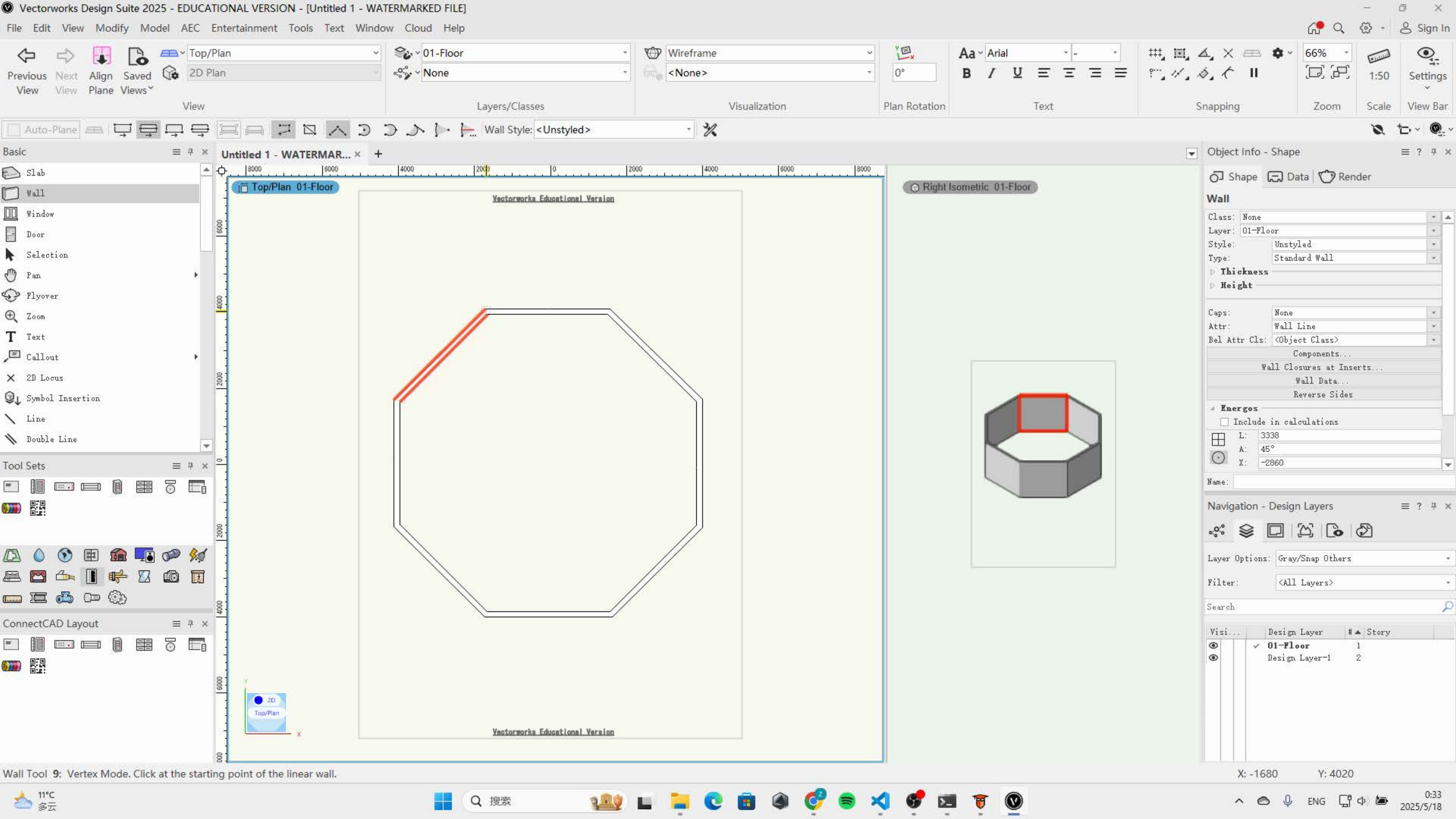}
    \caption{%
      \parbox[c][7\baselineskip][c]{\linewidth}{%
        \centering
        Last external wall\\
        \texttt{shortcut(combo='9')}\\
        \texttt{move\_mouse\_to(x=522,y=529)}\\
        \texttt{left\_click()}\\
        \texttt{move\_mouse\_to(x=641,y=410)}\\
        \texttt{left\_click()}\\
        \texttt{press\_enter()}
      }%
    }
    \label{fig:trajectory-b}
  \end{subfigure}\hfill
  \begin{subfigure}[b]{0.32\textwidth}
    \centering
    \includegraphics[width=\linewidth,keepaspectratio]{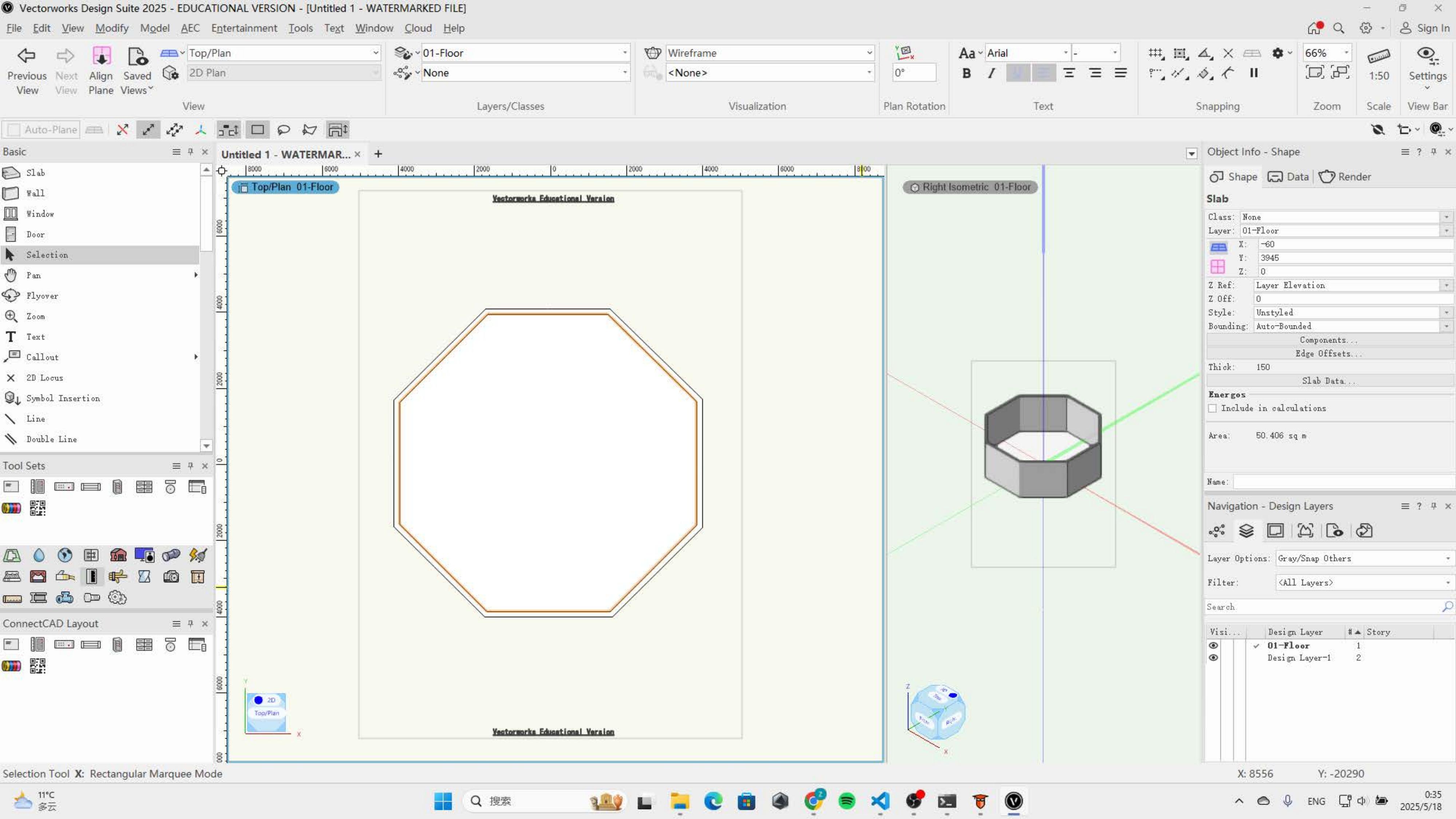}
    \caption{%
      \parbox[c][7\baselineskip][c]{\linewidth}{%
        \centering
        Create slab by picking external walls\\
        \texttt{shortcut(combo='alt+shift+2')}\\
        \texttt{move\_mouse\_to(x=681,y=410)}\\
        \texttt{left\_click()}\\
        …\\
        \texttt{left\_click()}\\
        \texttt{press\_enter()}
      }%
    }
    \label{fig:trajectory-c}
  \end{subfigure}

  \vspace{0.5em}

  % 第二行
  \begin{subfigure}[b]{0.32\textwidth}
    \centering
    \includegraphics[width=\linewidth,keepaspectratio]{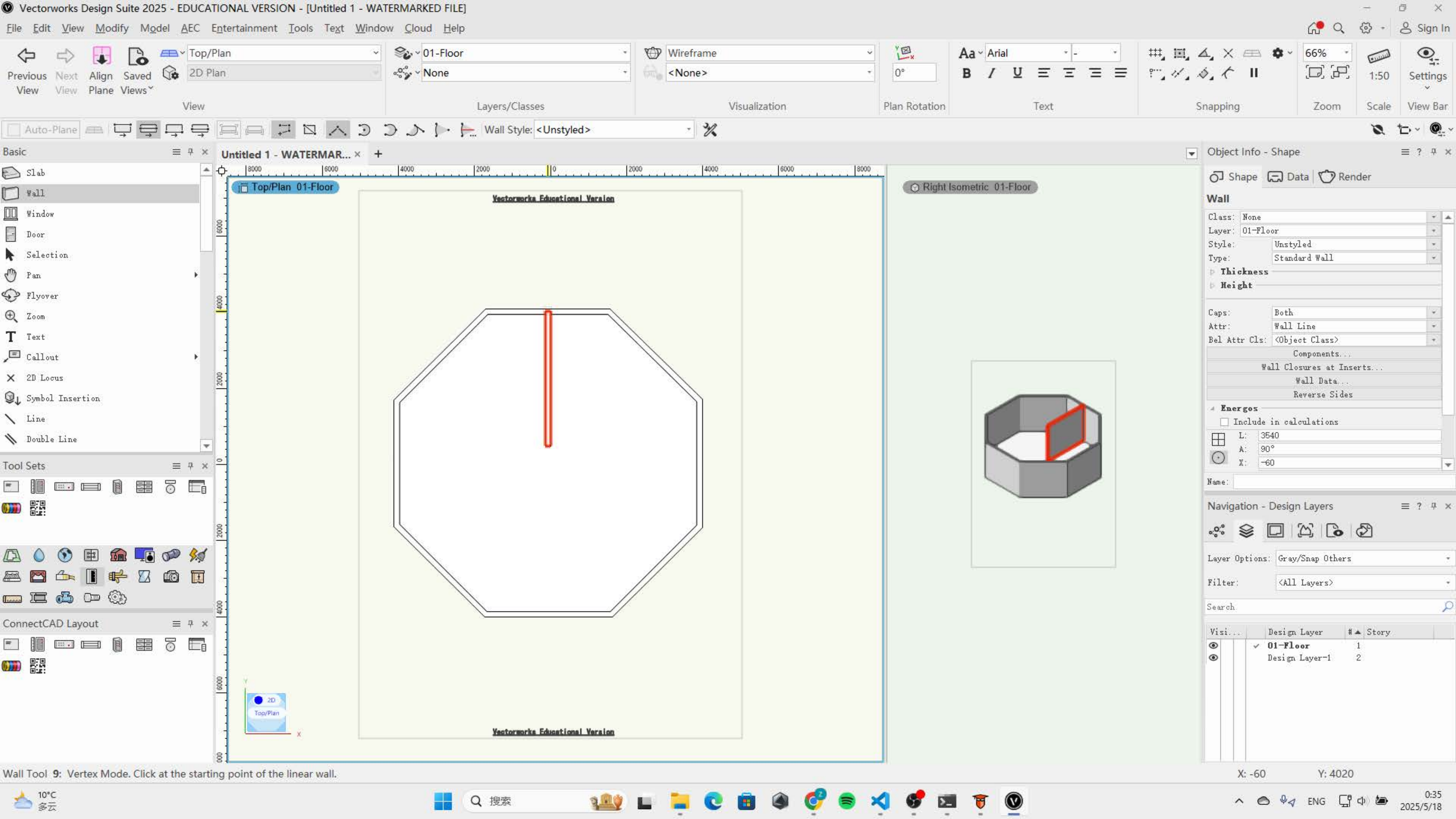}
    \caption{%
      \parbox[c][7\baselineskip][c]{\linewidth}{%
        \centering
        Create first internal wall\\
        \texttt{shortcut(combo='9')}\\
        \texttt{move\_mouse\_to(x=722,y=588)}\\
        \texttt{left\_click()}\\
        \texttt{move\_mouse\_to(x=722,y=410)}\\
        \texttt{left\_click()}\\
        \texttt{press\_enter()}
      }%
    }
    \label{fig:trajectory-d}
  \end{subfigure}\hfill
  \begin{subfigure}[b]{0.32\textwidth}
    \centering
    \includegraphics[width=\linewidth,keepaspectratio]{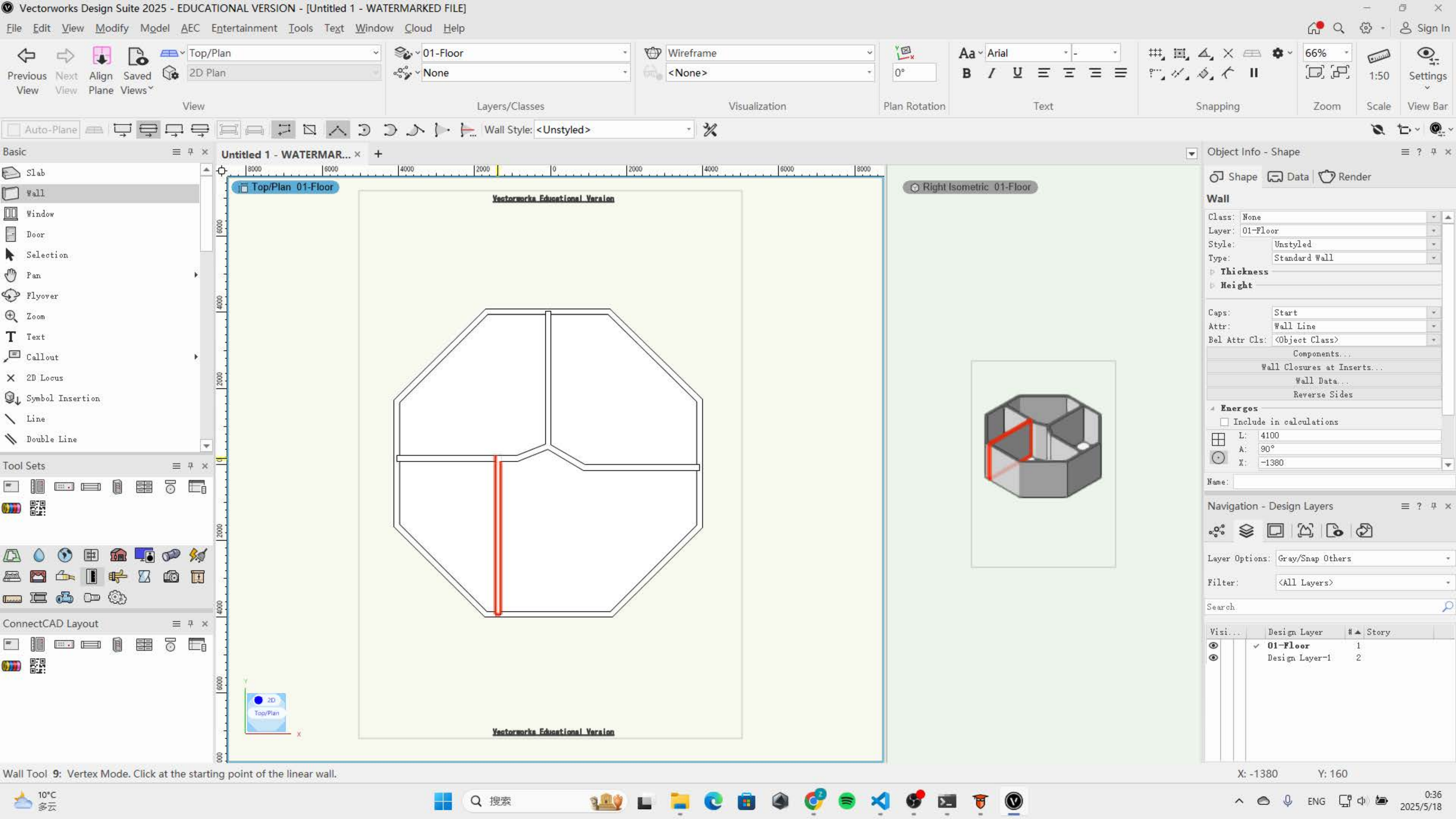}
    \caption{%
      \parbox[c][7\baselineskip][c]{\linewidth}{%
        \centering
        Final internal wall\\
        \texttt{shortcut(combo='9')}\\
        \texttt{move\_mouse\_to(x=656,y=810)}\\
        \texttt{left\_click()}\\
        \texttt{move\_mouse\_to(x=656,y=604)}\\
        \texttt{left\_click()}\\
        \texttt{press\_enter()}
      }%
    }
    \label{fig:trajectory-e}
  \end{subfigure}\hfill
  \begin{subfigure}[b]{0.32\textwidth}
    \centering
    \includegraphics[width=\linewidth,keepaspectratio]{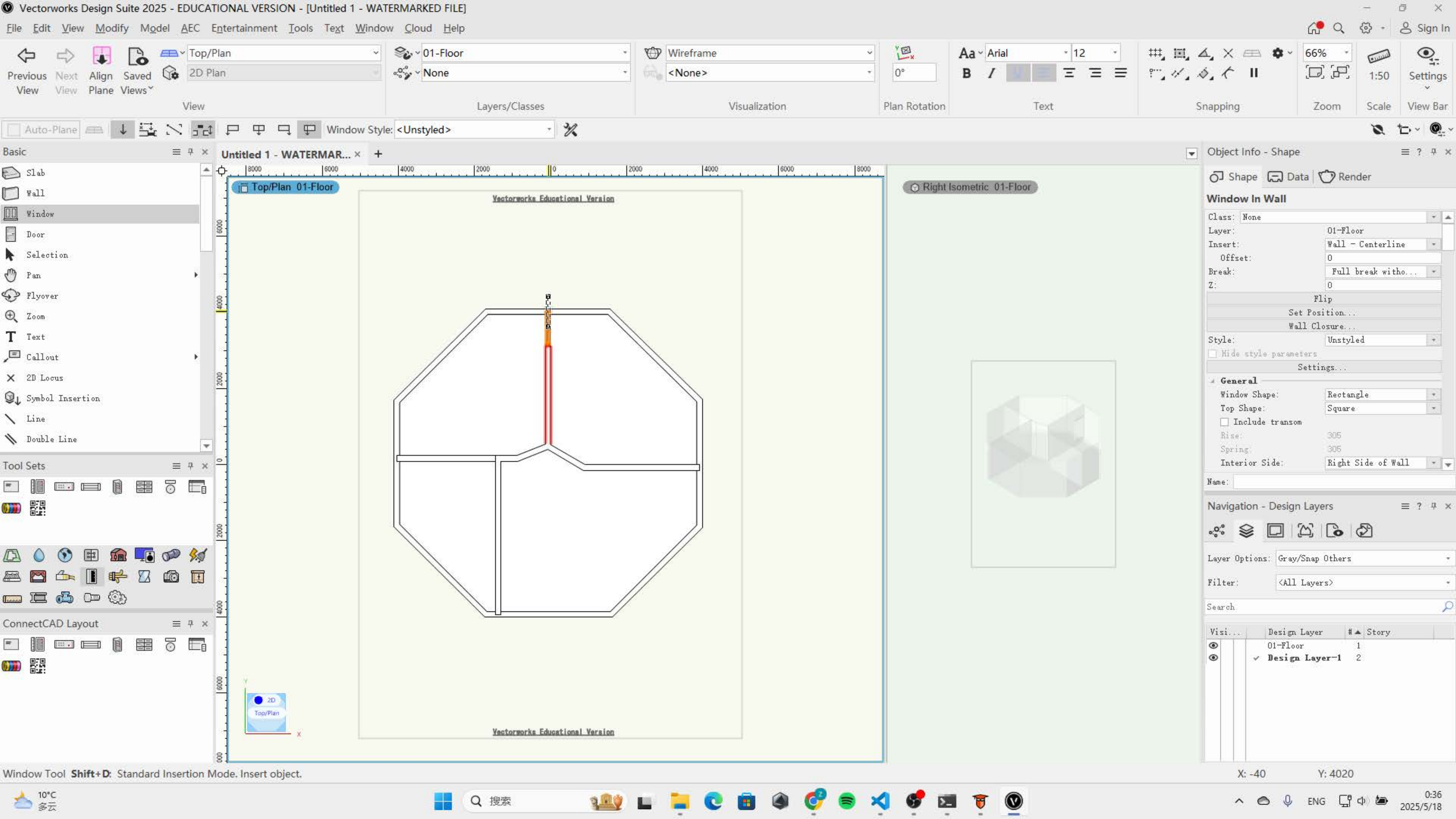}
    \caption{%
      \parbox[c][7\baselineskip][c]{\linewidth}{%
        \centering
        Insert first window\\
        \texttt{shortcut(combo='shift + d')}\\
        \texttt{move\_mouse\_to(x=723,y=410)}\\
        \texttt{left\_click()}\\
        \texttt{press\_enter()}\\
        --\\
        --
      }%
    }
    \label{fig:trajectory-f}
  \end{subfigure}

  \vspace{0.5em}

  % 第三行
  \begin{subfigure}[b]{0.32\textwidth}
    \centering
    \includegraphics[width=\linewidth,keepaspectratio]{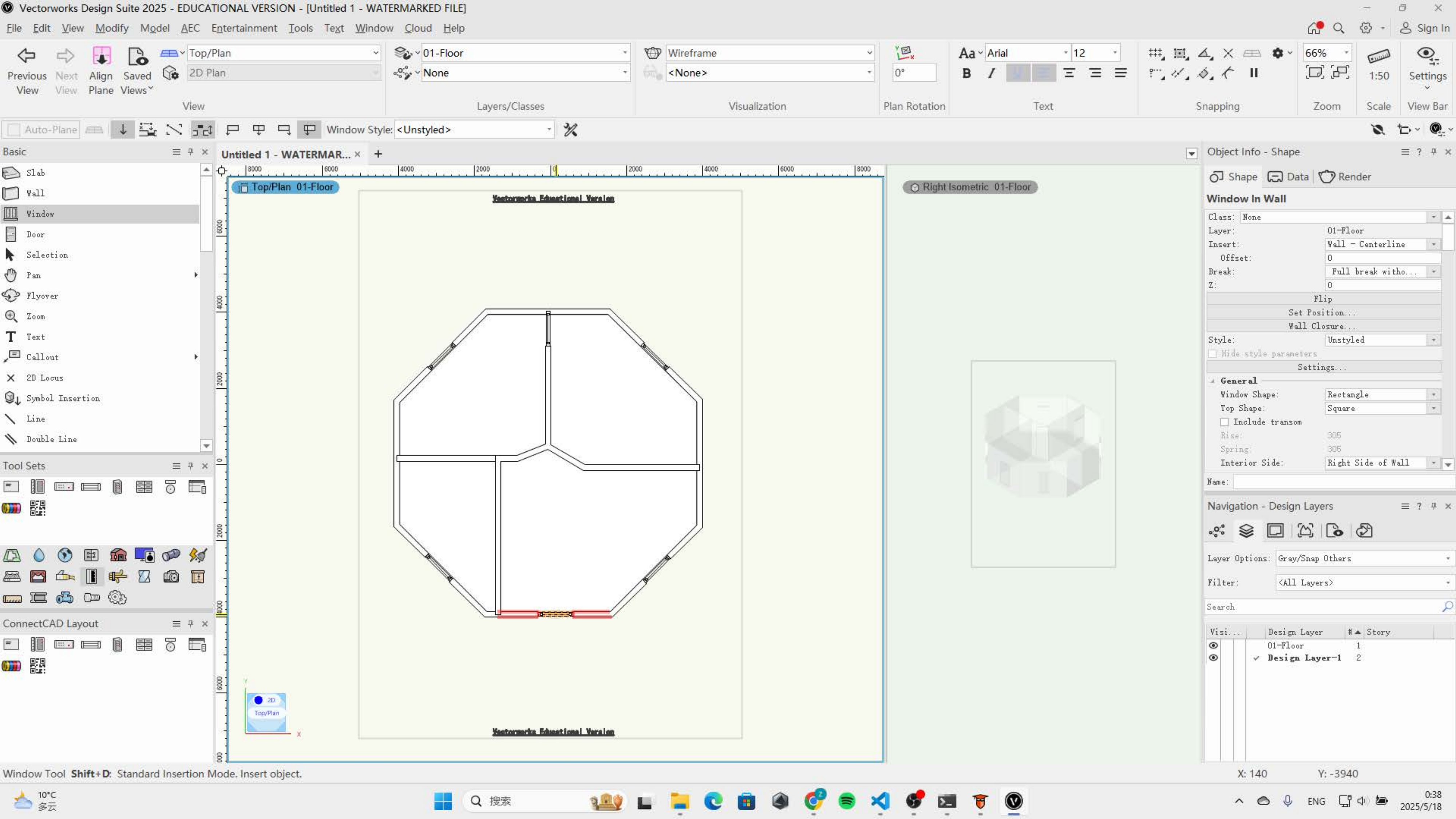}
    \caption{%
      \parbox[c][7\baselineskip][c]{\linewidth}{%
        \centering
        Insert final window\\
        \texttt{shortcut(combo='shift + d')}\\
        \texttt{move\_mouse\_to(x=732,y=810)}\\
        \texttt{left\_click()}\\
        \texttt{press\_enter()}\\
        --\\
        --
      }%
    }
    \label{fig:trajectory-g}
  \end{subfigure}\hfill
  \begin{subfigure}[b]{0.32\textwidth}
    \centering
    \includegraphics[width=\linewidth,keepaspectratio]{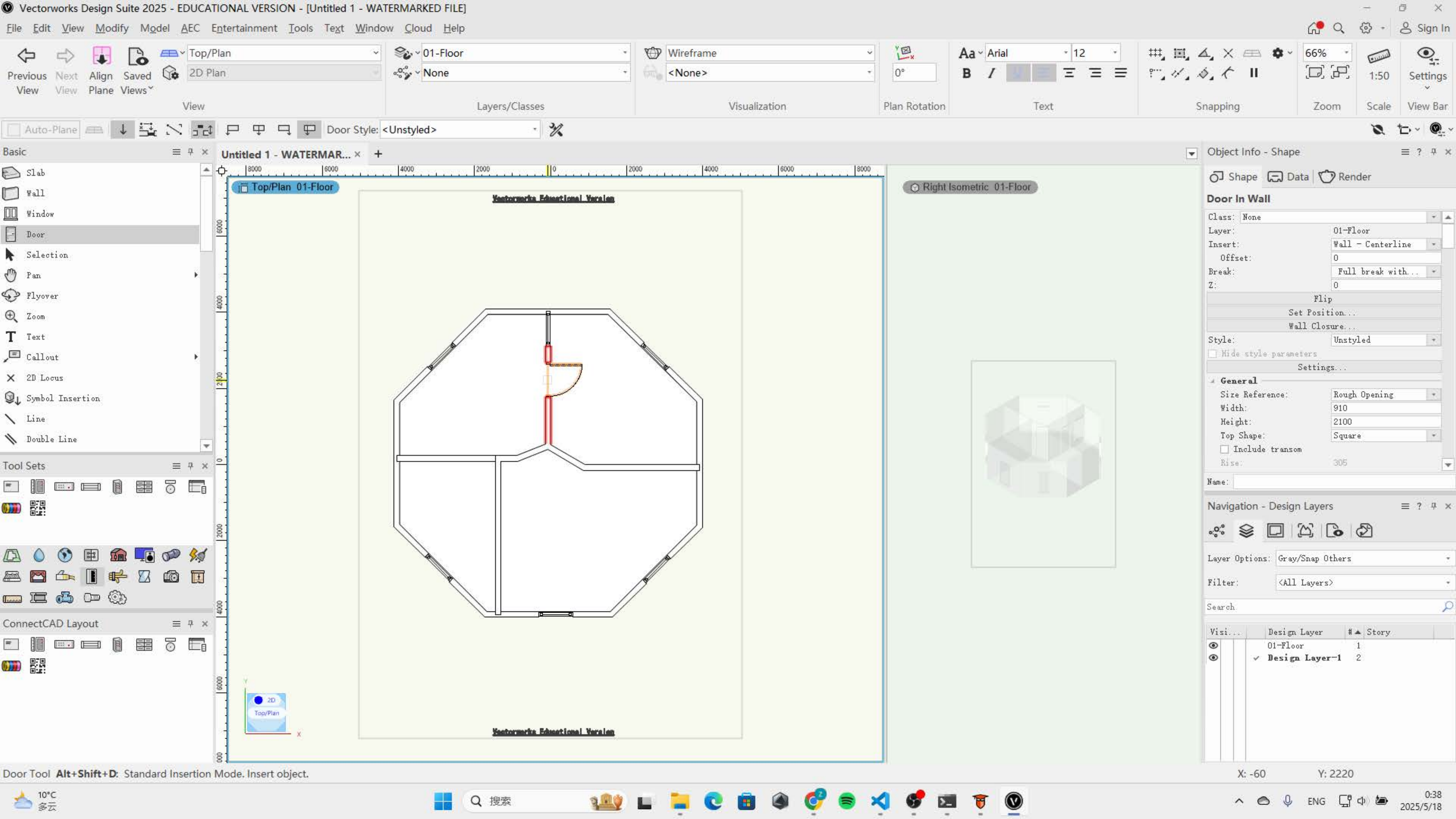}
    \caption{%
      \parbox[c][7\baselineskip][c]{\linewidth}{%
        \centering
        Insert first door\\
        \texttt{shortcut(combo='alt+shift+d')}\\
        \texttt{move\_mouse\_to(x=722,y=501)}\\
        \texttt{left\_click()}\\
        \texttt{press\_enter()}\\
        --\\
        --
      }%
    }
    \label{fig:trajectory-h}
  \end{subfigure}\hfill
  \begin{subfigure}[b]{0.32\textwidth}
    \centering
    \includegraphics[width=\linewidth,keepaspectratio]{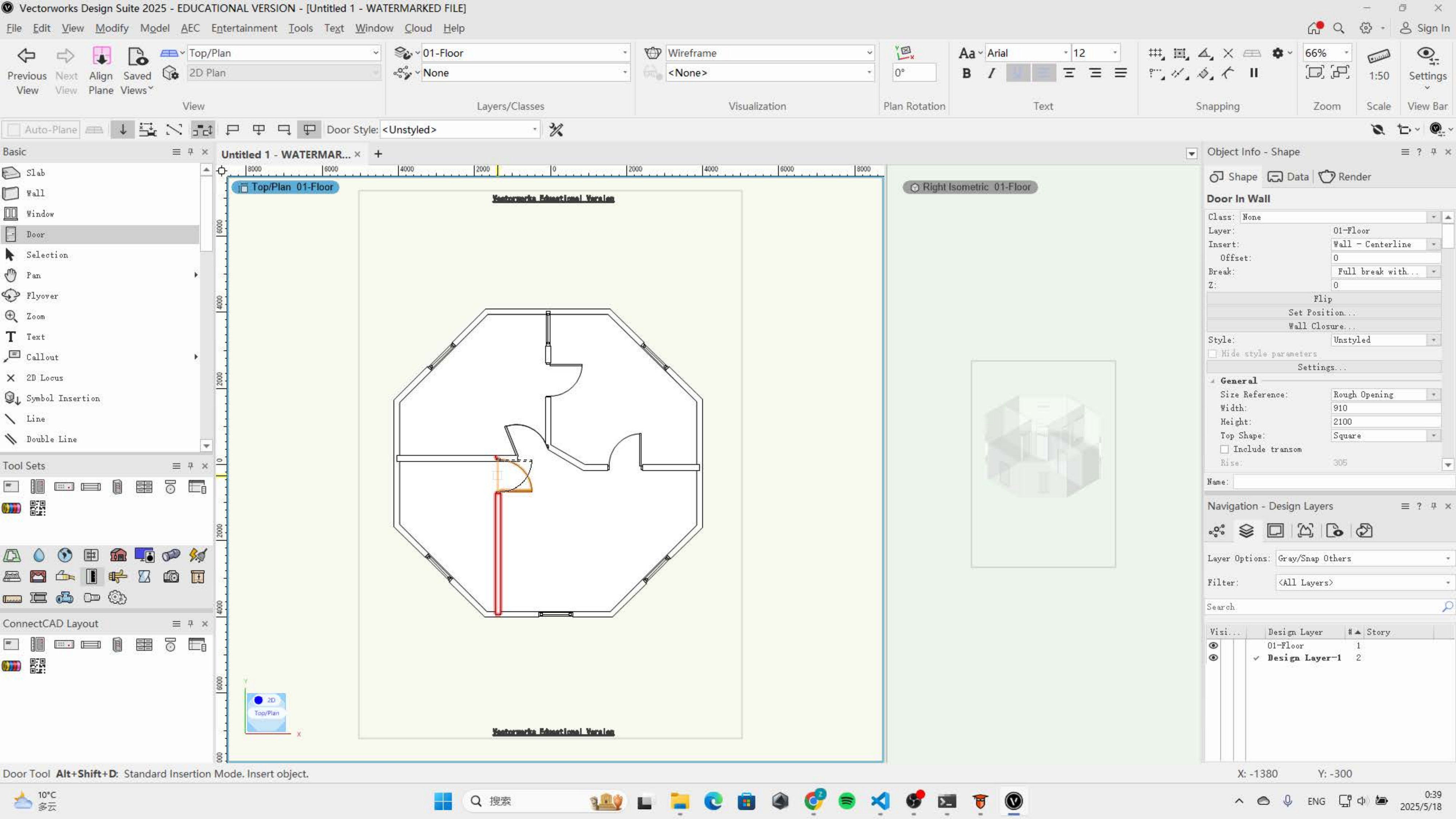}
    \caption{%
      \parbox[c][7\baselineskip][c]{\linewidth}{%
        \centering
        Insert final door\\
        \texttt{shortcut(combo='alt+shift+d')}
        \texttt{move\_mouse\_to(x=656,y=627)}\\
        \texttt{left\_click()}\\
        \texttt{press\_enter()}\\
        --\\
        --\\
      }%
    }
    \label{fig:trajectory-i}
  \end{subfigure}

  \caption{Screenshots of element creations.}
  \label{fig:elementcreation111}
\end{figure}

\begin{figure}[htbp]
  \centering

  % 第一行：两张图
  \begin{subfigure}[b]{0.45\textwidth}
    \centering
    \includegraphics[width=\linewidth,keepaspectratio]{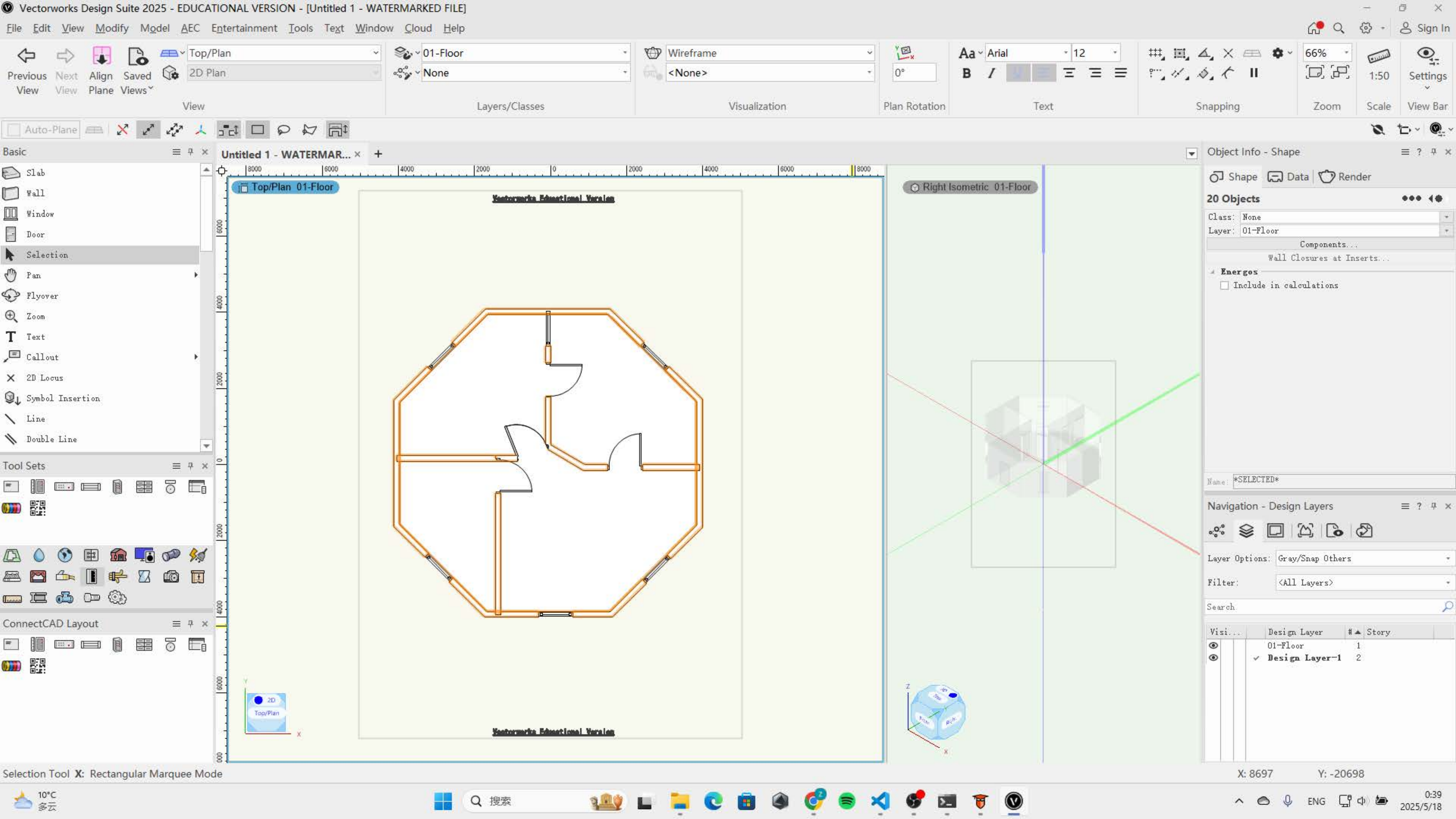}
    \caption{%
      \parbox[c][7\baselineskip][c]{\linewidth}{%
        \centering
        Select all components\\
        \texttt{select\_all()} \\
        --\\
        --
      }%
    }
    \label{fig:trajectory-a}
  \end{subfigure}\hfill
  \begin{subfigure}[b]{0.45\textwidth}
    \centering
    \includegraphics[width=\linewidth,keepaspectratio]{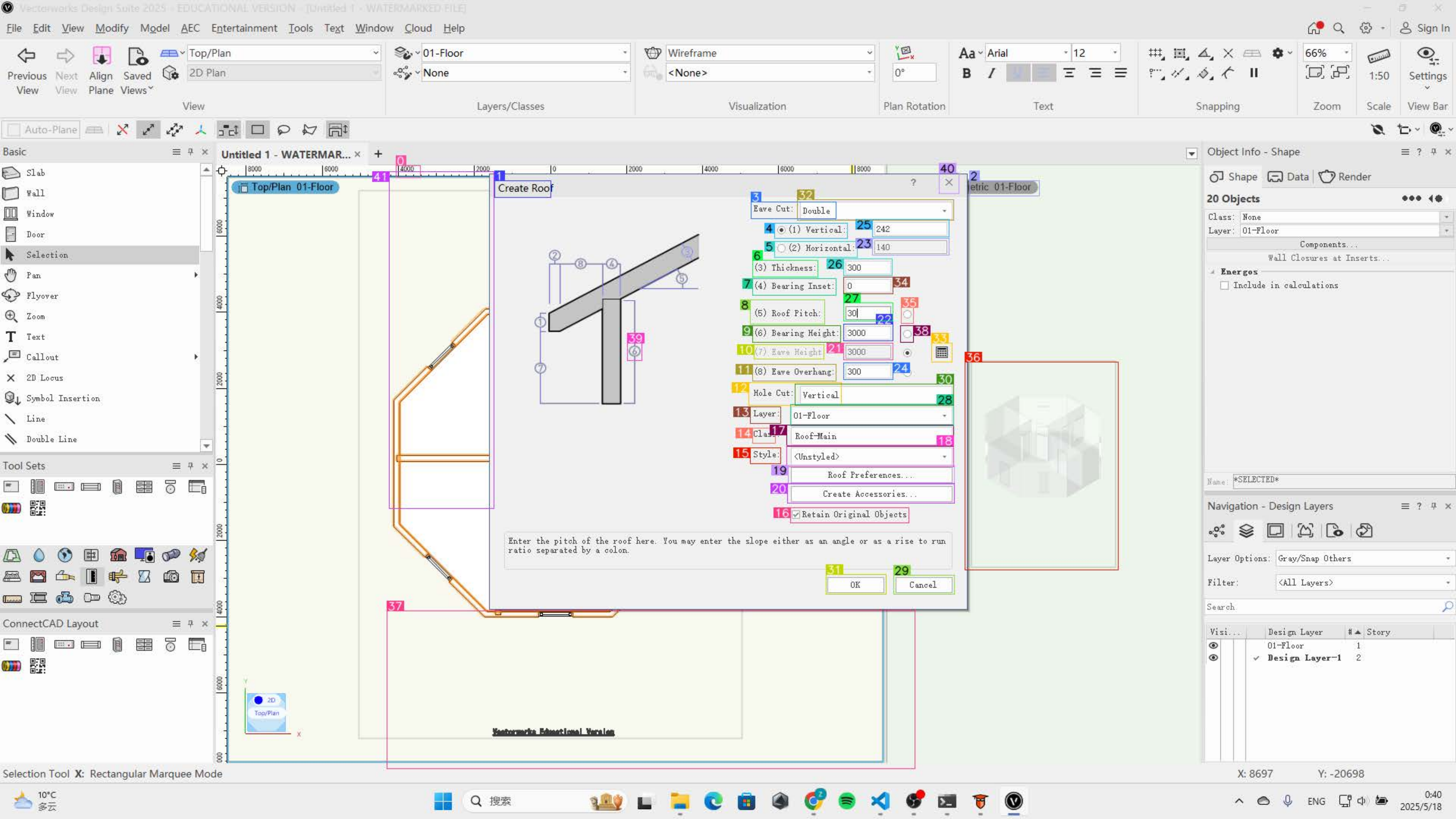}
    \caption{%
      \parbox[c][7\baselineskip][c]{\linewidth}{%
        \centering
        Active roof tool\\
        \texttt{shortcut(combo="ctrl + alt + shift + 1")}\\
        --\\
        --
      }%
    }
    \label{fig:trajectory-b}
  \end{subfigure}

  \vspace{0.5em}

  % 第二行：两张图
  \begin{subfigure}[b]{0.45\textwidth}
    \centering
    \includegraphics[width=\linewidth,keepaspectratio]{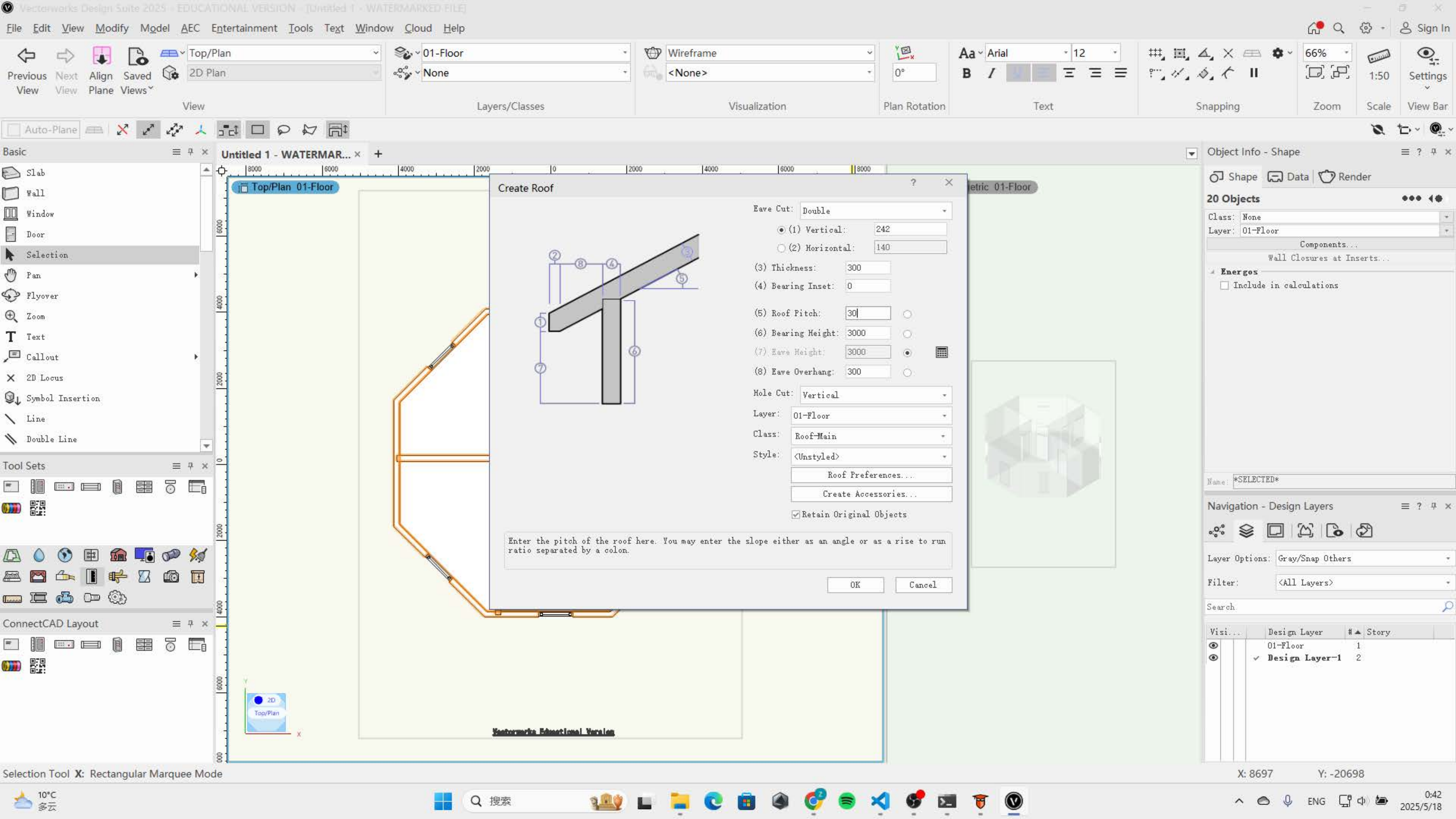}
    \caption{%
      \parbox[c][7\baselineskip][c]{\linewidth}{%
        \centering
        Set parameters\\
        \texttt{move\_mouse\_to(1145,353)}\\
        \texttt{left\_click()}\\
        \dots\\
        \texttt{type\_name("30")}
      }%
    }
    \label{fig:trajectory-c}
  \end{subfigure}\hfill
  \begin{subfigure}[b]{0.45\textwidth}
    \centering
    \includegraphics[width=\linewidth,keepaspectratio]{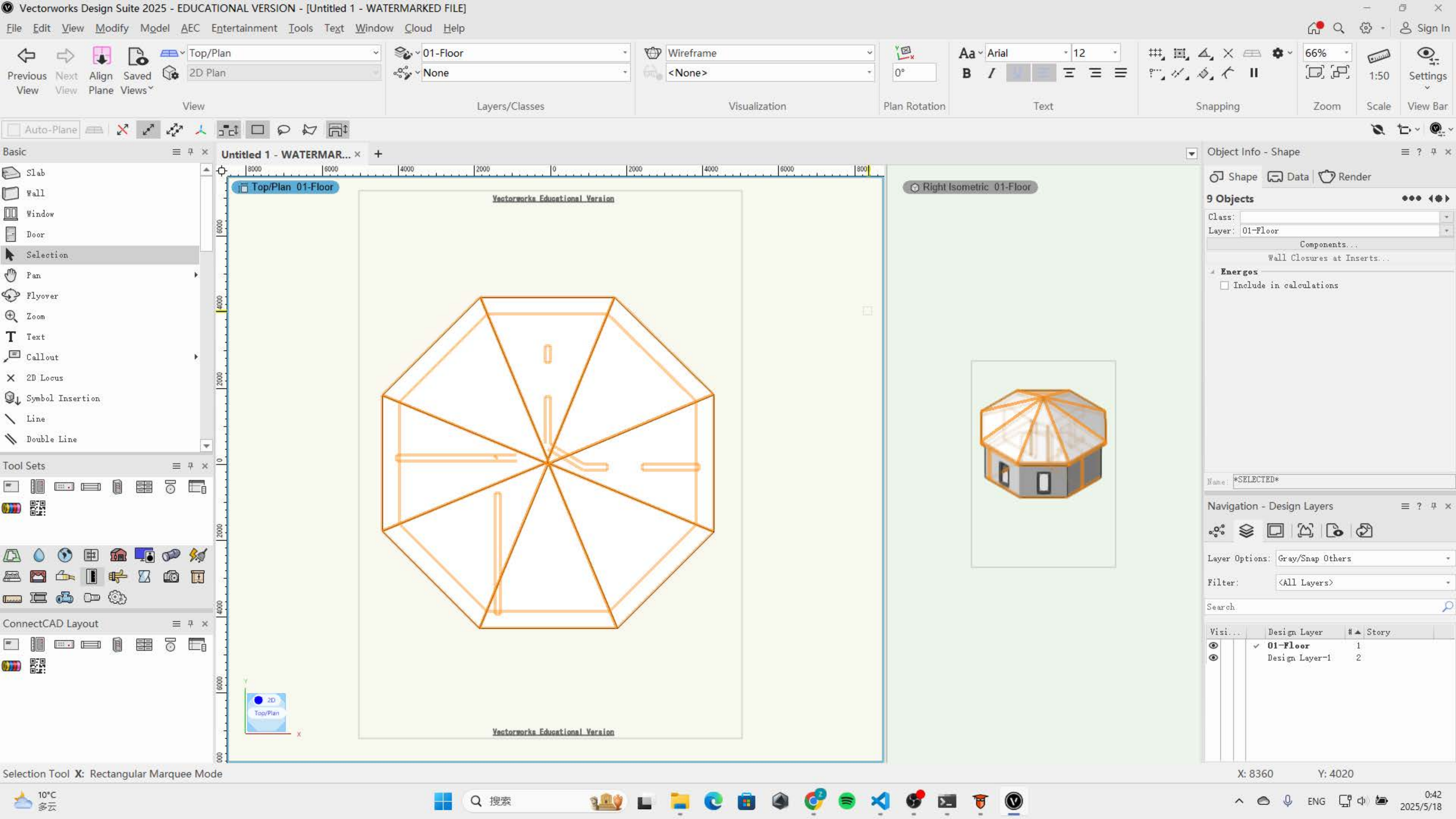}
    \caption{%
      \parbox[c][7\baselineskip][c]{\linewidth}{%
        \centering
        Confirm roof\\
        \texttt{press\_enter()}\\
        --\\
        --
      }%
    }
    \label{fig:trajectory-d}
  \end{subfigure}

  \caption{Screenshots of roof creation steps.}
  \label{fig:roof-creations111}
\end{figure}

Given the task: \textit{Generate a building model based on a hand-drawn octagon floorplan, modifying the interior layout to include four rooms instead of three.} The floorplan design and new design layer creation are illustrated in Figure~\ref{fig:desingandlayer111}, element creation is shown in Figure~\ref{fig:elementcreation111}, and roof creation is depicted in Figure~\ref{fig:roof-creations111}.

\newpage

\subsection{Failure Examples}
\begin{figure}[htbp]
  \centering

  % 三幅图同一行，顶部对齐
  \begin{subfigure}[t]{0.32\textwidth}
    \centering
    \parbox[c][5\baselineskip][c]{\linewidth}{%
    \centering\textit{\fcolorbox{red}{white}{Set the Elevation (Z) to 1000} and the layer wall height (\(\Delta Z\)) to the default (3000\,mm) for the first storey.}%
    }
    \caption{Planning info: Elevation (Z) is wrong.}
    \label{fig:trajectory-a}
  \end{subfigure}\hfill
  \begin{subfigure}[t]{0.32\textwidth}
    \centering
    \includegraphics[width=\linewidth,keepaspectratio]{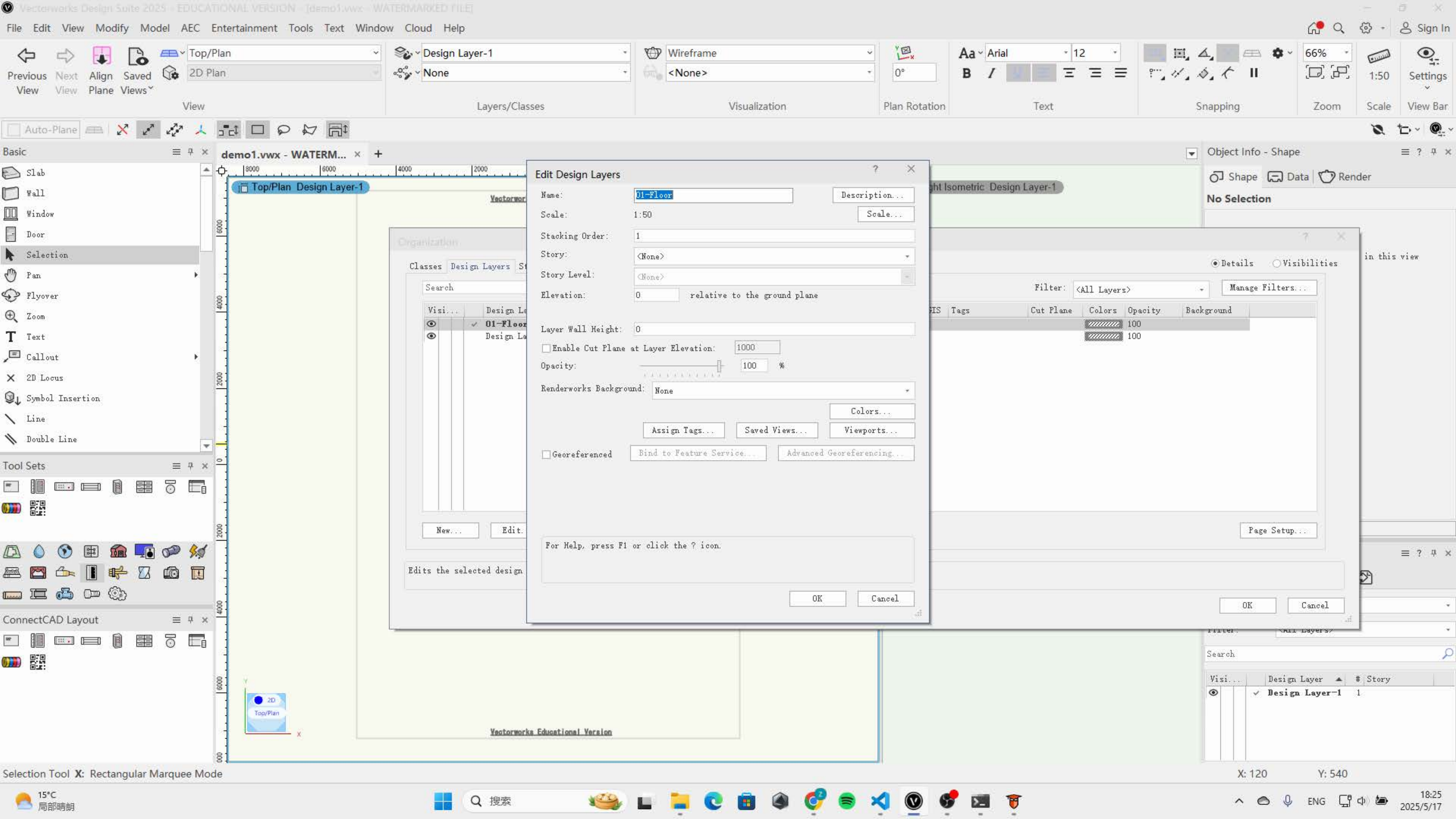}
    \caption{\parbox[c][5\baselineskip][c]{\linewidth}{\centering
        Click ‘Edit…’\\
      \texttt{move\_mouse\_to(684,698)}, \texttt{left\_click()}
    }}
    \label{fig:trajectory-b}
  \end{subfigure}\hfill
  \begin{subfigure}[t]{0.32\textwidth}
    \centering
    \includegraphics[width=\linewidth,keepaspectratio]{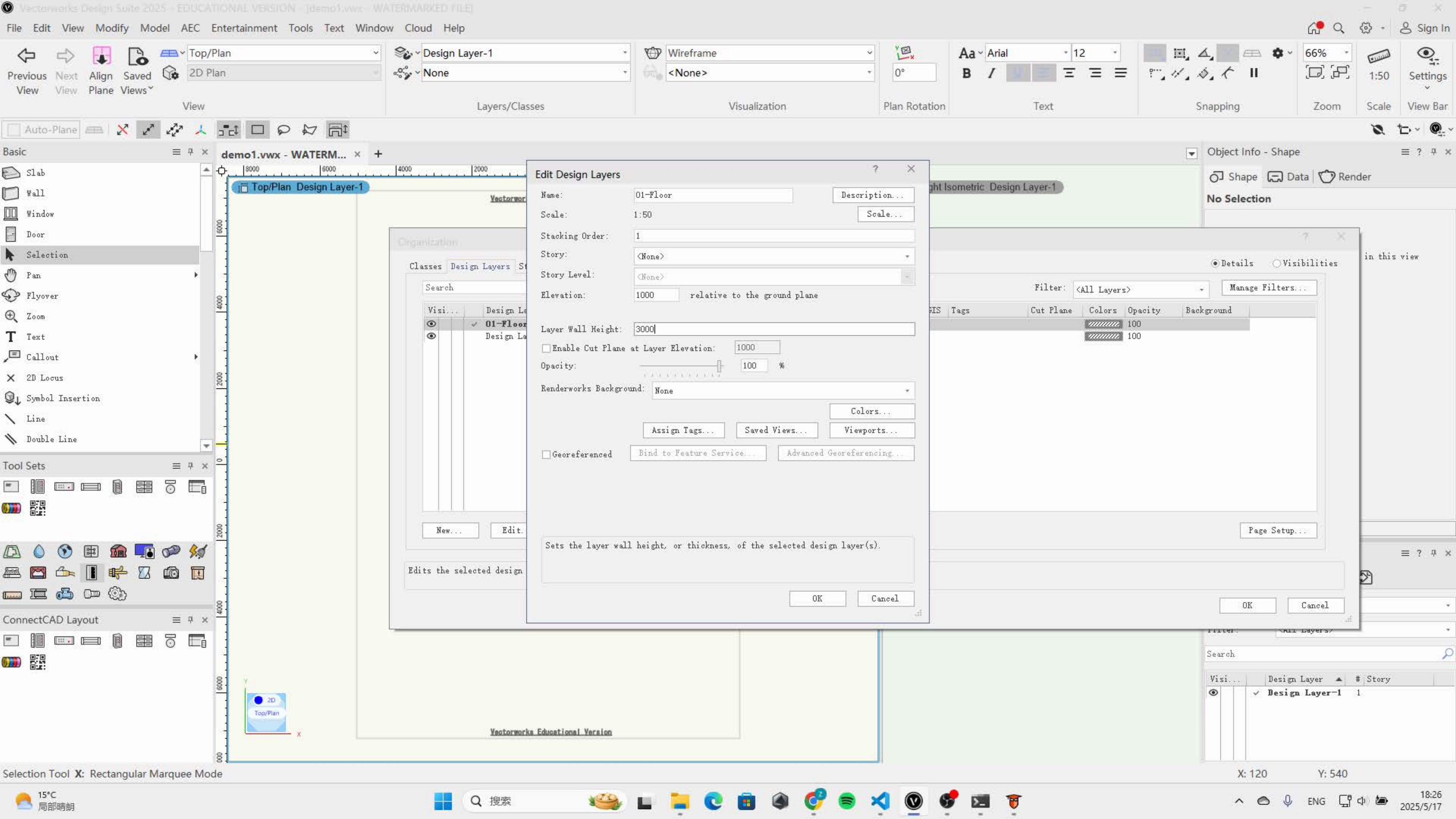}
    \caption{\parbox[c][5\baselineskip][c]{\linewidth}{\centering
        Set elevation\\
      \texttt{move\_mouse\_to(865,383)}, \texttt{left\_click()},\\
      \texttt{select\_all()}, \texttt{type\_name(1000)}\\
    }}
    \label{fig:trajectory-111c}
  \end{subfigure}

    \caption{A failed task caused by a planning error: the elevation (\textit{Z}) was incorrectly set to \texttt{1000} instead of \texttt{0}.}
    
  \label{fig:planningerro111r}
\end{figure}

\begin{figure}[htbp]
  \centering
  % 三幅图同一行
  \begin{subfigure}[b]{0.32\textwidth}
    \centering
    \includegraphics[width=\linewidth,keepaspectratio]{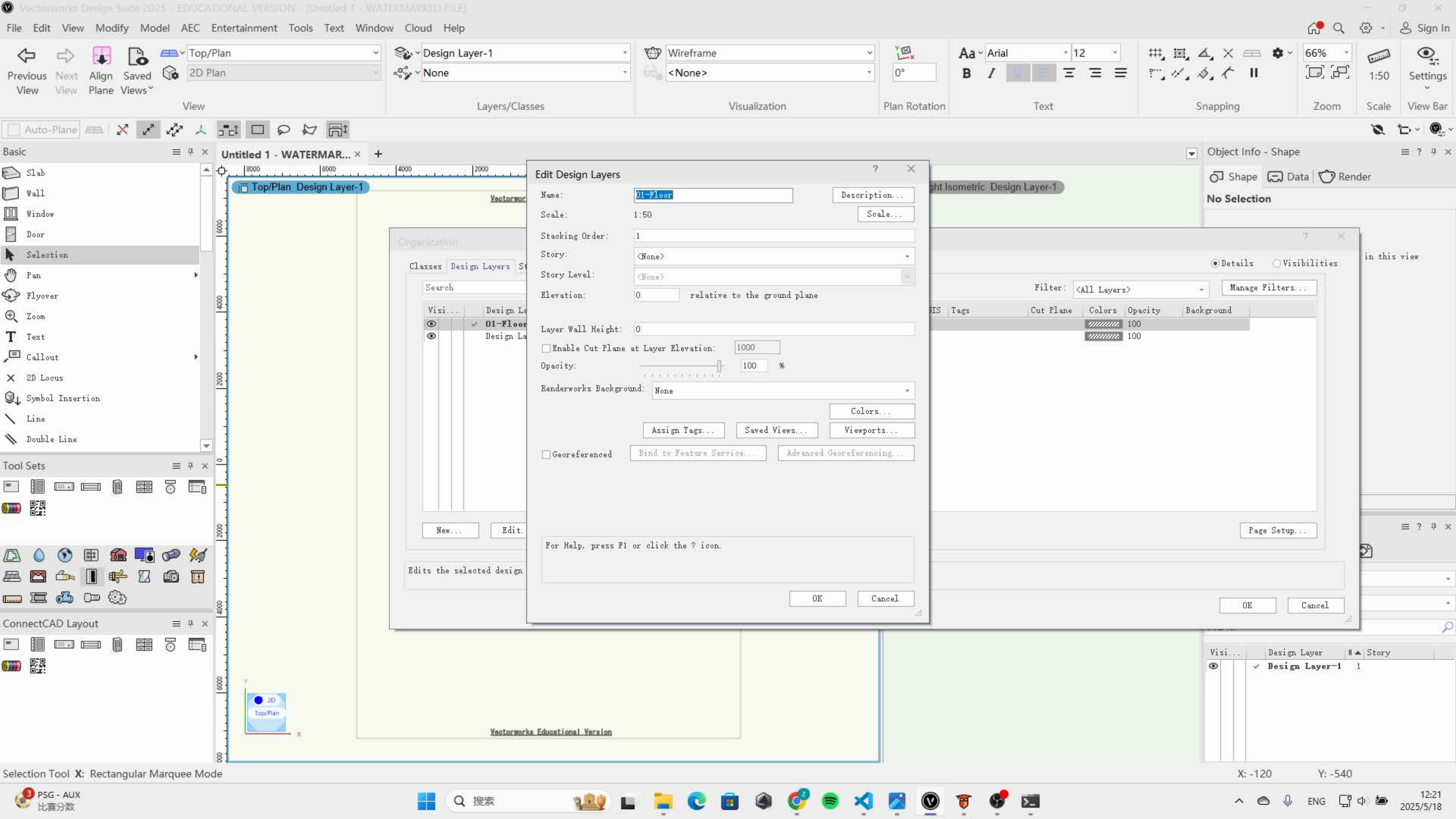}
    \caption{\parbox[c][6\baselineskip][c]{\linewidth}{\centering
      Click ‘Edit…’\\
      \texttt{move\_mouse\_to(684,698)}, \texttt{left\_click()}
    }}
    \label{fig:trajectory-a}
  \end{subfigure}\hfill
  \begin{subfigure}[b]{0.32\textwidth}
    \centering
    \includegraphics[width=\linewidth,keepaspectratio]{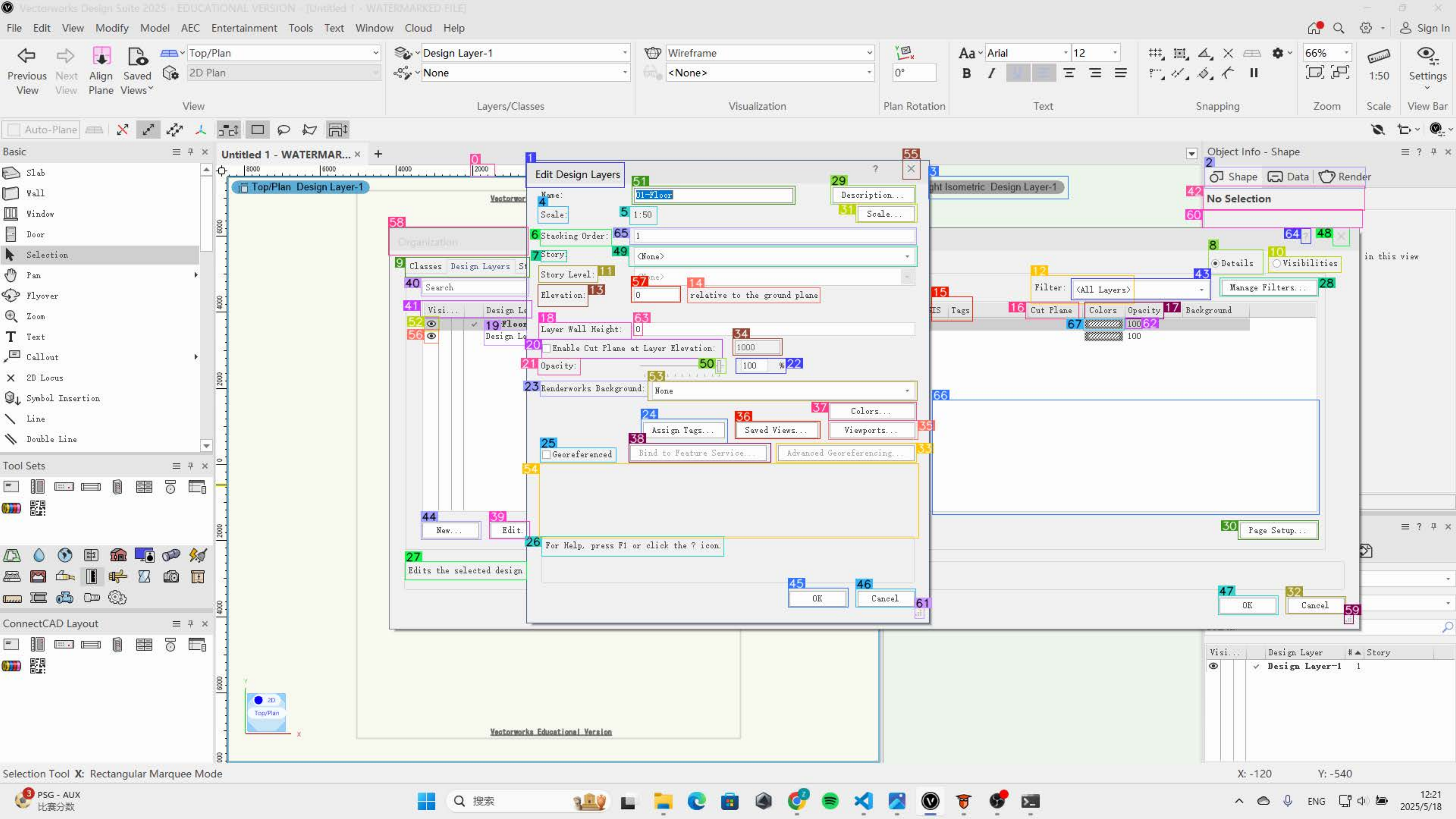}
    \caption{\parbox[c][6\baselineskip][c]{\linewidth}{\centering
      Grounding error\\
      Detect bounding box of 51 as elevation instead of 63
    }}
    \label{fig:trajectory-b}
  \end{subfigure}\hfill
  \begin{subfigure}[b]{0.32\textwidth}
    \centering
    \includegraphics[width=\linewidth,keepaspectratio]{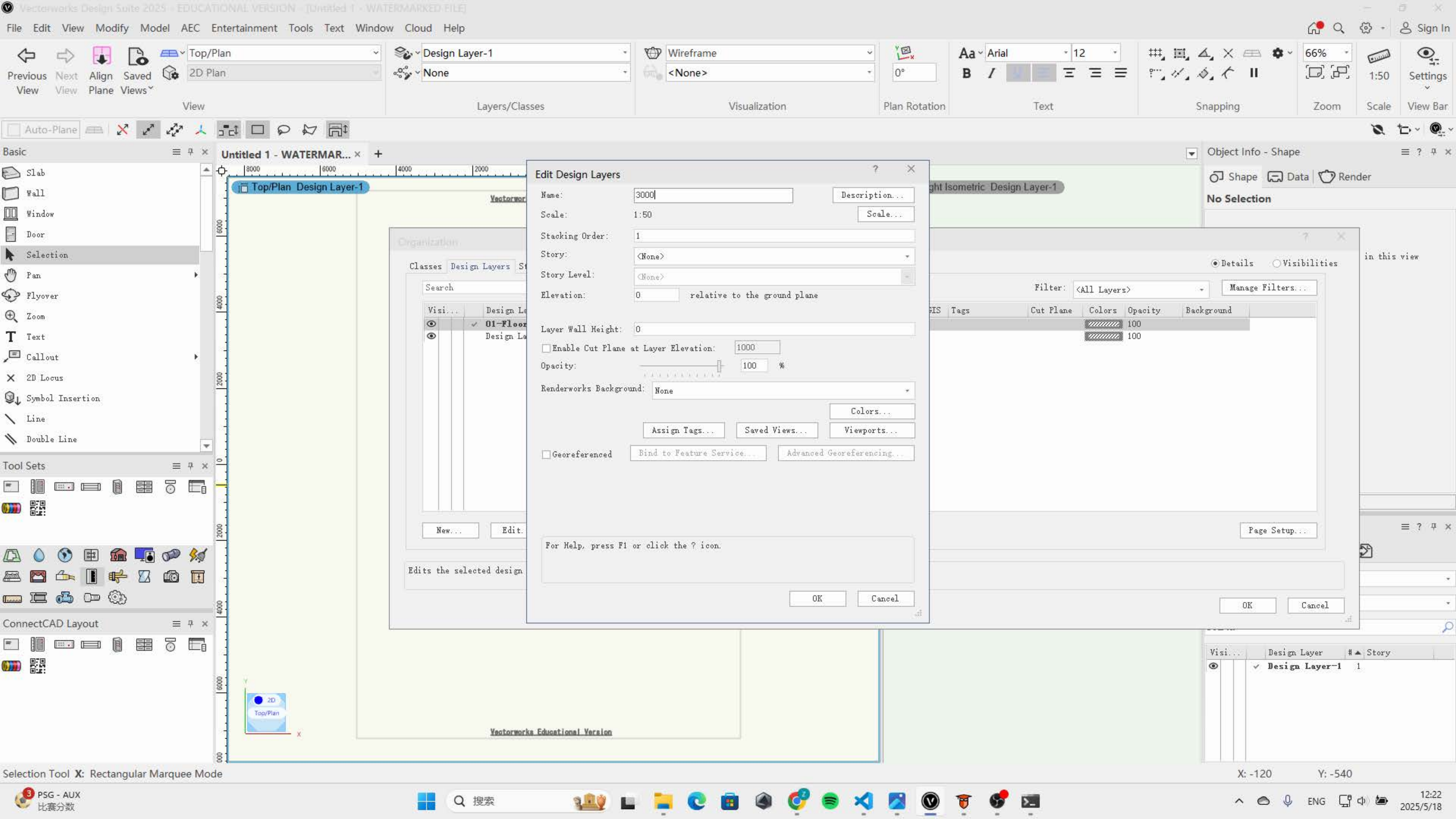}
    \caption{\parbox[c][6\baselineskip][c]{\linewidth}{\centering
      Edit parameters\\
      \texttt{move\_mouse\_to(950,445)}, \texttt{left\_click()},\\
      \texttt{select\_all()}, \texttt{type\_name("3000")}
    }}
    \label{fig:trajectory-c}
  \end{subfigure}

\caption{A failed task caused by a grounding error: the mouse clicked on bounding box 51 instead of 63, resulting in the \textbf{Name} being incorrectly changed to \texttt{3000}, while the \textbf{Elevation} value remained unchanged.}

  \label{fig:planningerror}
\end{figure}

\begin{figure}[htbp]
  \centering
  % 三幅图同一行，顶部对齐
  \begin{subfigure}[t]{0.32\textwidth}
    \centering
    % 子图说明文字
    \parbox[c][5\baselineskip][c]{\linewidth}{%
      \centering\textit{Create the main floor slab for the second floor, covering the area defined by the external walls.}%
    }
    \caption{
      Subtask: Create the slab for the second floor
    }
    \label{fig:slab-info}
  \end{subfigure}\hfill
  \begin{subfigure}[t]{0.32\textwidth}
    \centering
    \includegraphics[width=\linewidth,keepaspectratio]{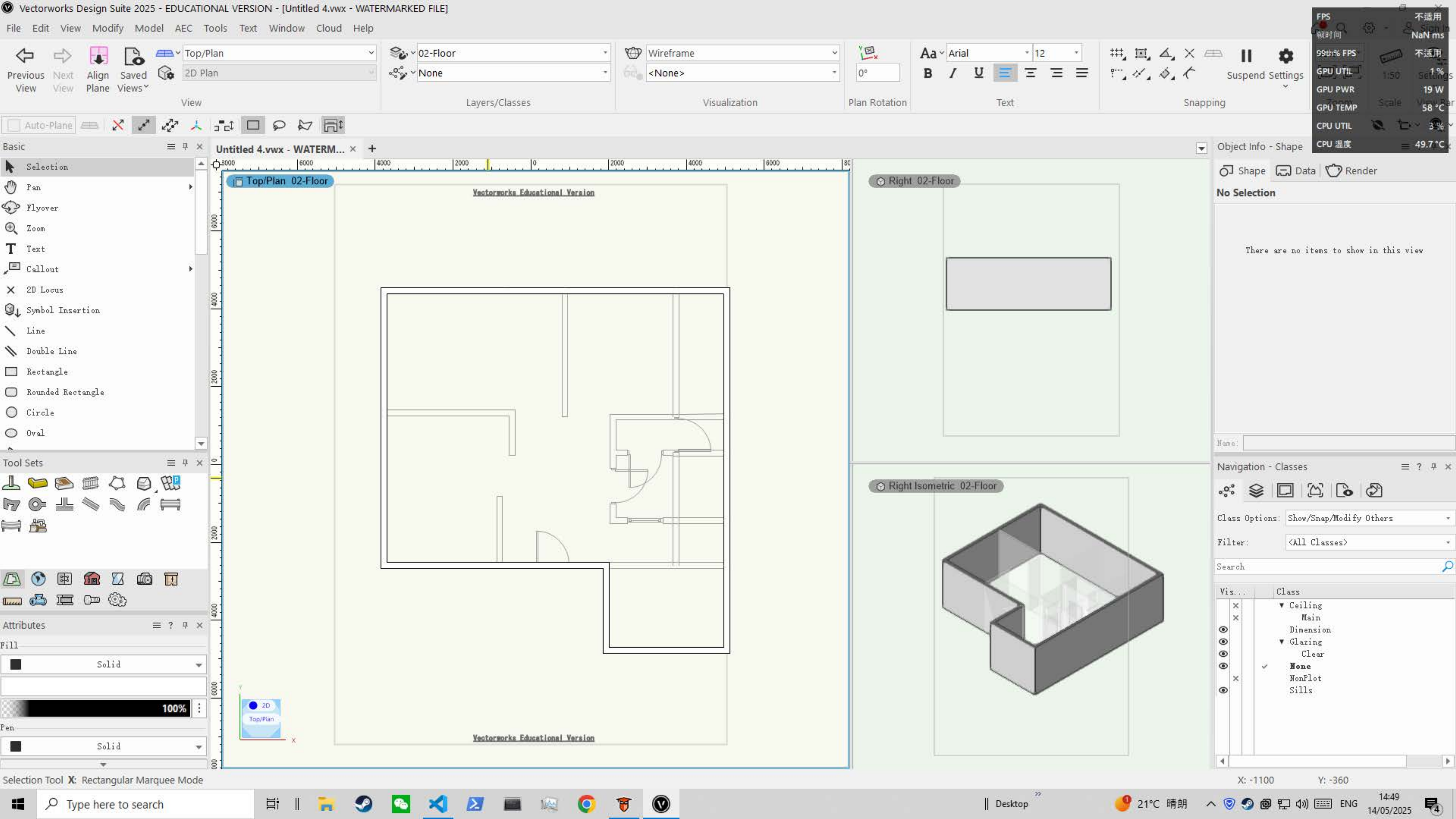}
    \caption{\parbox[c][5\baselineskip][c]{\linewidth}{\centering
      Active slab tool\\
      \texttt{shortcut(combo = "alt + shift + 2")}
    }}
    \label{fig:slab-tool}
  \end{subfigure}\hfill
  \begin{subfigure}[t]{0.32\textwidth}
    \centering
    \includegraphics[width=\linewidth,keepaspectratio]{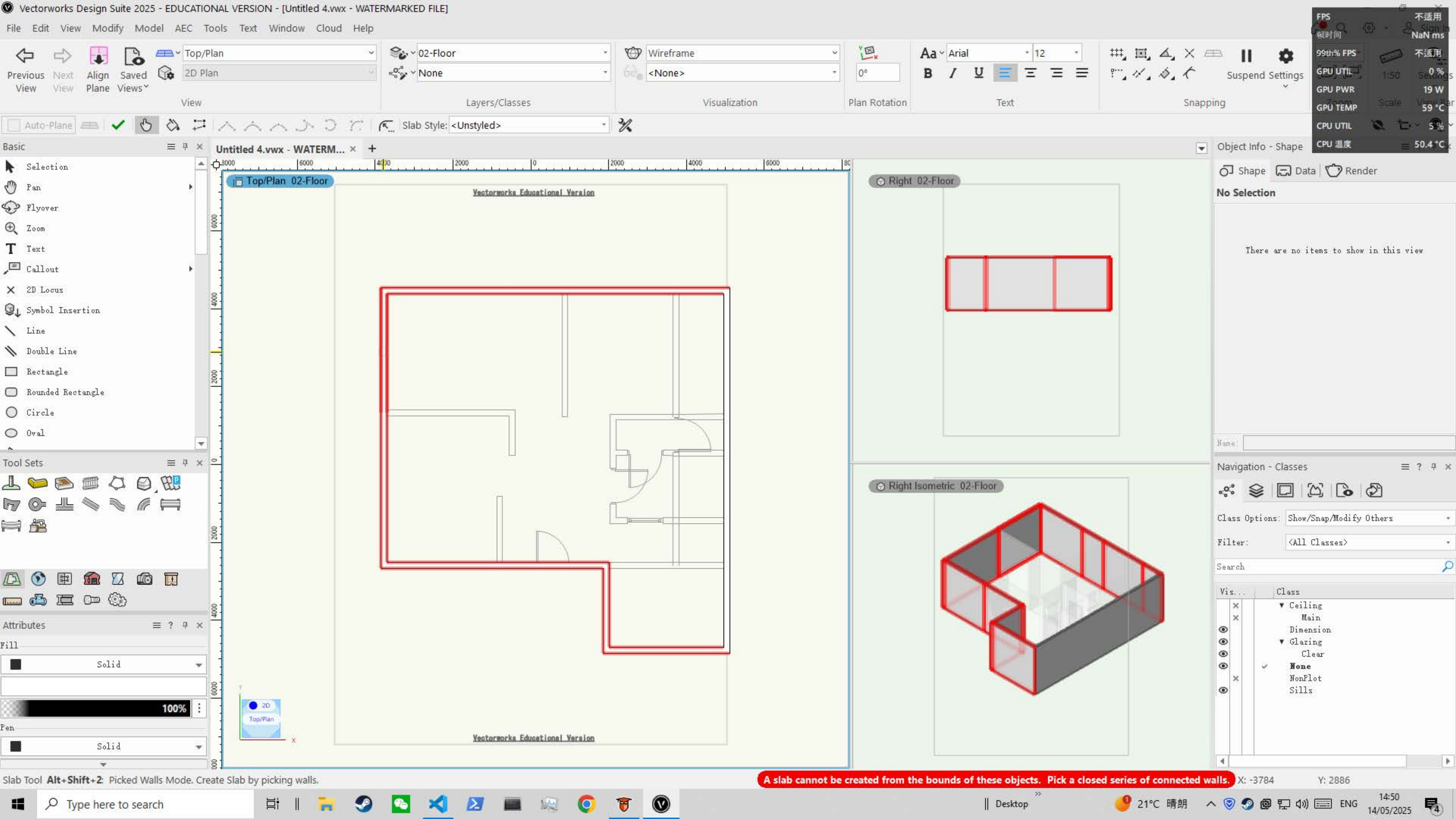}
    \caption{\parbox[c][5\baselineskip][c]{\linewidth}{\centering
      Create slab\\
      \texttt{move\_mouse\_to(x=585, y=383)}, \texttt{left\_click()}, …, \texttt{move\_mouse\_to(x=585, y=383)}, \texttt{left\_click()}, \texttt{press\_enter()}
    }}
    \label{fig:slab-actions}
  \end{subfigure}

  \caption{A failed task caused by an execution error: Create the main floor slab for the second floor, covering the area defined by the external walls. Generated action clicked a same wall twice, which cannot form a closed boundary.}
  \label{fig:slab-error111}
\end{figure}

We present examples of the three types of errors that occurred during the experiment. As illustrated in Figure~\ref{fig:planningerro111r} to Figure~\ref{fig:slab-error111}, these examples demonstrate Grounding Error, Execution Error, and Planning Error, respectively.

\newpage

\section{Visualization of Generated Models}
\label{app:visual}

\begin{figure}[H]
  \centering
  \includegraphics[width=0.45\textwidth]{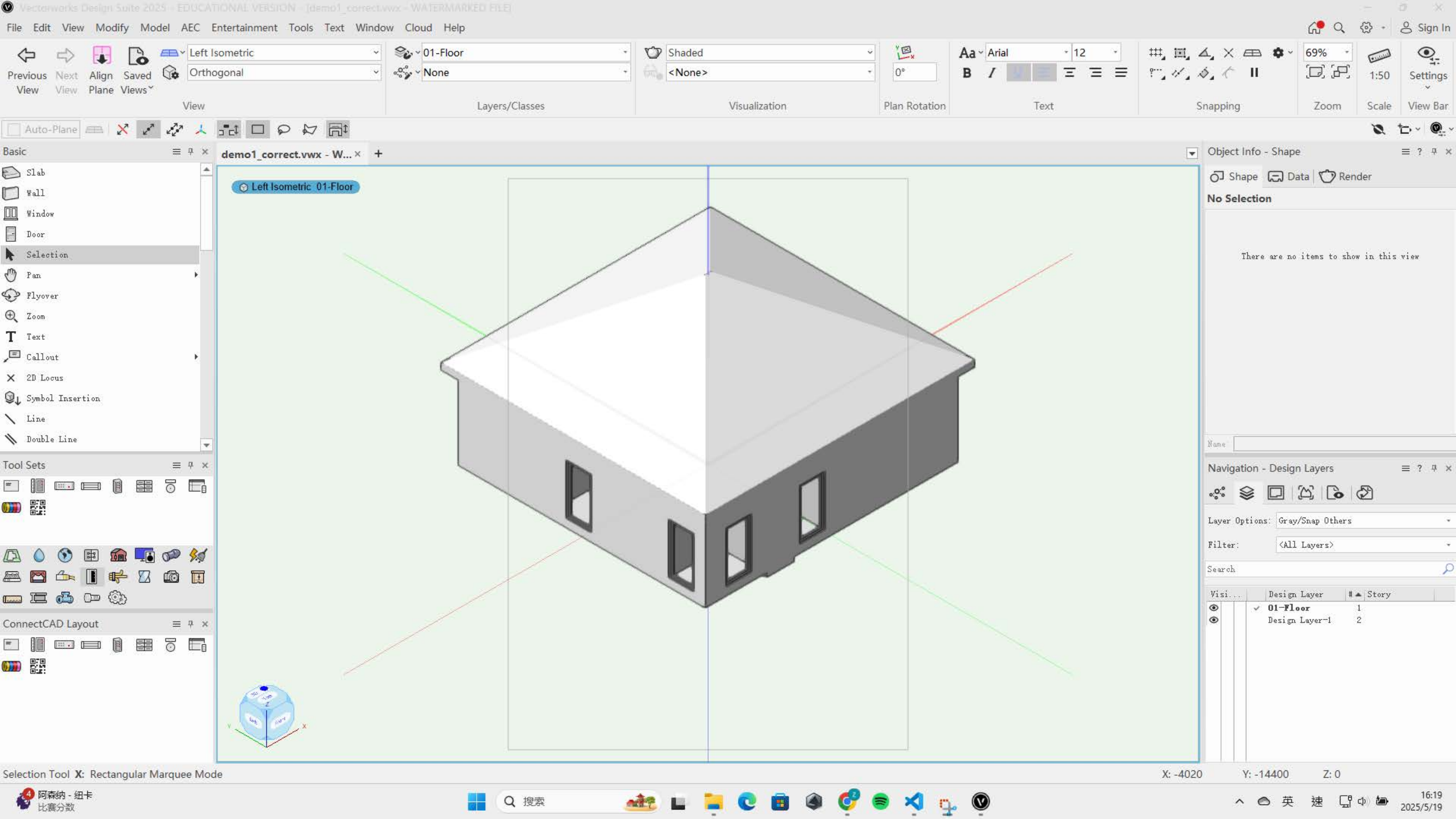}
  \includegraphics[width=0.45\textwidth]{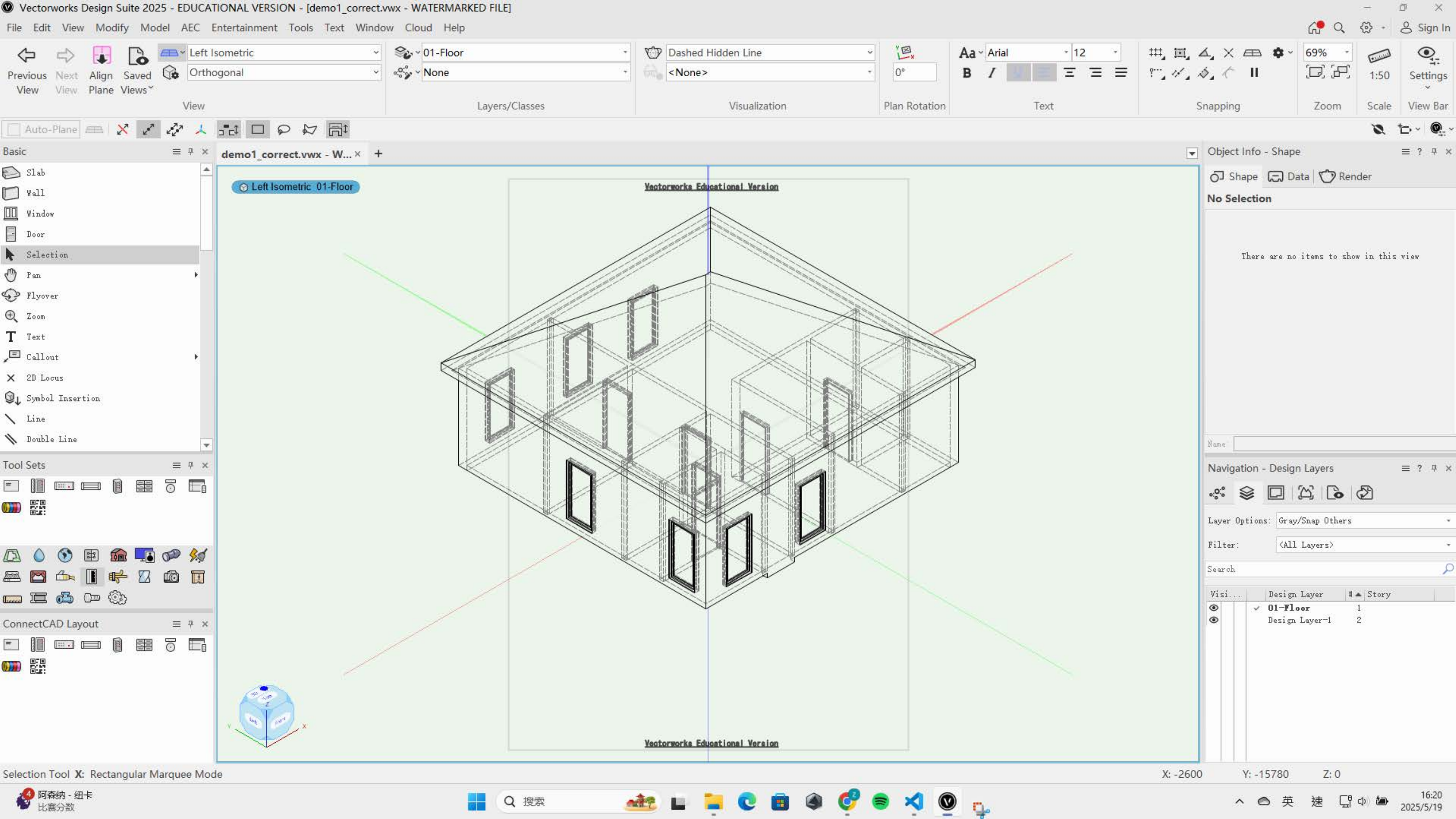}
  \caption{Task 1: \textit{Generate a one-storey office building with a large open workspace
occupying most of the floor area. The layout should also include two
enclosed meeting rooms, a manager’s office, a small pantry, and two
restrooms.} Shown are the resulting building model in shaded (left) and wireframe (right) modes.}
\end{figure}
\begin{figure}[H]
  \centering
  \includegraphics[width=0.45\textwidth]{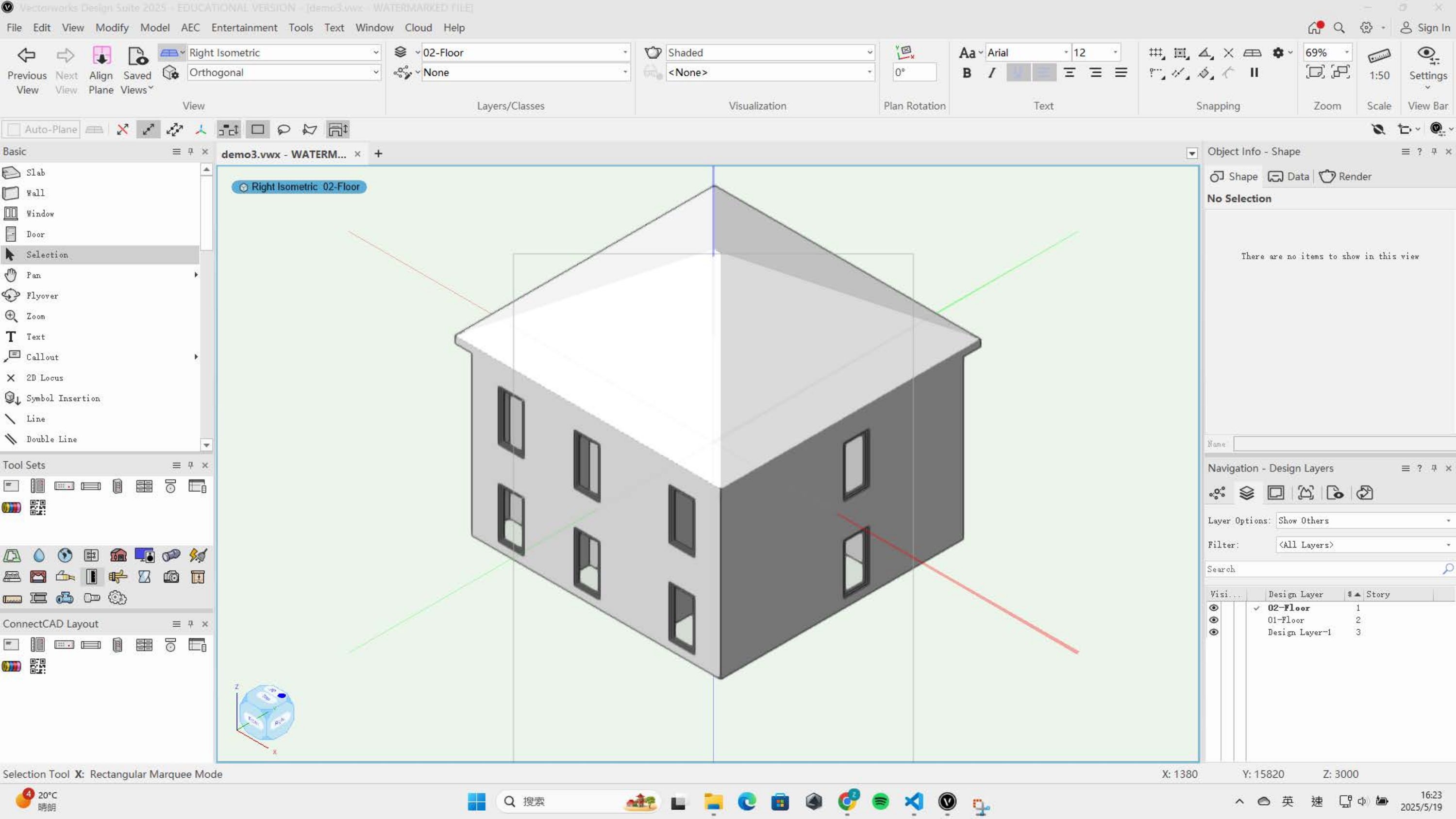}
  \includegraphics[width=0.45\textwidth]{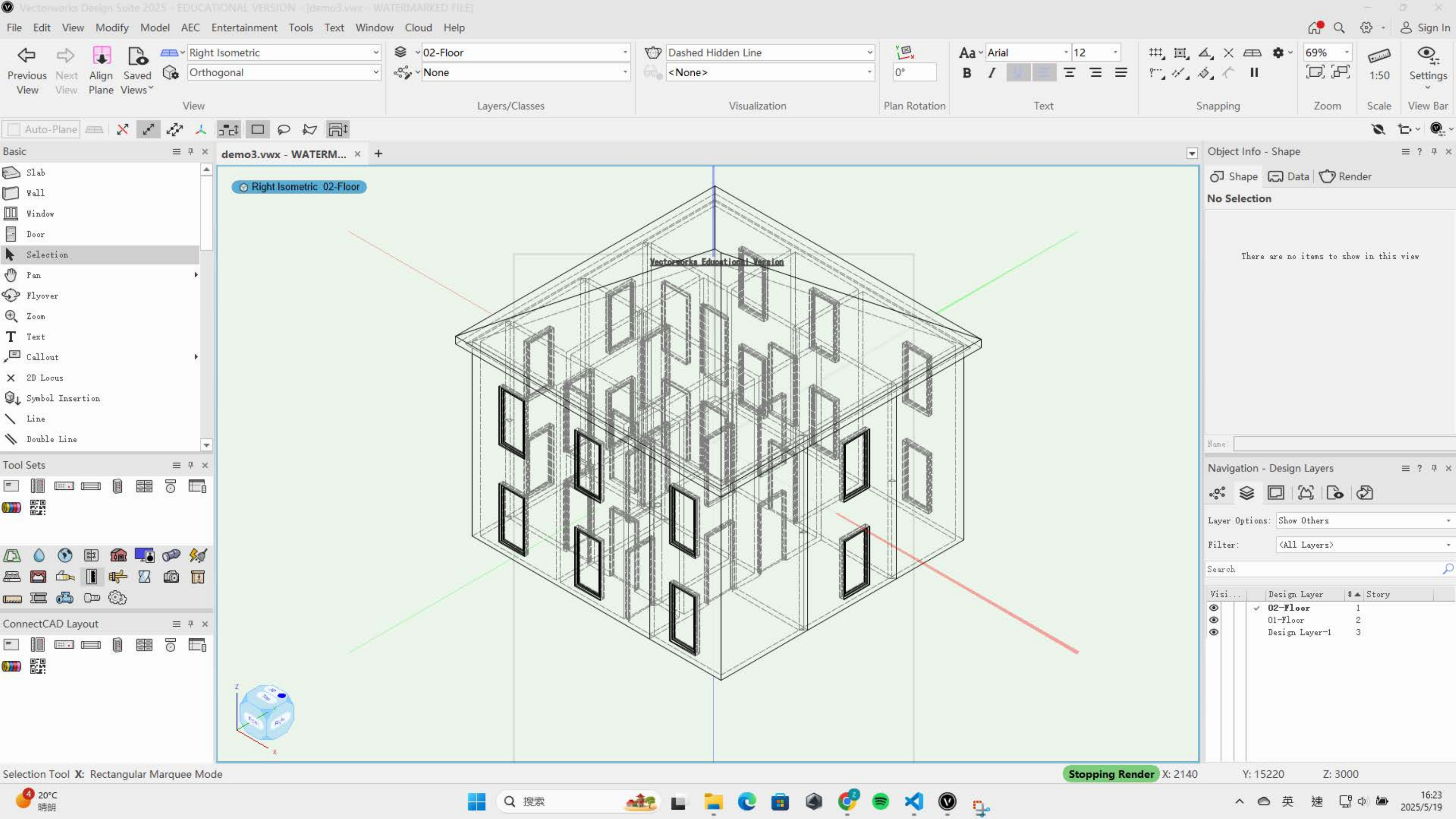}
  \caption{Task 4: \textit{Design a two-storey hospital building with eight distinct rooms. These must include a reception area, two consultation rooms, one minor surgery room,
one waiting area, two patient rooms, and one staff room.} Shown are the resulting building model in shaded (left) and Dashed Hidden Line (right) modes.}
\end{figure}
\begin{figure}[H]
  \centering
  \includegraphics[width=0.45\textwidth]{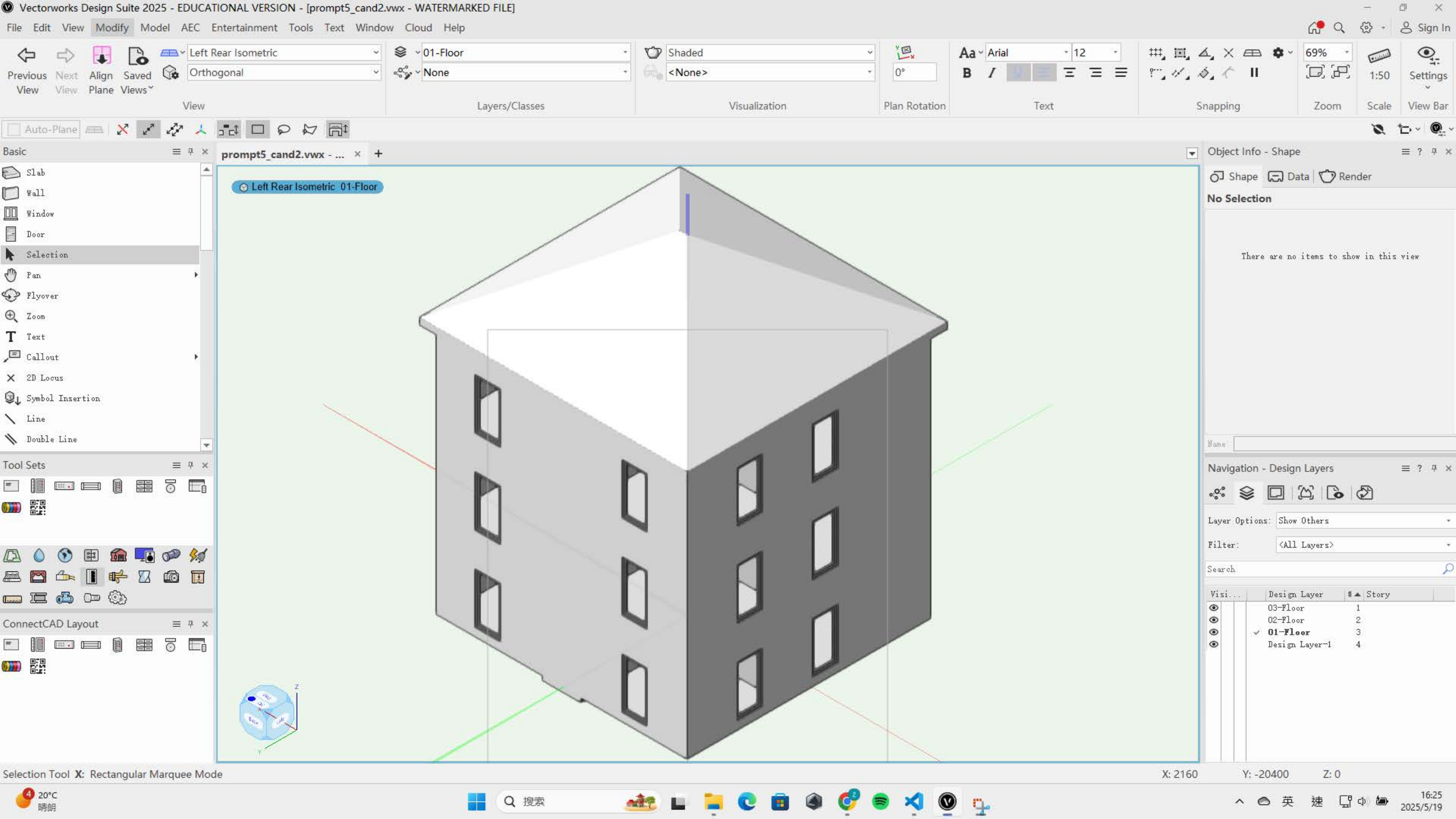}
  \includegraphics[width=0.45\textwidth]{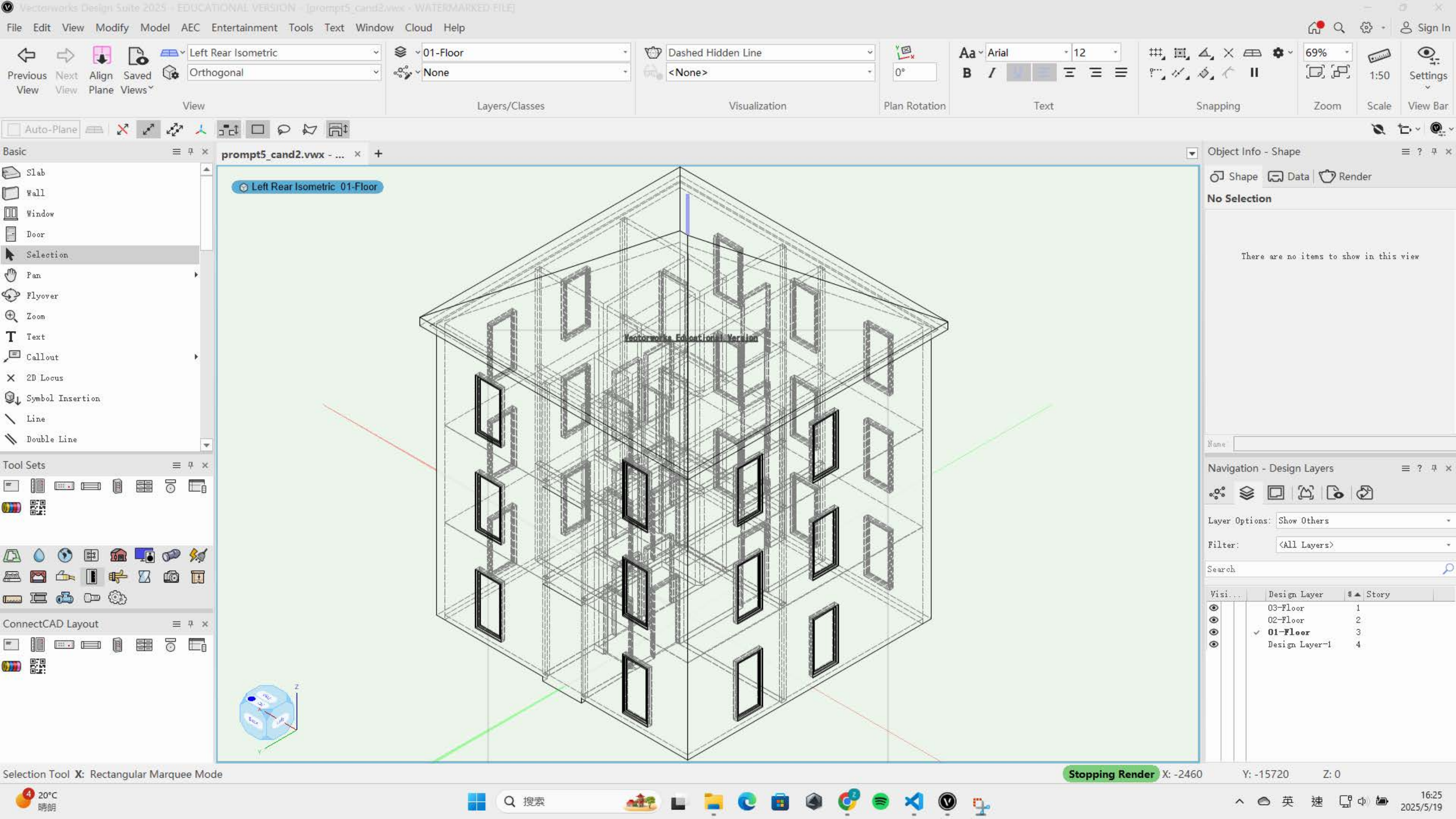}
  \caption{Task 5: \textit{Create a three-storey commercial office building where each floor has the same layout. Each floor must include two large office rooms, a small meeting room, a restroom, and a central corridor. The entrance to the building is located on the ground floor and leads directly to the corridor.} Shown are the resulting building model in shaded (left) and Dashed Hidden Line (right) modes.}
\end{figure}
\begin{figure}[H]
  \centering
  \includegraphics[width=0.45\textwidth]{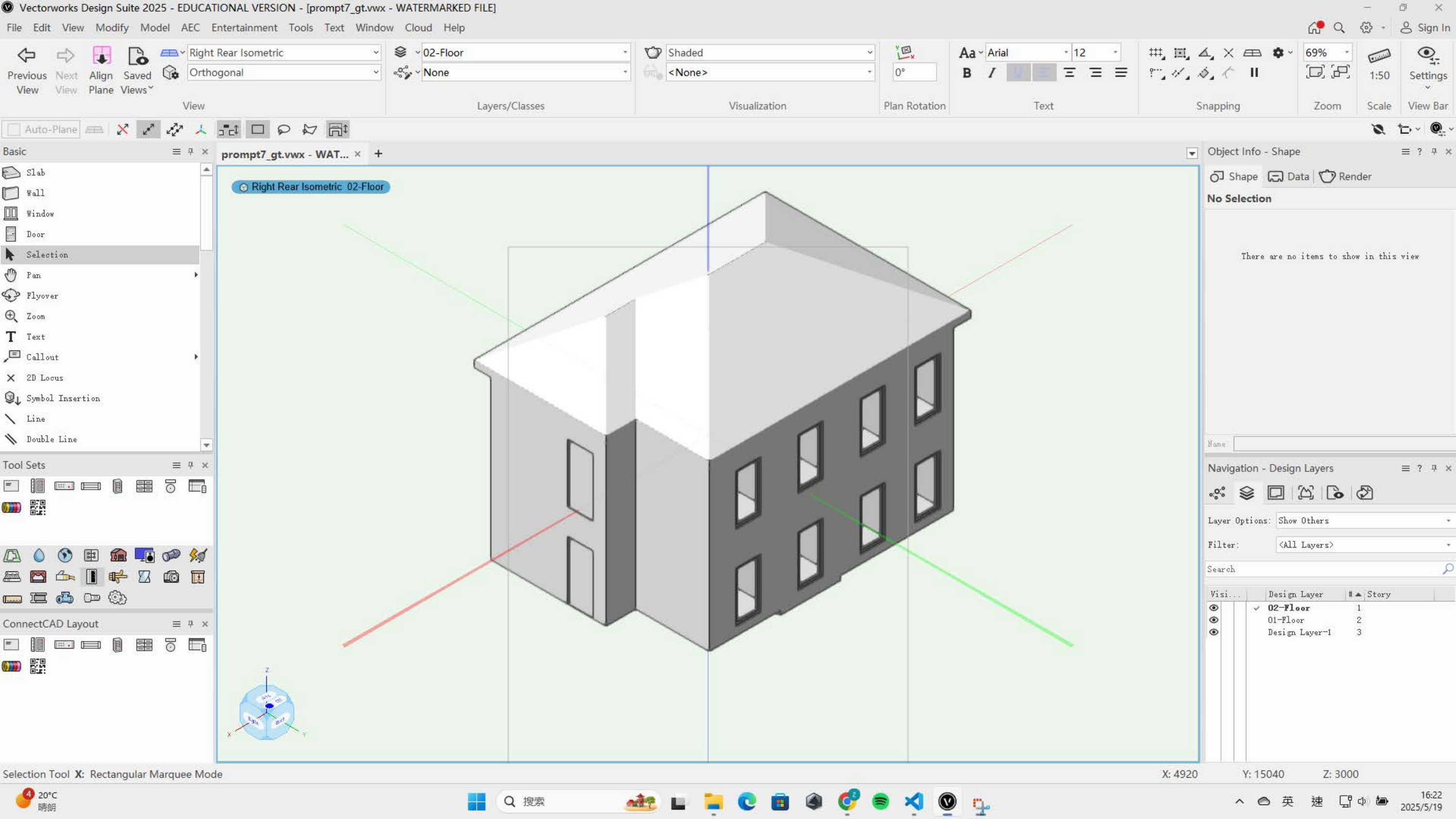}
  \includegraphics[width=0.45\textwidth]{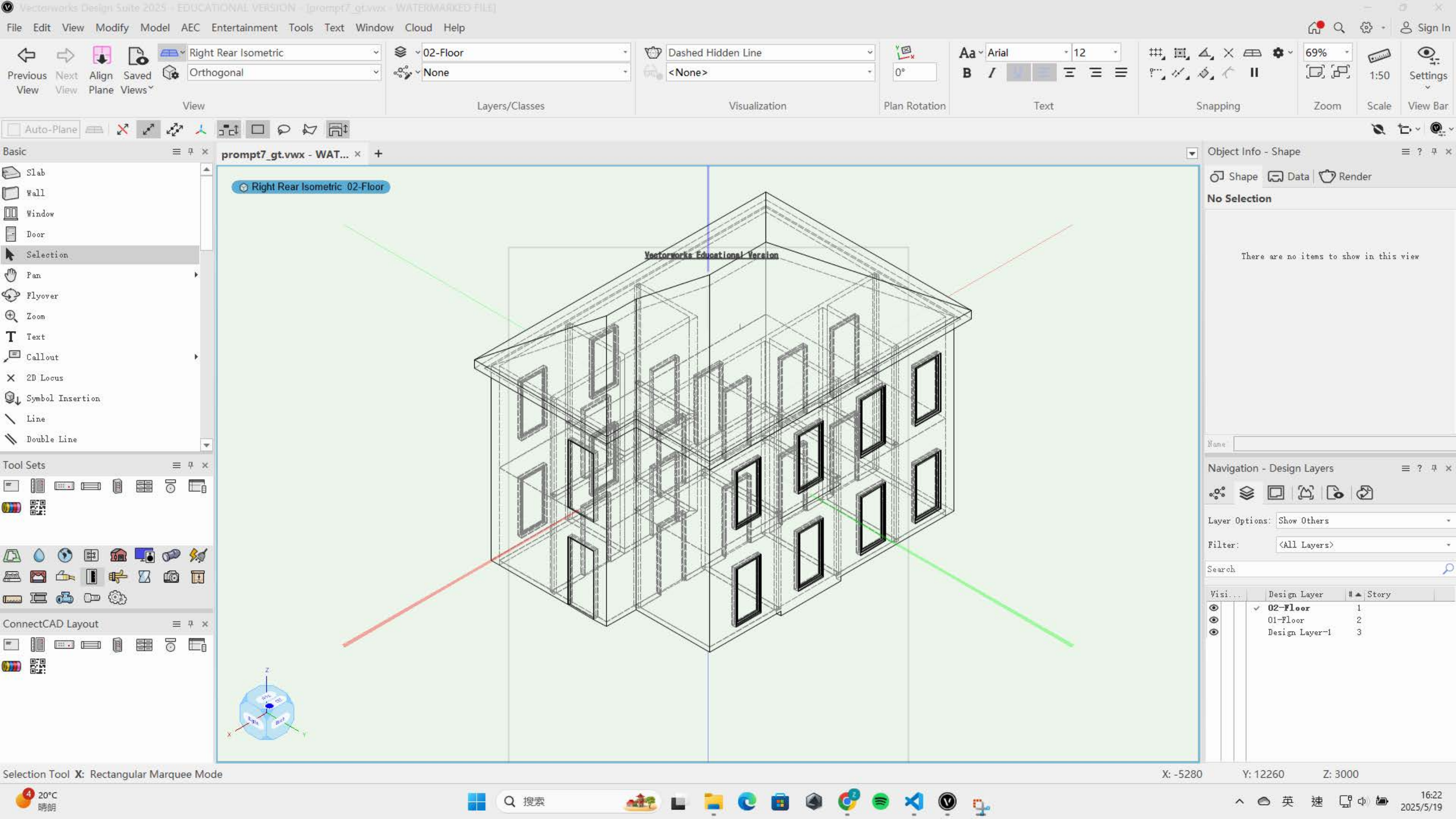}
  \caption{Task 7: \textit{Generate a two-storey building based on the sketch.} Shown are the resulting building model in shaded (left) and Dashed Hidden Line (right) modes.}
\end{figure}
\begin{figure}[H]
  \centering
  \includegraphics[width=0.45\textwidth]{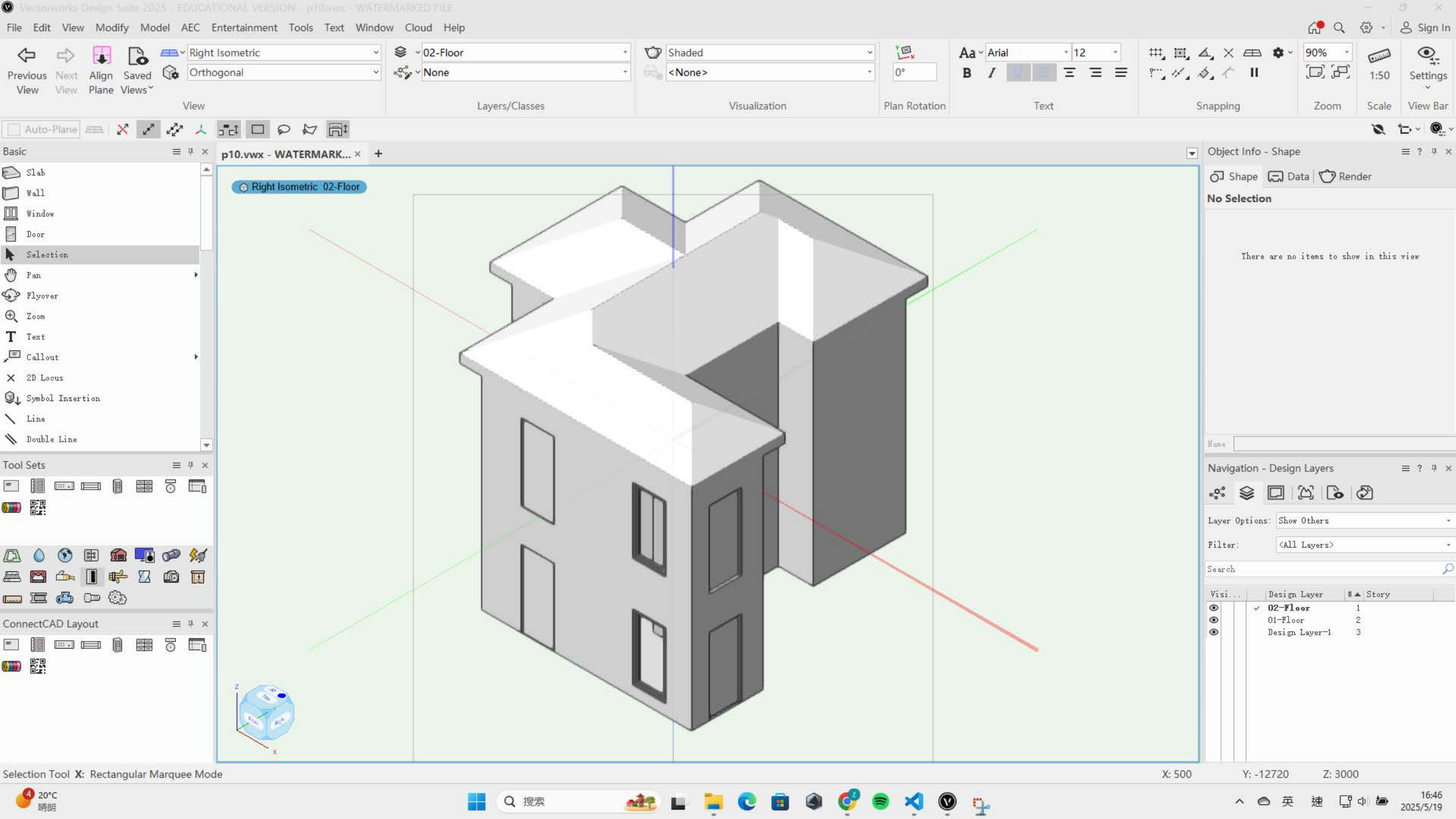}
  \includegraphics[width=0.45\textwidth]{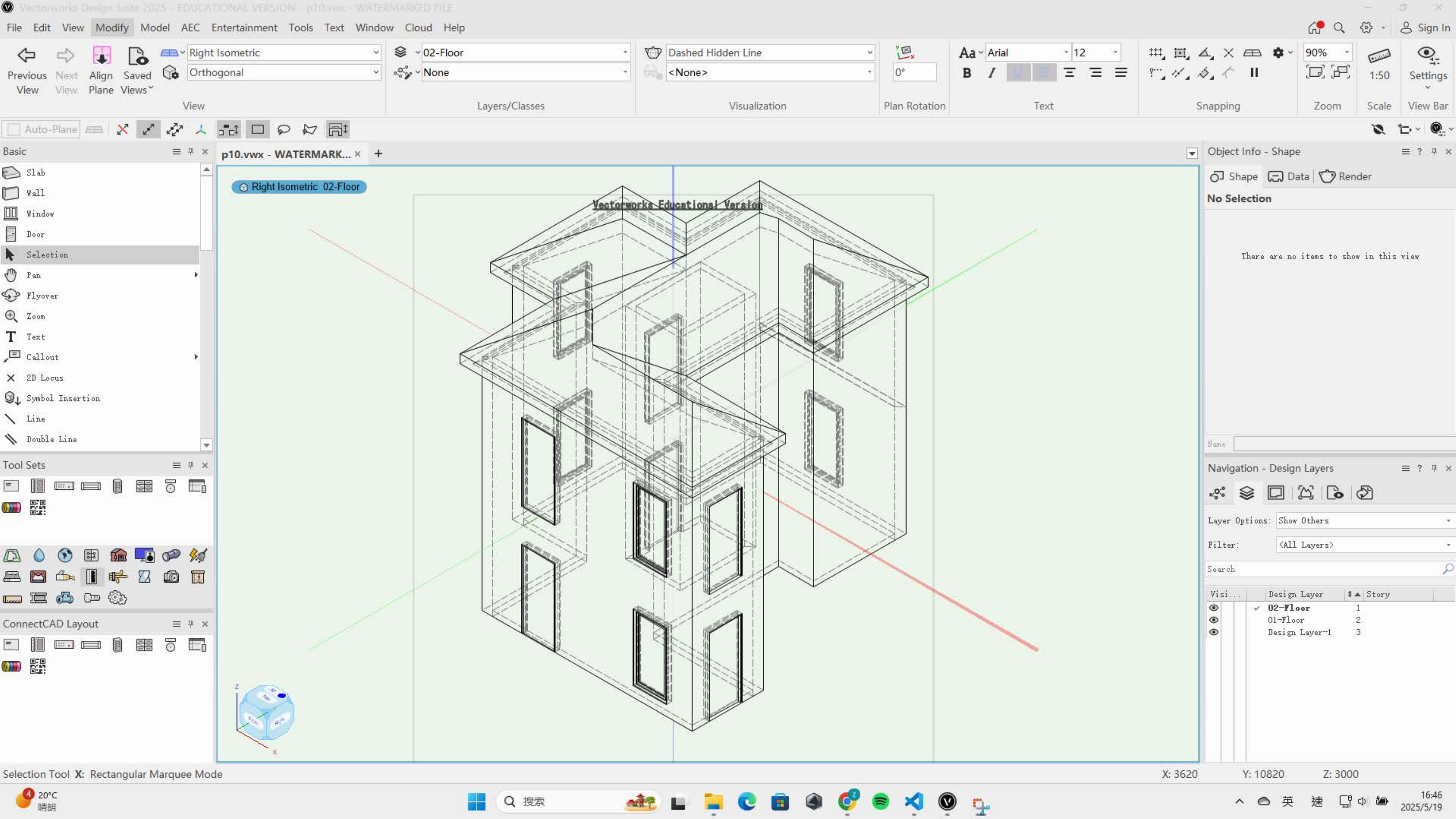}
  \caption{Task 10: \textit{Generate a one-storey building based on the sketch.} Shown are the resulting building model in shaded (left) and Dashed Hidden Line (right) modes.}
\end{figure}

\begin{figure}[H]
  \centering
  \includegraphics[width=0.45\textwidth]{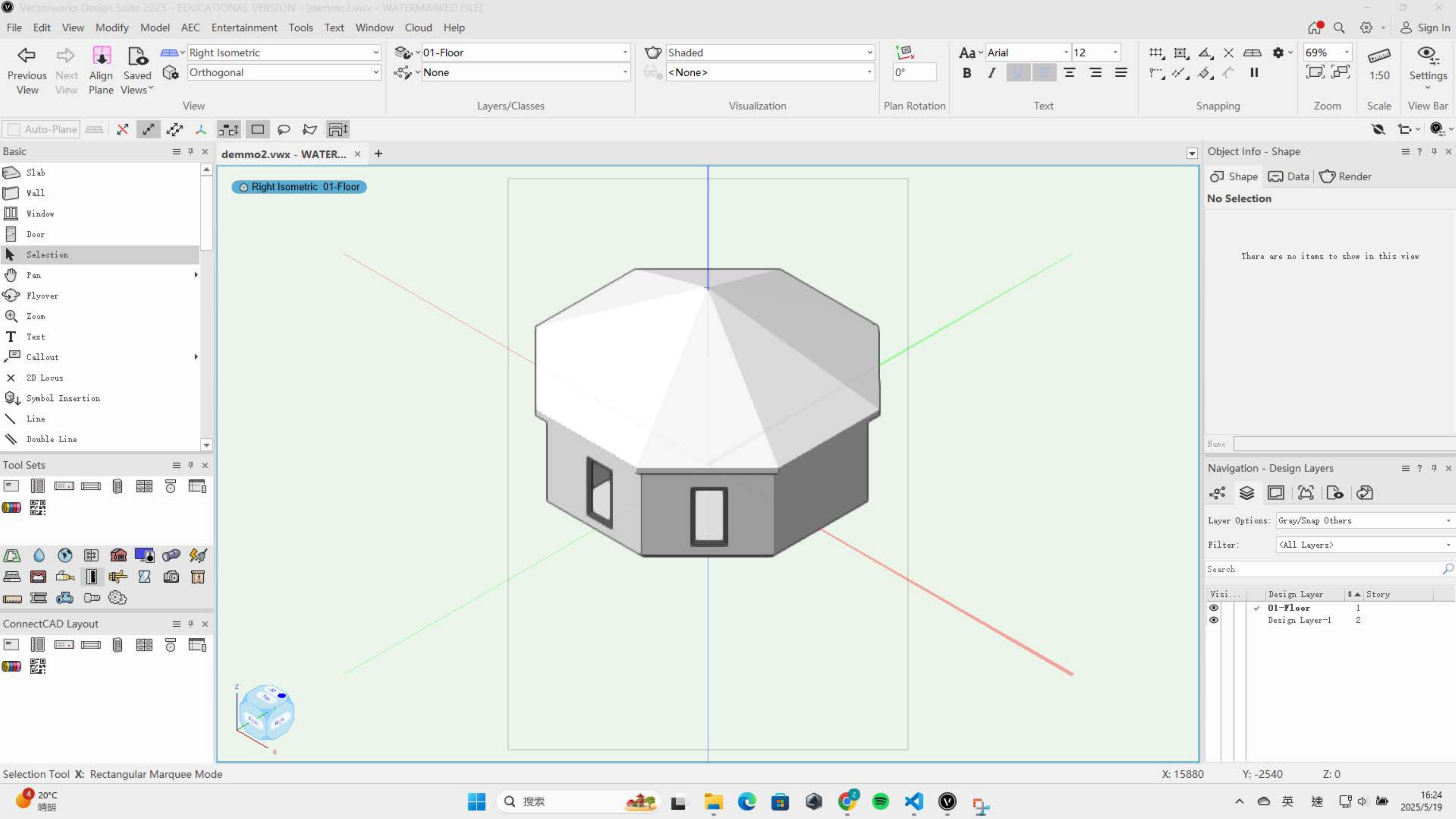}
  \includegraphics[width=0.45\textwidth]{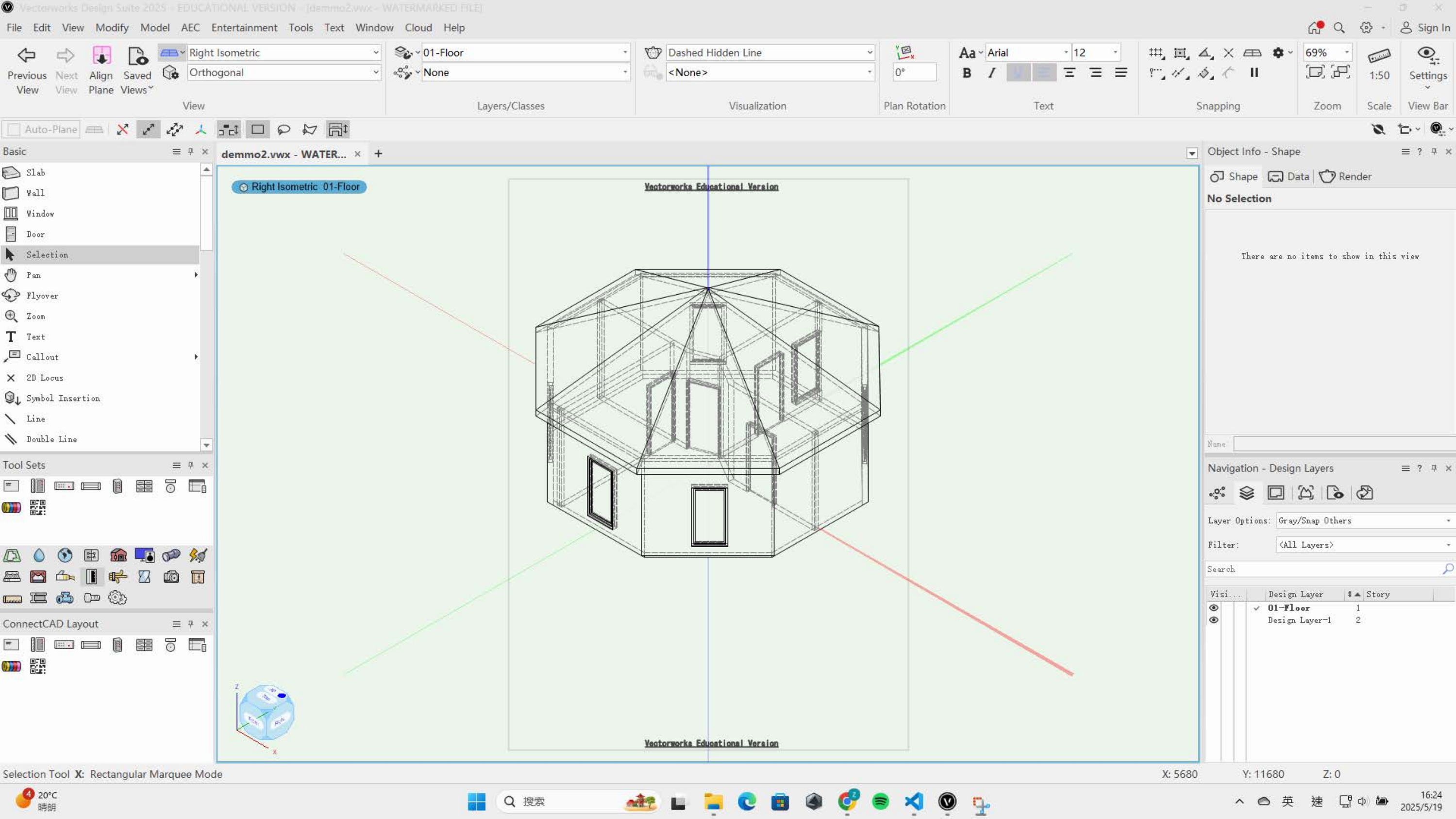}
  \caption{Task 16: \textit{Generate a building model based on a hand-drawn octagon floorplan, modifying the interior layout to include
four rooms instead of three.} Shown are the resulting building model in shaded (left) and Dashed Hidden Line (right) modes.}
\end{figure}

\begin{figure}[H]
  \centering
  \includegraphics[width=0.45\textwidth]{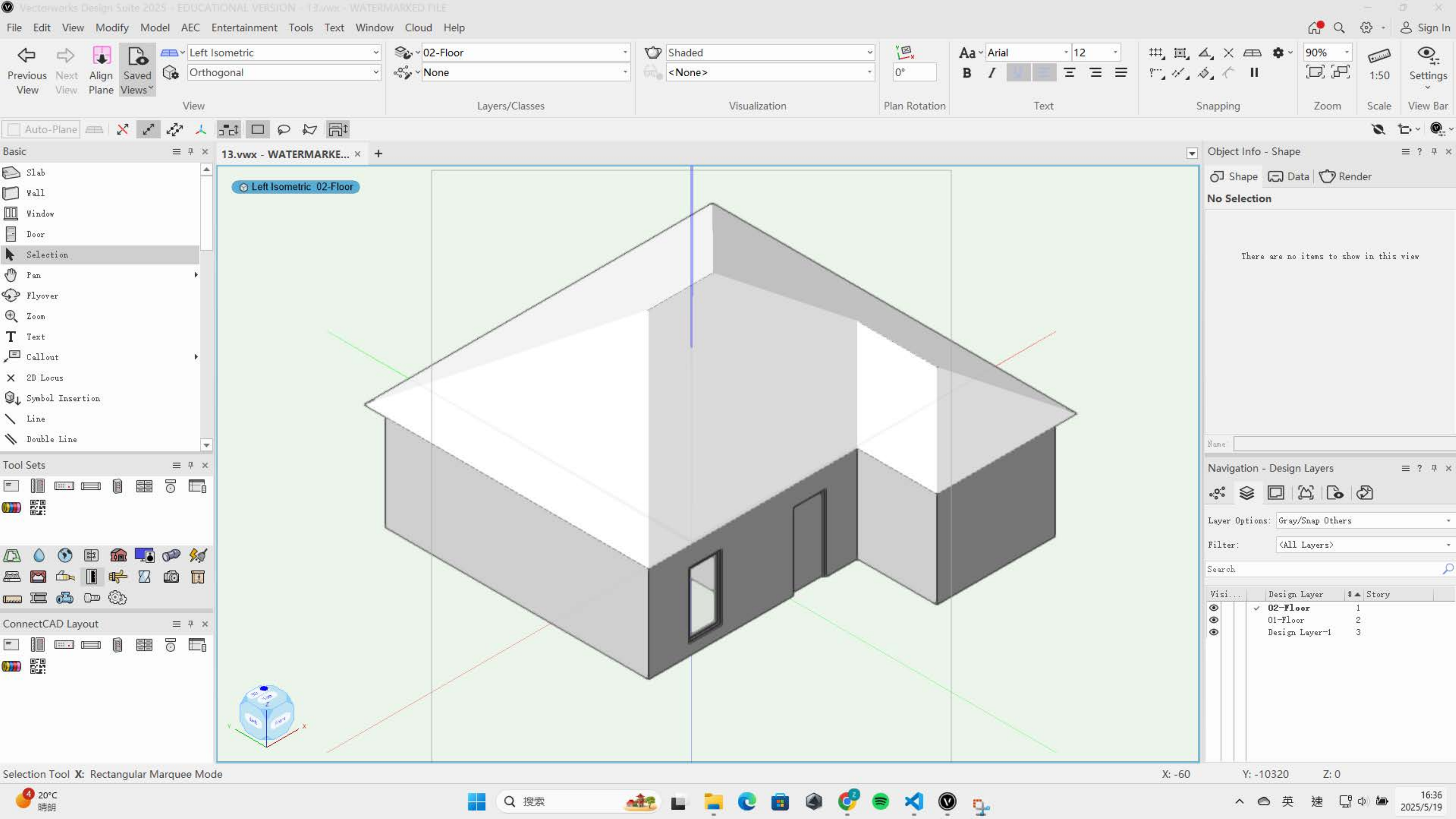}
  \includegraphics[width=0.45\textwidth]{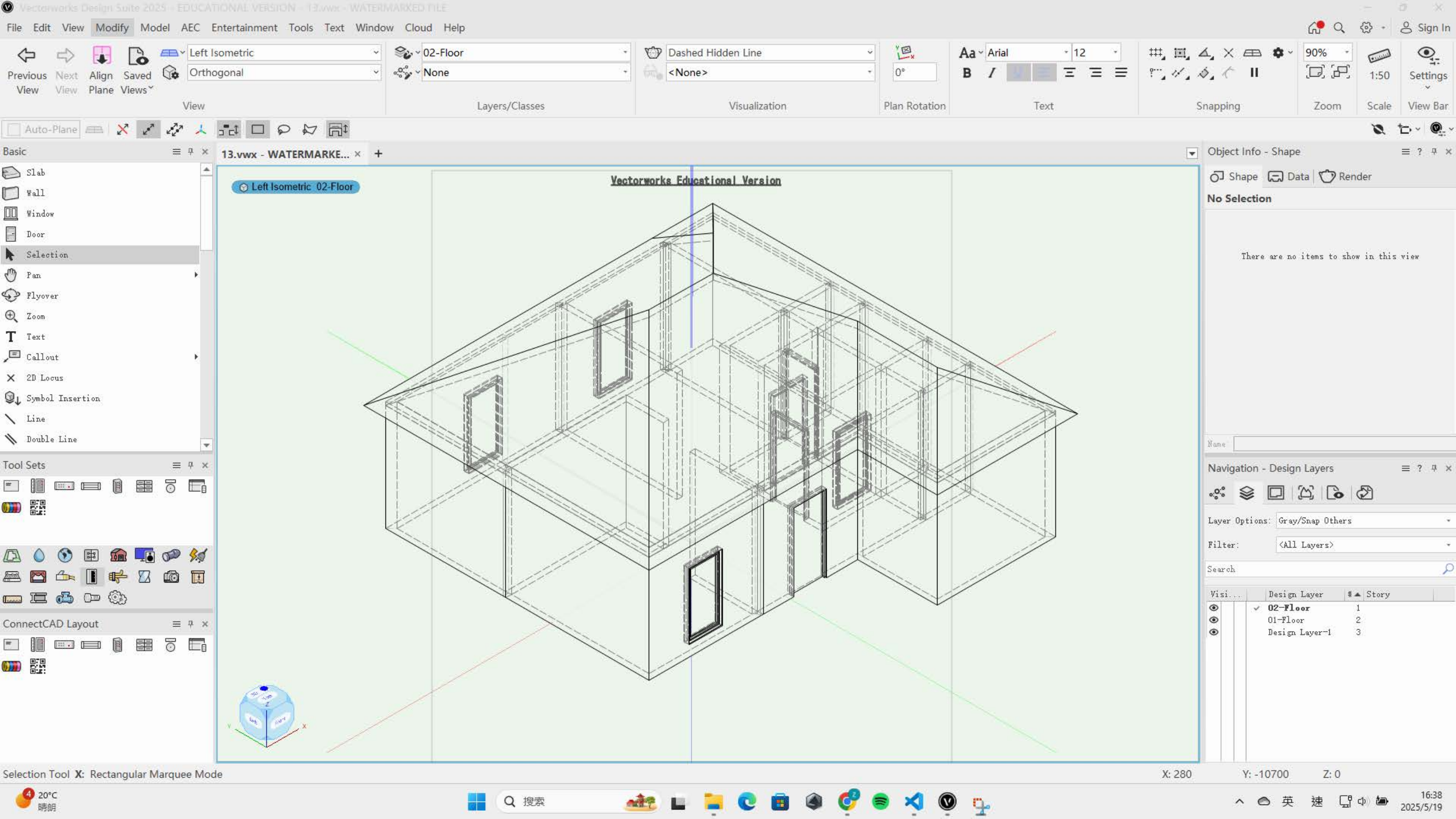}
  \caption{Task 22: \textit{Make a one-floor apartment based on the image but
with updates. Add an additional room in the bottom-right corner.} Shown are the resulting building model in shaded (left) and Dashed Hidden Line (right) modes.}
\end{figure}

\newpage
\section{Floorplan Generation Comparison }
\label{compare_floorplan}

\begin{figure}[htbp]
  \centering
  %------------ First image ------------%
  \begin{subfigure}[b]{0.33\textwidth}
    \includegraphics[height=4cm, keepaspectratio]{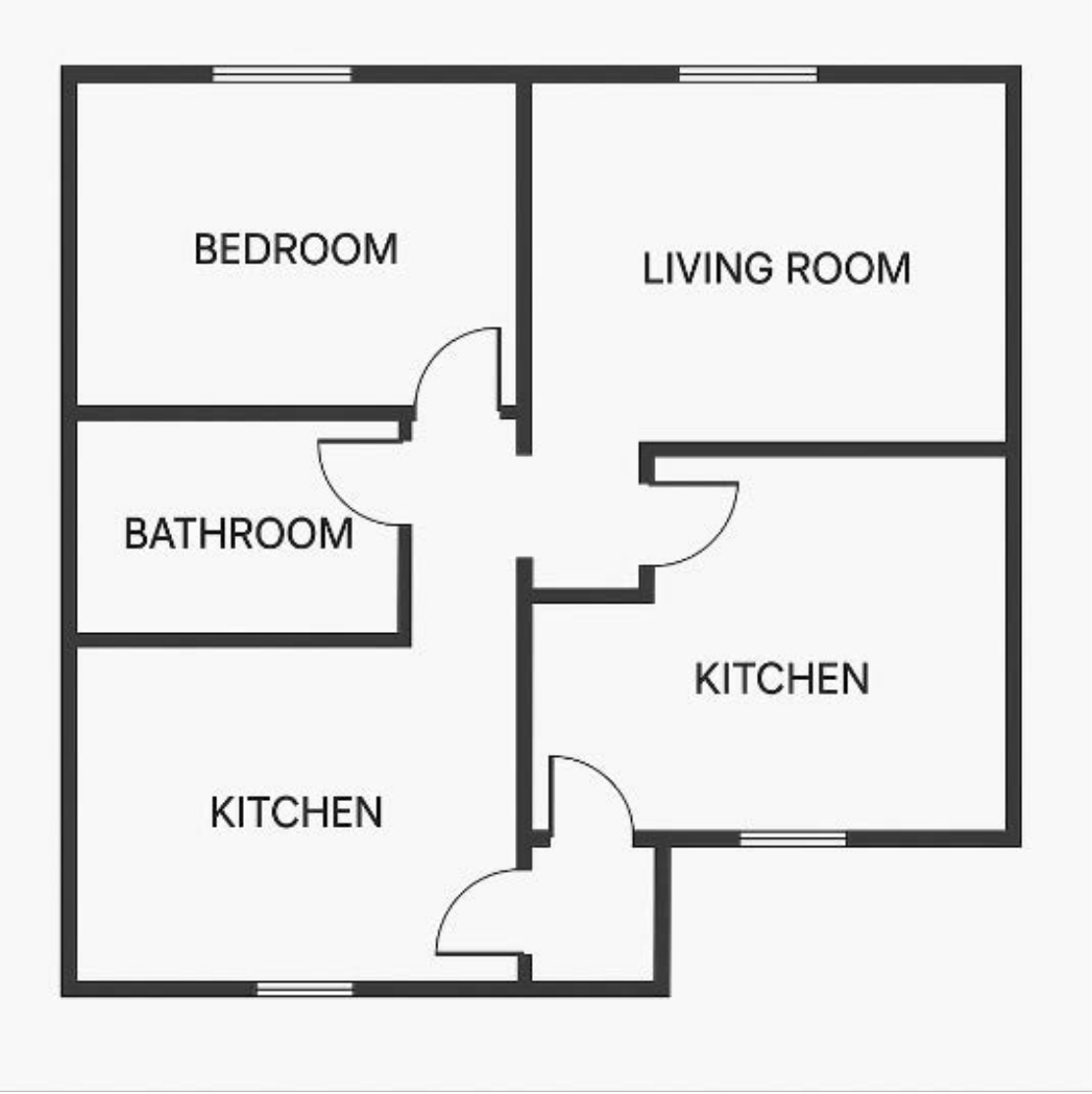}
    \caption{GPT-Image-1}
    \label{fig:design-a}
  \end{subfigure}\hfill
  %------------ Second image -----------%
  \begin{subfigure}[b]{0.33\textwidth}
    \includegraphics[height=4cm, keepaspectratio]{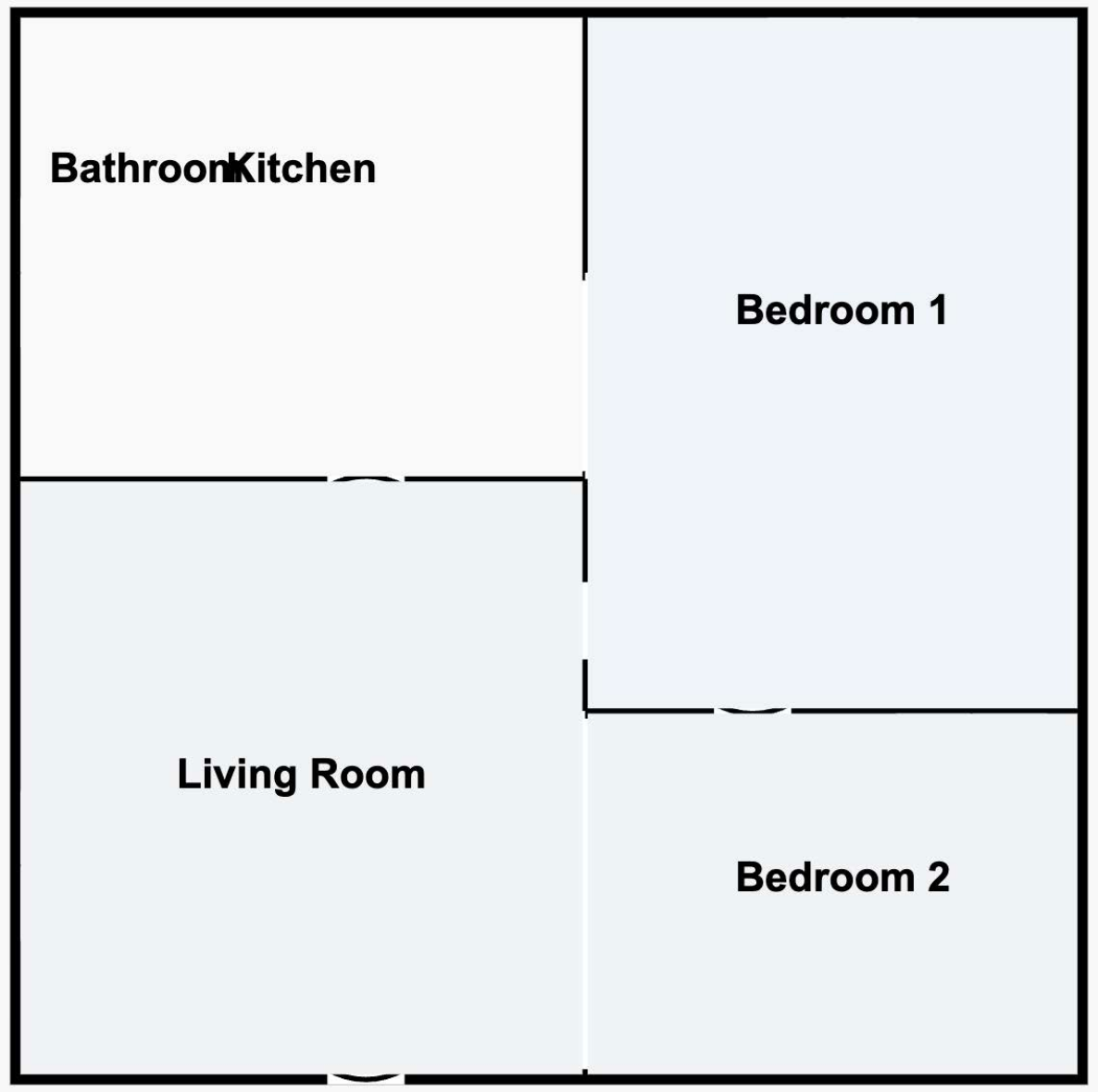}
    \caption{Claude 3.7}
    \label{fig:claude_ex1}
  \end{subfigure}\hfill
  %------------ Third image ------------%
  \begin{subfigure}[b]{0.33\textwidth}
    \includegraphics[height=4cm, keepaspectratio]{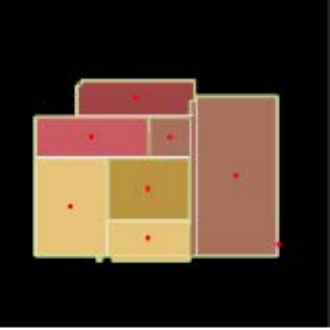}
    \caption{House-GAN++}
    \label{fig:gan_ex1}
  \end{subfigure}

  \caption{Example 1: \textit{Generate a residential floorplan with 5 rooms: 2 bedrooms, 1 living room, 1 kitchen, and 1 bathroom.}}
  \label{fig:DESIGN111}
\end{figure}

\begin{figure}[htbp]
  \centering
  %------------ First image ------------%
  \begin{subfigure}[b]{0.33\textwidth}
    \includegraphics[height=4cm, keepaspectratio]{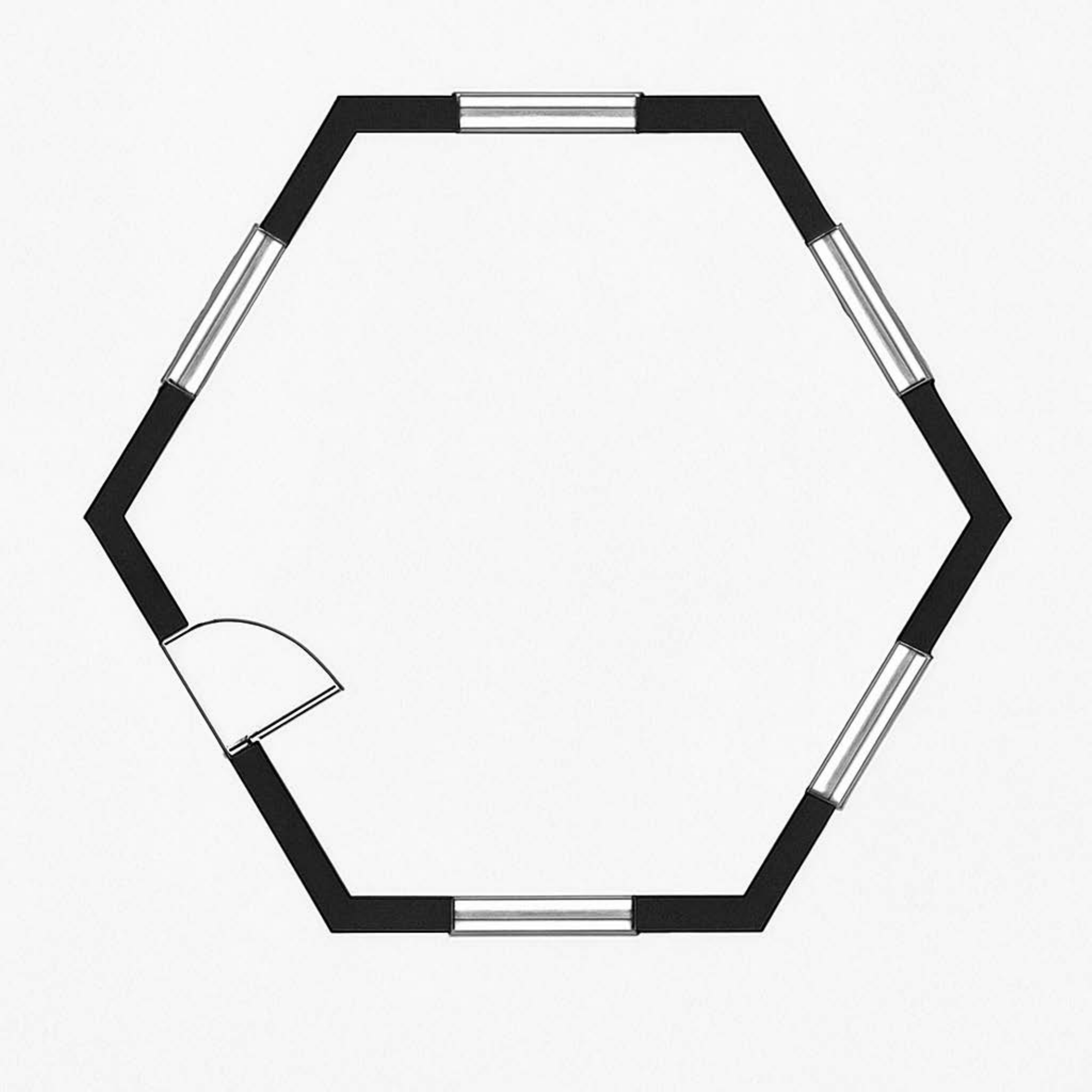}
    \caption{GPT-Image-1}
    \label{fig:design-a}
  \end{subfigure}\hfill
  %------------ Second image -----------%
  \begin{subfigure}[b]{0.33\textwidth}
    \includegraphics[height=4cm, keepaspectratio]{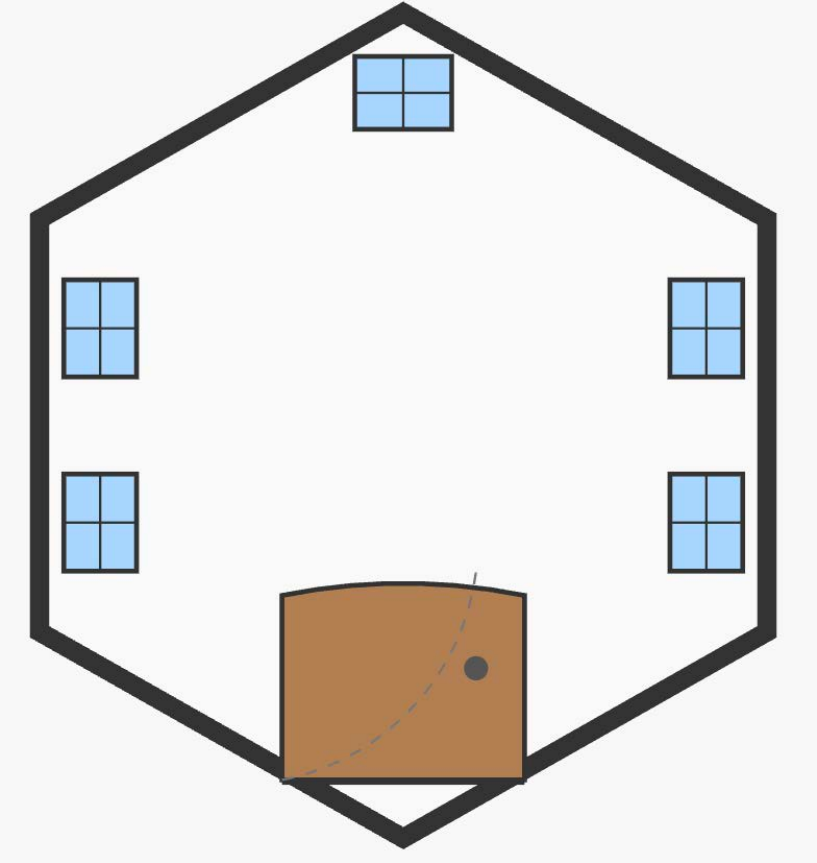}
    \caption{Claude 3.7}
    \label{fig:design-b}
  \end{subfigure}\hfill
  %------------ Third image ------------%
  \begin{subfigure}[b]{0.33\textwidth}
    \includegraphics[height=4cm, keepaspectratio]{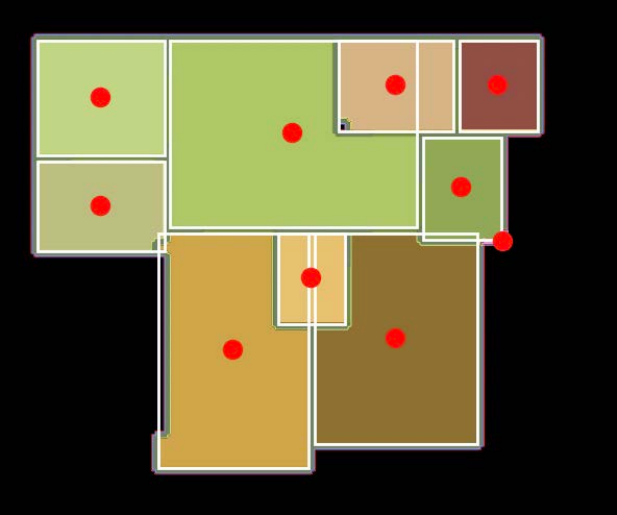}
    \caption{House-GAN++}
    \label{fig:design-c}
  \end{subfigure}

  \caption{Example 2: \textit{Design a floorplan with a complex polygonal footprint (hexagonal).}}
  \label{fig:DESIGN222}
\end{figure}

\begin{figure}[htbp]
  \centering
  %------------ First image ------------%
  \begin{subfigure}[b]{0.33\textwidth}
    \includegraphics[height=4cm]{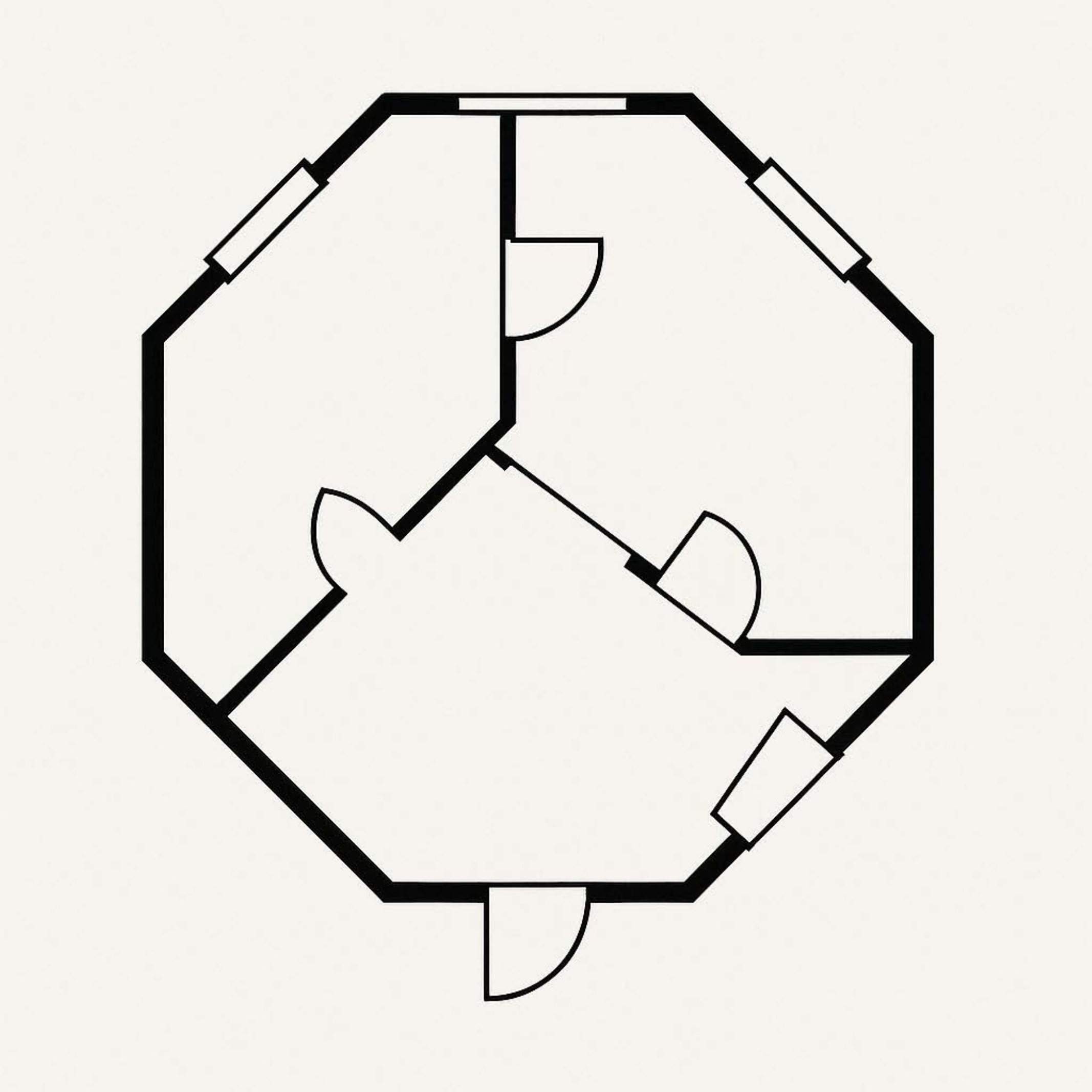}
    \caption{GPT-Image-1}
    \label{fig:our_ex3}
  \end{subfigure}\hfill
  %------------ Second image -----------%
  \begin{subfigure}[b]{0.33\textwidth}
    \includegraphics[height=4cm]{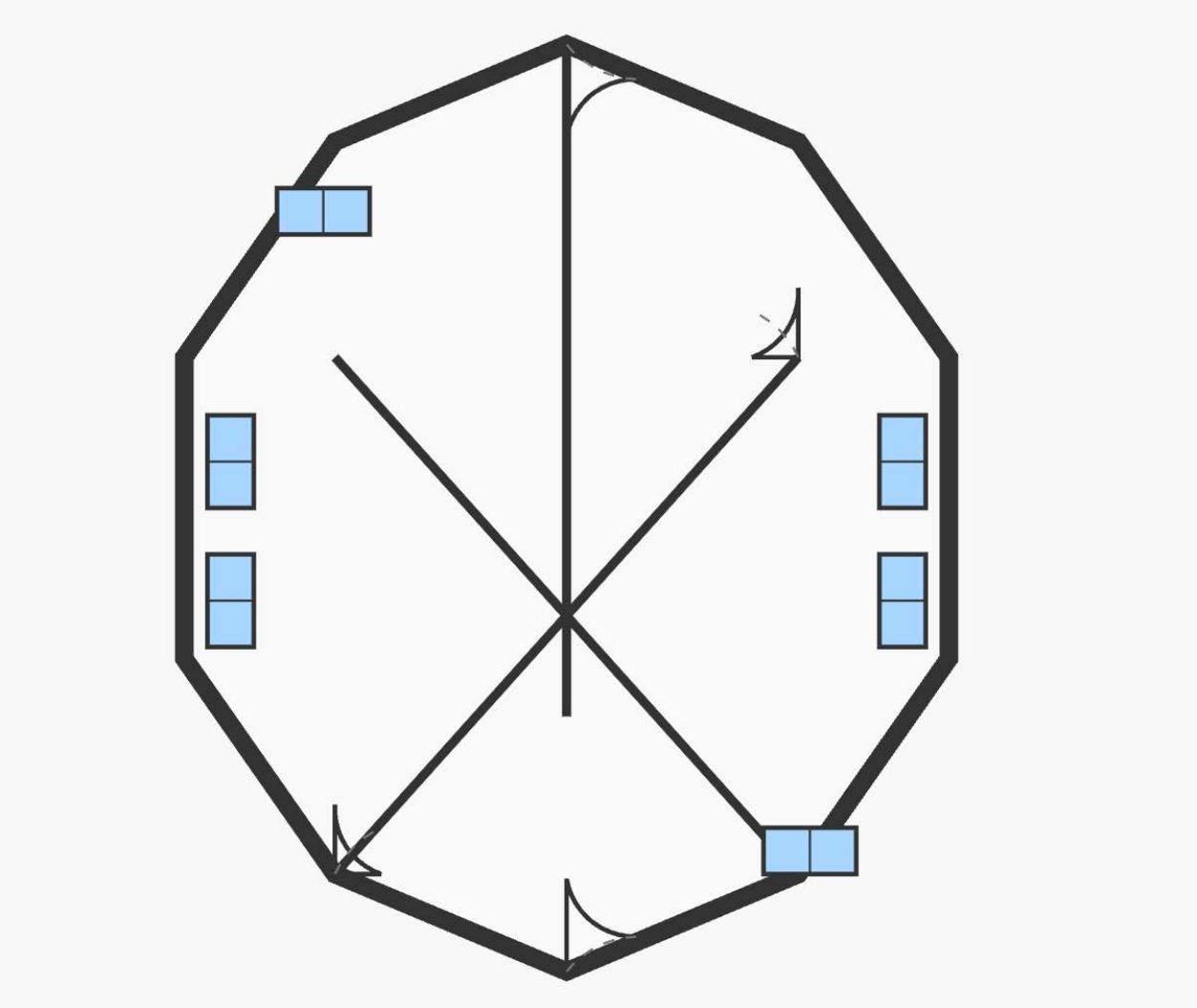}
    \caption{Claude 3.7}
    \label{fig:claude_ex3}
  \end{subfigure}\hfill
  %------------ Third text-only -----------%
  \begin{subfigure}[b]{0.33\textwidth}
    \begin{minipage}[c][4cm][c]{\linewidth} % fixed height minipage
      \centering
      \textit{GAN not supported}
    \end{minipage}
    \caption{House-GAN++}
    \label{fig:design-c}
  \end{subfigure}

  \caption{Example 3: \textit{Generated floorplans for Task 6 in the Mini Building Benchmark.} The GAN model does not support this task.}
  \label{fig:DESIGN333}
\end{figure}

In this section, we present results of floorplan image generation using three different approaches: gpt-image-1, Claude 3.7, and House-GAN++ \cite{nauata2021house}. Figures~\ref{fig:DESIGN111} to~\ref{fig:DESIGN333} show three representative examples. As illustrated, gpt-image-1 produces the most satisfactory results, offering layouts that are not only reasonable but also well-suited for downstream segmentation tasks.

\newpage
\section{BIMgent Prompts}

\subsection{Prompt for High-level Planner}

\begin{lstlisting}[style=prompt, caption={Example prompt used for High-level Planner}, label={lst:example-prompt}]
You are an assistant acting as a high-level planner for a building design and construction project using BIM software. You have been provided with the complete floorplan metadata and a design task from the architect. Your role is to outline the construction process as a sequence of high-level, logically ordered steps that guide the modeling workflow within the BIM environment.

Task Description:
<Task Description>

A structured floorplan including coordinates and types of walls, doors, windows, etc.:
<Interpreted Floorplan Metadata>


Here are some hints to support your decision-making process.
1. Generate high-level construction steps following the typical architectural workflow, for example: creating layers, placing walls ...
2. For each construction step, follow these rules:
    - Determine the number of storeys from the task description. For each storey, create a corresponding layer.
    - All floors are identical. For each floor, construct all required components based on the floorplan. 
    - For each component, specify both its name and its assigned floor (e.g., floor2). 
...


You must respond strictly in the following format. Do not include any comments, explanations, or additional information. Output only the requested data exactly as specified below:

{    
    "step 1": {
        "class": "layer"
        "component": "layer_floorx",
        "description": "Detailed which design layer should be created currently for the current floor.",
    },
    "step 2": {
        "class": "external walls"
        "component": "wallx_floorx, ...",
        "description": "Detailed description of what needs to be done"
    },
    "step x": {
        "class": "the class of the drawing components."
        "component": "the components in specific floor that should be draw here",
        "description": "Detailed description of what needs to be done"
    }
}

\end{lstlisting}
\newpage

\subsection{Prompt for Low-level Planner}

\begin{lstlisting}[style=prompt, caption={Example prompt used for Low-level Planner}, label={lst:example-prompt}]
You are an assistant acting as a low-level planner, responsible for generating detailed action steps based on software guidance and the provided high-level plan.

your current task:
<Current General Step>

tool guidance:
<Software Documentation Retrieved via RAG>

floorplan_metadata:
<Interpreted Floorplan Metadata>       

        
Here are some hints to support your decision-making process.
You have to decide two types of actions:
Vision-Driven: Actions for which explicit coordinates are not yet provided, including shortcut-based operations.
...

Pure-Action: Actions for which coordinates are explicitly provided in the floorplan metadata, such as wall creation.
<Action Definitions>
<Speculative Multi-action Execution>
...

You must respond strictly in the following format. Do not include any comments, explanations, or additional information. Output only the requested data exactly as specified below:

{{        
    "sub_step_1": {{               
        "action name": "...",
        "action_type": "Vision-Driven"
        "description": "Detailed current step's goal"                        
    }},
    "sub_step_2": {{
        "action name": "...",
        "action_type": "Pure-Action"
        "actions": ['action1()', 'action2(x=..., y=...)', ...],
        "coordinates": [[x1, y1], ...],
        "description": "Detailed current step's goal"   
    }},
    "sub_step_x": {{
        "action name": "...",
        "action_type": "Pure-Action"
        "actions": ['action1()', 'action2(x=..., y=...)', ...],
        "coordinates": [[x1, y1], ...],
        "description": "Detailed current step's goal"   
    }}
    "sub_step_x": {{
        "action name": "...",
        "action_type": "Vision-Driven"
        "description": "Detailed x step's goal"
    }}
}}

\end{lstlisting}

\newpage
\subsection{Prompt for Action Generator}

\begin{lstlisting}[style=prompt, caption={Example prompt used for Action Generator}, label={lst:example-prompt}]
You are an action generator. Your task is to implement the provided actions by combining and sequencing them effectively to accomplish the given subtask. Ensure that all actions are context-aware, precise, and optimized for the tools and workflows in Vectorworks 2025. Below is some helpful information to assist your decision-making.
 
you will be provided with an image of the current screenshot image, which is already segmented, and the meta information of the provided image.

Your task:
<Current Substep>

meta_information of the labeled image:
<Interpreted Floorplan Metadata>   

Feedback from Supervisor:
<Reasons>   

Based on the image and metadata, you must respond by following the rules below:
1. Generate a workflow consisting of all necessary actions required to complete the given task.
2.Your output for the actions field must be a plain string representing a list of actions in the following format: "['action1()', 'action2(x=..., y=...)', ...]". Do not include any additional text, explanation, or formatting.
...

You must respond strictly in the following format. Do not include any comments,
explanations, or additional information. Output only the requested data exactly as specified below:   
{{               
    "action name": "...",
    "actions": "['move_mouse_to(x=, y=)', ... ]"
}}

\end{lstlisting}

\newpage
\subsection{Prompt for Supervisor -- Vision-Driven Workflow}

\begin{lstlisting}[style=prompt, caption={Example prompt used for Supervisor -- Vision-Driven Workflow}, label={lst:example-prompt}]
You will be provided with a screenshot of the current GUI state. 

Additionally, you are given the current task content: 
<Current Substep>

The list of executed actions: 
<Executed Actions>


Based on the information provided, you must respond according to the following rules:     
approved_value:
1. Open Dialog: If the task is to open a dialog, check whether the dialog is visible in the screenshot.
2. Enter Name: If the task is to enter a name, verify that the correct name has been typed into the appropriate input field.
...

reasons:
1. If the result of approved_value is "fail", you must provide clear reasoning for your decision. If the result is "success", simply return "success" in this field.
...

You should respond strictly in the following format, and you must not output any comments, explanations, or additional information. Don't include anything beside the requested data represented in the following format

approved_value:
success/fail

reasons:
...

\end{lstlisting}

\newpage

\subsection{Prompt for Supervisor -- Pure-Action Workflow}

\begin{lstlisting}[style=prompt, caption={Example prompt used for Supervisor -- Pure-Action Workflow}, label={lst:example-prompt}]
You will be provided with an image that contains the object information of the created elements and the executed actions. 

Additionally, you are given the current task content: 
<Current Substep>

The list of executed actions: 
<Executed Actions>


You must respond by following the rules below:
approved_value:
At the right side of the image, information about the most recently created or selected object is displayed. You must identify the component name and determine whether it corresponds to the current task content. If the created object matches the task, respond with "success"; otherwise, respond with "fail".

actions:
If the object does not correspond to the task, you must regenerate a corrected list of actions based on the current executed_actions, following the defined action generation rules:
... 


You should respond strictly in the following format, and you must not output any comments, explanations, or additional information. Don't include anything beside the requested data represented in the following format

approved_value:
success/fail

actions:
{{               
    "action name": "...",
    "actions": "['move_mouse_to(x=, y=)', ... ]"
}}

\end{lstlisting}

% \newline

%%%%%%%%%%%%%%%%%%%%%%%%%%%%%%%%%%%%%%%%%%%%%%%%%%%%%%%%%%%%%%%%%%%%%%%%%%%%%%%
%%%%%%%%%%%%%%%%%%%%%%%%%%%%%%%%%%%%%%%%%%%%%%%%%%%%%%%%%%%%%%%%%%%%%%%%%%%%%%%

\end{document}